\PassOptionsToPackage{unicode}{hyperref}
\PassOptionsToPackage{hyphens}{url}
\PassOptionsToPackage{dvipsnames,svgnames,x11names}{xcolor}
\documentclass[
]{article}

\usepackage{amsmath,amssymb}
\usepackage{mathbbol}
\usepackage{iftex}
\ifPDFTeX
  \usepackage[T1]{fontenc}
  \usepackage[utf8]{inputenc}
  \usepackage{textcomp} 
\else 
  \usepackage{unicode-math}
  \defaultfontfeatures{Scale=MatchLowercase}
  \defaultfontfeatures[\rmfamily]{Ligatures=TeX,Scale=1}
\fi
\usepackage{lmodern}
\ifPDFTeX\else  
\fi
\IfFileExists{upquote.sty}{\usepackage{upquote}}{}
\IfFileExists{microtype.sty}{
  \usepackage[]{microtype}
  \UseMicrotypeSet[protrusion]{basicmath} 
}{}
\makeatletter
\@ifundefined{KOMAClassName}{
  \IfFileExists{parskip.sty}{%
    \usepackage{parskip}
  }{
    \setlength{\parindent}{0pt}
    \setlength{\parskip}{6pt plus 2pt minus 1pt}}
}{
  \KOMAoptions{parskip=half}}
\makeatother
\usepackage{xcolor}
\ifx\paragraph\undefined\else
  \let\oldparagraph\paragraph
  \renewcommand{\paragraph}[1]{\oldparagraph{#1}\mbox{}}
\fi
\ifx\subparagraph\undefined\else
  \let\oldsubparagraph\subparagraph
  \renewcommand{\subparagraph}[1]{\oldsubparagraph{#1}\mbox{}}
\fi

\usepackage{longtable,booktabs,array}
\usepackage{calc} 
\usepackage{etoolbox}
\makeatletter
\patchcmd\longtable{\par}{\if@noskipsec\mbox{}\fi\par}{}{}
\makeatother
\IfFileExists{footnotehyper.sty}{\usepackage{footnotehyper}}{\usepackage{footnote}}
\makesavenoteenv{longtable}
\usepackage{graphicx}
\makeatletter
\def\maxwidth{\ifdim\Gin@nat@width>\linewidth\linewidth\else\Gin@nat@width\fi}
\def\maxheight{\ifdim\Gin@nat@height>\textheight\textheight\else\Gin@nat@height\fi}
\makeatother
\setkeys{Gin}{width=\maxwidth,height=\maxheight,keepaspectratio}
\makeatletter
\def\fps@figure{htbp}
\makeatother
\NewDocumentCommand\citeproctext{}{}
\NewDocumentCommand\citeproc{mm}{%
  \begingroup\def\citeproctext{#2}\cite{#1}\endgroup}
\makeatletter
 \let\@cite@ofmt\@firstofone
 \def\@biblabel#1{}
 \def\@cite#1#2{{#1\if@tempswa , #2\fi}}
\makeatother
\newlength{\cslhangindent}
\newlength{\csllabelwidth}
\newenvironment{CSLReferences}[2] 
 {\begin{list}{}{%
  \setlength{\itemindent}{0pt}
  \setlength{\leftmargin}{0pt}
  \setlength{\parsep}{0pt}
  \ifodd #1
   \setlength{\leftmargin}{\cslhangindent}
   \setlength{\itemindent}{-1\cslhangindent}
  \fi
  \setlength{\itemsep}{#2\baselineskip}}}
 {\end{list}}
\usepackage{calc}

\usepackage{arxiv}
\usepackage{orcidlink}
\usepackage{amsmath}
\usepackage[T1]{fontenc}
\makeatletter
\@ifpackageloaded{caption}{}{\usepackage{caption}}
\AtBeginDocument{%
\ifdefined\contentsname
  \renewcommand*\contentsname{Table of contents}
\else
  \newcommand\contentsname{Table of contents}
\fi
\ifdefined\listfigurename
  \renewcommand*\listfigurename{List of Figures}
\else
  \newcommand\listfigurename{List of Figures}
\fi
\ifdefined\listtablename
  \renewcommand*\listtablename{List of Tables}
\else
  \newcommand\listtablename{List of Tables}
\fi
\ifdefined\figurename
  \renewcommand*\figurename{Figure}
\else
  \newcommand\figurename{Figure}
\fi
\ifdefined\tablename
  \renewcommand*\tablename{Table}
\else
  \newcommand\tablename{Table}
\fi
}
\@ifpackageloaded{float}{}{\usepackage{float}}
\floatstyle{ruled}
\@ifundefined{c@chapter}{\newfloat{codelisting}{h}{lop}}{\newfloat{codelisting}{h}{lop}[chapter]}
\floatname{codelisting}{Listing}

\makeatother
\makeatletter
\@ifpackageloaded{caption}{}{\usepackage{caption}}
\@ifpackageloaded{subcaption}{}{\usepackage{subcaption}}
\makeatother
\ifLuaTeX
  \usepackage{selnolig}  
\fi
\usepackage{bookmark}

\IfFileExists{xurl.sty}{\usepackage{xurl}}{} 
\hypersetup{
  pdftitle={Reducing Selection Bias in Large Language Models},
  pdfauthor={Jonathan E. Eicher; Rafael F. Irgoli},
  pdfkeywords={Large Language Models, Cognitive load, List
Selection, Bias, Guard rails},
  colorlinks=true,
  linkcolor={blue},
  filecolor={Maroon},
  citecolor={Blue},
  urlcolor={Blue},
  pdfcreator={LaTeX via pandoc}}

\newcommand{\runninghead}{A Preprint }
\renewcommand{\runninghead}{A Preprint }
\title{Reducing Selection Bias in Large Language Models}
\def\asep{\\\\\\ } 
\author{\textbf{Jonathan E.
Eicher}~\orcidlink{0000-0002-2631-3385atu}\\\href{mailto:jonathan@elsworth.phd}{jonathan@elsworth.phd}\asep\textbf{Rafael
F.
Irgoli\v{c}}\\\href{mailto:cbuclrsbllm@irgolic.com}{cbuclrsbllm@irgolic.com}}
\date{}
\begin{document}
\maketitle
\begin{abstract}
Large Language Models (LLMs) like gpt-3.5-turbo-0613 and
claude-instant-1.2 are vital in interpreting and executing semantic
tasks. Unfortunately, these models' inherent biases adversely affect
their performance Particularly affected is object selection from lists;
a fundamental operation in digital navigation and decision-making. This
research critically examines these biases and quantifies the effects on
a representative list selection task. To explore these biases, we
experiment manipulating temperature, list length, object identity,
object type, prompt complexity, and model. We isolated and measured the
influence of the biases on selection behavior. Our findings show that
bias structure is strongly dependent on the model, with object type
modulating the magnitude of the effect. With a strong primacy effect,
causing the first objects in a list to be disproportionately represented
in outputs. The usage of guard rails, a prompt engineering method of
ensuring a response structure, increases bias and decreases instruction
adherence when to a selection task. The bias is ablated when the guard
rail step is separated from the list sampling step, lowering the
complexity of each individual task. We provide LLM applications and
theoretically suggest that LLMs experience a form of cognitive load that
is compensated for with bias.
\end{abstract}
{\bfseries \emph Keywords}
\def\sep{\textbullet\ }
Large Language Models \sep Cognitive load \sep List
Selection \sep Bias \sep 
Guard rails

\begin{figure}

\centering{

\includegraphics{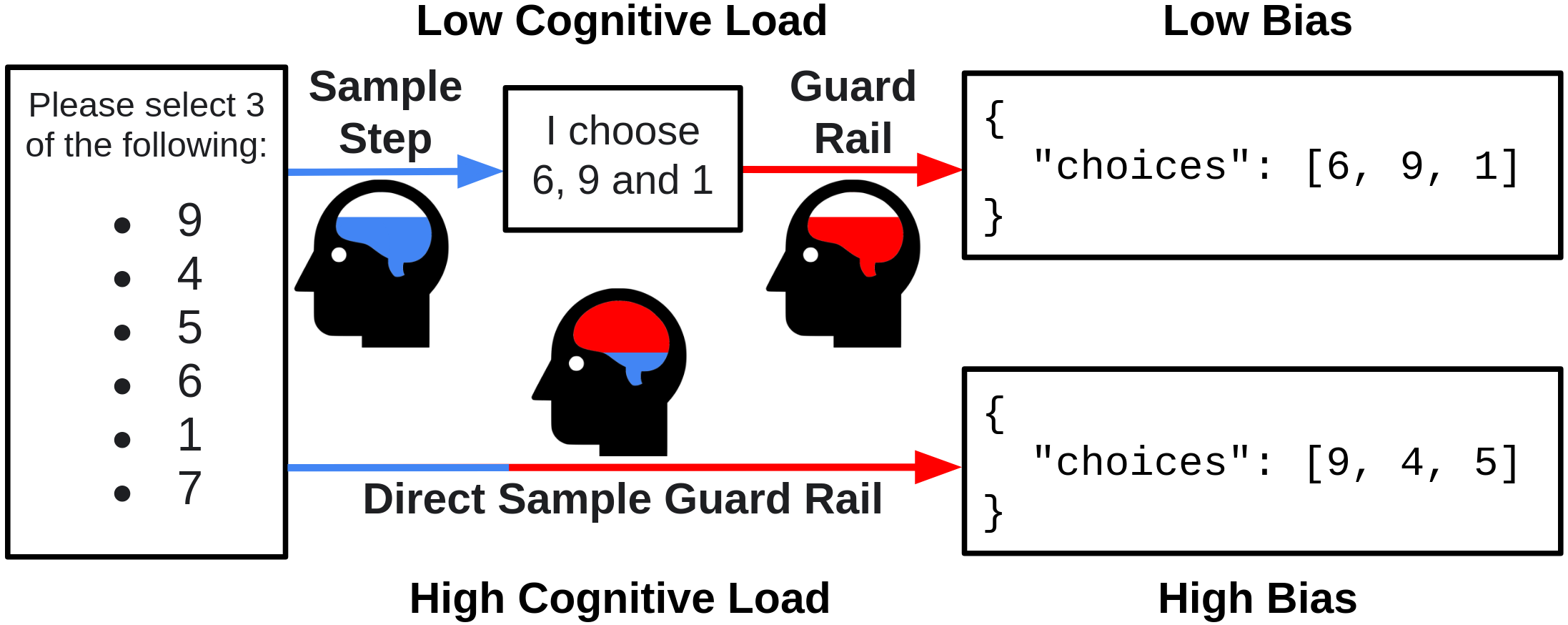}

}

\caption{\label{fig-graphical-abstract}The two data collection methods
used in this paper: a two-step method which separates the sample step
and guard rail, and a direct guard rail method which combines them into
a single step. Each method is annotated with its relative cognitive load
and bias as a result of the method.}

\end{figure}%

\section{Introduction}\label{introduction}

Bias is a quality exhibited by humans and AI alike (Li
(\citeproc{ref-liPrimacyEffectRecency2010}{2010}), Nauts et al.
(\citeproc{ref-nautsFormingImpressionsPersonality2014}{2014})). The bias
we shall concern ourselves with in this paper is selection, where the
order of objects in a list affects selection preference. This bias has
been found in humans, such as in the case of multiple choice exams where
item order affects the average for exams (Balch
(\citeproc{ref-balchItemOrderAffects1989}{1989})). This phenomenon of
multiple choice bias is also well-documented in large language models
(LLMs) (Zheng et al.
(\citeproc{ref-zhengLargeLanguageModels2023}{2023})). Bias is something
central to the human experience and our ability to reduce cognitive load
when making decisions (Allred et al.
(\citeproc{ref-allredWorkingMemorySpatial2016}{2016})). That said,
problems arise when we do not account for how the presentation of
information biases the choices made (Kirk et al.
(\citeproc{ref-kirkBiasOutoftheBoxEmpirical2021}{2021}), Weidinger et
al. (\citeproc{ref-weidingerTaxonomyRisksPosed2022}{2022})).

This is also true of LLMs, which inherit the structural bias of the sum
total of human language (Kinniment et al.
(\citeproc{ref-kinnimentEvaluatingLanguageModelAgents2024}{2024})).
Selection bias is a danger for agentic systems, where a LLM selects
options from a list to decide on the direction of a program, such as
selecting whether to respond to a user, gather more information, or ask
a clarifying question. As agentic systems become more widely adopted
careful consideration must be paid to bias mitigation strategies
(Weidinger et al.
(\citeproc{ref-weidingerTaxonomyRisksPosed2022}{2022})). Of paritcular
note is the ``lost in the middle'' phenomena wherein the LLM is unable
to properly include information at the center of their context
highlights this danger (Liu et al.
(\citeproc{ref-liuLostMiddleHow2023}{2023})). This risk is only
compounded by the observed extant bias present in the training data, and
by extension the output of the LLMs (Kirk et al.
(\citeproc{ref-kirkBiasOutoftheBoxEmpirical2021}{2021}), Touileb,
Øvrelid, and Velldal
(\citeproc{ref-touilebOccupationalBiasesNorwegian2022}{2022}), Wolfe and
Caliskan (\citeproc{ref-wolfeLowFrequencyNames2021}{2021})). While there
are efforts to minimize bias in LLMs, there is a significant lack of
research into the fundamental question of how bias itself manifests in
LLMs (Liang et al.
(\citeproc{ref-liangUnderstandingMitigatingSocial2021}{2021}))

The bias we set out to analyze is how the order in which a LLM is
presented an object affects the choice of object. To extend this
question we investigate how the complexity of the output effects that
bias. There have been some attempts to quantify the bias of LLMs dealing
with selecting objects from lists(Zheng et al.
(\citeproc{ref-zhengLargeLanguageModels2023}{2023})). While interesting
for understanding how multiple choice questions are answered, the
co-variance of the differing ability for a LLM to know the right answer
to a question muddles the direct effect of list order. Similarly, there
has been work in output assurance via guard rails -- an input prompt
structure that ensures the output of a LLM is in a required format,
e.g.~JSON(Han et al.
(\citeproc{ref-hanInformationExtractionSolved2023}{2023}), Rebedea et
al. (\citeproc{ref-rebedeaNeMoGuardrailsToolkit2023}{2023})). Although,
to our knowledge there has been no work on the effects of guard rails on
LLM bias.

\begin{figure}

\centering{

\includegraphics{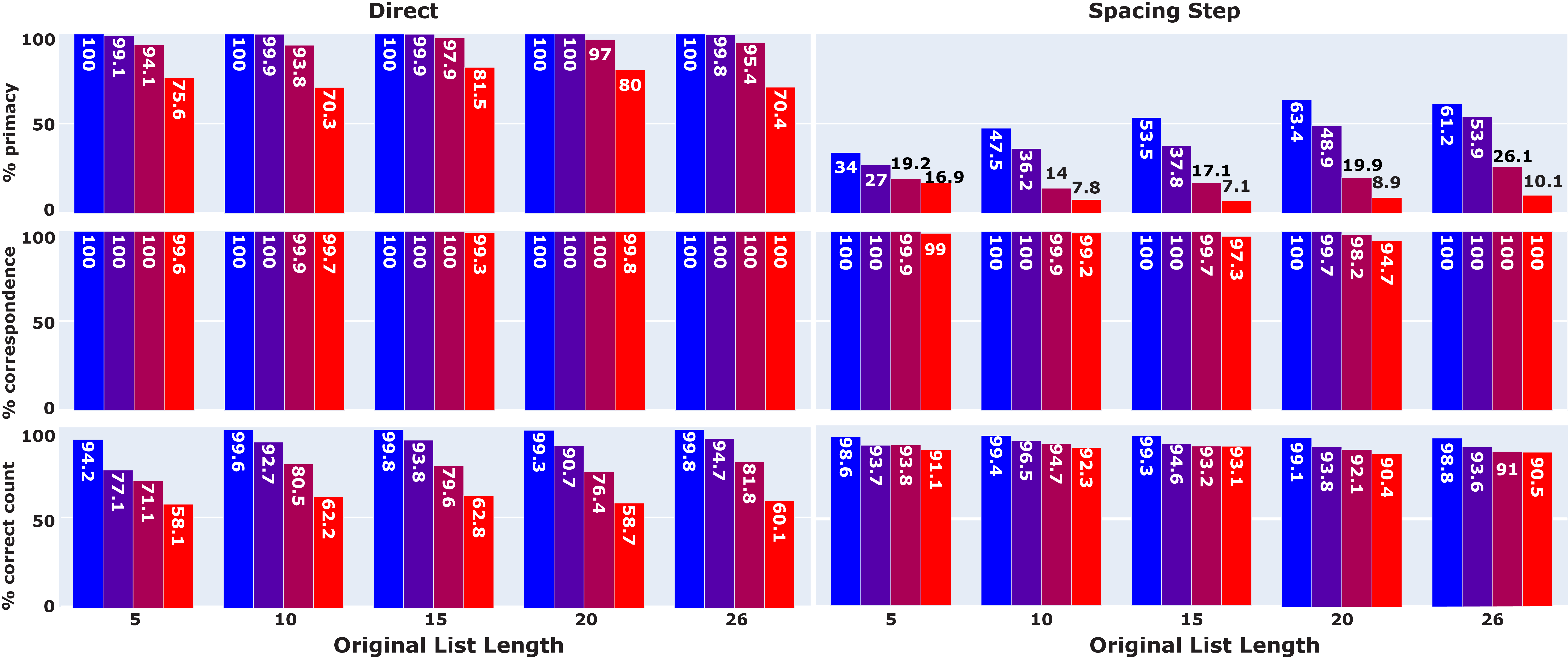}

}

\caption{\label{fig-guard-rails}The percent of responses displaying
primacy bias, correspondence, and correct count for both a direct guard
rail and sample step sampling methodology when selecting numbers from a
list. This was performed via gpt-3.5-turbo-0613 and for a variety of
temperatures (blue 0 to red 1.5 with a step size of 0.5) and list
lengths.}

\end{figure}%

In order to address this gap we developed a simple model problem and
rigorously analyzed the selection probabilities of two LLMs:
gpt-3.5-turbo-0613 and claude-instant-1.2. The toy problem we devised
was the selection of three objects from a variable-length list. This toy
example can be used to extract guiding principles for the analysis of
bias in LLMs and how to implement agentic frameworks that prevent
exacerbation of these biases.

We selected several variables for our toy problem of selecting three
objects from a list: the list length, object identities, temperature,
and model. We analyzed the probabilities of the values being selected
given an object's identity and its position; allowing us to
systematically evaluate the effects of temperature modulation, list
length modulation, model type, object identity, object type, and
position on the selection probabilities. We used guard rails derived
from Guardrails AI to parse our outputs and test their effects on
selection biases.

We found that primacy bias is indeed a problem in our context, but is
significantly model dependent (Figure~\ref{fig-bias-cot}). The identity
and position of our objects significantly altered the selection
probabilities (Figure~\ref{fig-letter-position-gpt}). When comparing
models we found that biases on position and letters were not consistent,
with variation in every form of bias measured
(Figure~\ref{fig-letter-position-claude}). A guard rail significantly
alters the primacy bias and instruction adherence for all considered
models (Figure~\ref{fig-guard-rails}). We hypothesize that this is a
result of the cognitive load of the guard rail, which may be causing
human-like compensatory behavior (de Jong
(\citeproc{ref-dejongCognitiveLoadTheory2010}{2010}), Xu et al.
(\citeproc{ref-xuCognitiveOverloadJailbreaking2023}{2023})).

\section{Methods}\label{methods}

Two LLMs were accessed via their APIs: anthropic-instant-1.2 and
gpt-3.5-turbo. We tested a range of temperatures for each model:
\([0, 0.5, 1, 1.5]\) for gpt-3.5-turbo, and \([-1, 0, 0.5, 1]\) for
claude-instant-1.2. These models were presented with a list of letters
or numbers of modular length, prepended with an instruction. The output
of the first LLM call (Listing~\ref{lst-prompt-construct}) was then fed
into a guard rail to extract the choices
(Listing~\ref{lst-spacing-step-collection}), as opposed to the direct
method, which implemented list selection and guard rail application in
one LLM call (Listing~\ref{lst-direct-data-collection}). This process
was repeated \(N=1000\) times for each temperature, input list length,
and model. This allowed us to complete all analysis for a given model,
temperature, input, and list length. If the output was not in a JSON
format or the output was not the same length as the input the trial was
discarded and noted as having failed the correct count condition.

The input list was selected from a pool of object \(\mathbb{X}_p\) with
a length of \(n_p\), which in our study was 26, referring to the numbers
1-26 and the letters A-Z: \[\mathbb{x}_{p}=\begin{bmatrix}
1 \\
2 \\
\dots \\
n_p \\
\end{bmatrix}\]

\(\mathbb{X}_p\) was then uniformly sampled with a list length, \(n_t\),
of 5, 10, 15, 20, or 26. The input list was ordered so when analyzing we
considered it as an \(n_t \times 2\) matrix where \(\ell\) is the object
and \(p\) is the position:

\[\mathbb{x}_{t}= \begin{bmatrix}
a,1 \\
d,2 \\
q,3 \\
\dots \\
\ell, p_{t}
\end{bmatrix} \]

The outputs were sent to a LLM through either a direct method or with a
sample step. The direct method presents the input list alongside a guard
rail to the LLM, ensuring the output is in a JSON format. The sample
step queries the LLM to select objects from the list and the output is
then sent to a guard rail for extraction. The number of objects
requested from the list, \(n_s\) in our study, was always \(3\). As the
output was also an ordered list, the position of the selected objects
was also recorded, meaning the final output of a given LLM list sampling
was a \(n_s \times 3\) matrix.

\[\phi(\mathbb{x}_t) \to \mathbb{x}_{s}= \begin{bmatrix}
a,1,1 \\
c,4,2 \\
d,2,3 \\
\dots \\
\ell,p_{t}, p_{u}
\end{bmatrix} \]

Where \(p_u\) is a position in the output list, \(p_t\) is a position in
the input list, and \(\ell\) is an object. In order to compute the
probability of an object being selected we summed the number of times it
was selected and divided it by the total number of possible selections
that could have occurred, \(\ell_{i} \in  \mathbb{X}_{t}\). Similarly we
computed the probability of a position being selected by summing the
number of times that position was selected and dividing it by the total
number of possible selections that could have occurred,
\(p_{t,i} \in  \mathbb{X}_{t}\).

\[P(p_{t,i} \in \mathbb{x}_{s})=\frac{\# \hspace{0.1 cm} p_{t,i} \in \mathbb{X}_{s}}{\# \hspace{0.1 cm} p_{t,i} \in  \mathbb{X}_{t}}\]
\[P(\ell_{i} \in \mathbb{x}_{s})=\frac{\# \hspace{0.1 cm} \ell_{i} \in  \mathbb{X}_{s}}{\# \hspace{0.1 cm} \ell_{i} \in  \mathbb{X}_{t}}\]

The outputs were then processed to extract relevant features. First, we
tested for primacy bias; whether the outputs were of the first three
positions, returned in the exact same order. Primacy can be expressed
as:
\[\% \text{Primacy} = \frac{\# \mathbb{p}_{t} \subset \mathbb{x}_{s} : \mathbb{p}_t = \{1,2,3\}}{N}\]

Next, they were tested for correspondence hallucinations; whether the
output objects were present in the input list. We can define this
correspondence mathematically as:
\[\% \text{Correspondence} = \frac{ \# \hspace{0.1 cm} \mathbb{l}_{s} \subset \mathbb{l}_{t}}{N}\]

Finally, they were examined for adherence to instructions; whether the
exact number of objects specified were selected from the input list
(Correct Count). We can define this adherence mathematically as:
\[\% \text{Correct count} = \frac{\# n_{s,target} = n_s}{N}\]

We are assuming that the selections of objects and positions are
independent of each other. In the case that they are not, we can more
weakly assume the random distribution of the objects and positions
should produce an average probability. In order to compute the
probability and error robustly we performed bootstrapping using 3000
samples with replacement.

While this tells us information about the probability of an object or
position being selected it does not tell us about the probability of an
object being selected given a position. To compute the joint probability
\(P(\ell_i \cap p_i,t)\), we count the number of times an object was
selected from a position and divide it by the total number of times that
position and object occurred concurrently.

\[P(\ell_i \cap p_i,t) = \frac{\# \hspace{0.1 cm} (\ell_i, p_{i,t}) \in  \mathbb{x}_s}{\# \hspace{0.1 cm} (\ell_i, p_{i,t}) \in  \mathbb{x}_t}\]

Mutual information, a measure of how much information the identity of
one variable provides on another variable, was computed using the
following equation:

\[I(L_{s};P_{t})=\sum_{\ell \in L_{s}} \sum_{p_{t} \in P_{t}} P_{(L_{s},P_{t})}(\ell,p)\log\left( \frac{P_{(L_{s},P_{t})}(\ell,p)}{P_{L_{s}}(\ell)P_{_{P_{t}}}(p)} \right)\]

Where \(L_{s}\) is the distribution of inputs in the selected list, and
\(P_{t}\) is the distribution of positions from the input list that were
selected. In this case the mutual information describes how knowledge of
an input being selected gives us certainty on which position it was in.

In short, we designated a set of 26 objects from the alphabet (A-Z) or
the numbers(1-26). We then composed a set of lists of lengths 5, 10, 15,
20, or 26 from these objects. We then calculated the probabilities of an
object being selected given a uniform distribution over these lists and
compared this against the observed probability. This was repeated with
the probability that a position was selected, and the position that was
placed in. We also calculated the number of hallucinated answers
(correspondence), answers where instructions were not followed (correct
count), and answers that were just the first three objects in the list
returned in order (primacy). Finally we determined the relationship
between positions and objects by computing the mutual information.

\section{Results}\label{results}

\subsection{Primacy, Hallucination and
Adherence}\label{primacy-hallucination-and-adherence}

Primacy bias, for the purpose of this paper, is defined as the bias for
a LLM to select the first three objects in a list regardless of their
identity. Correspondence refers to the ability of the LLM to return only
objects in the input list, a measure of how common hallucination was.
Correct count refers to the ability of the LLM to return the correct
number of objects from the list, a measure of the LLM's ability to
adhere to instructions.

\begin{figure}

\centering{

\includegraphics{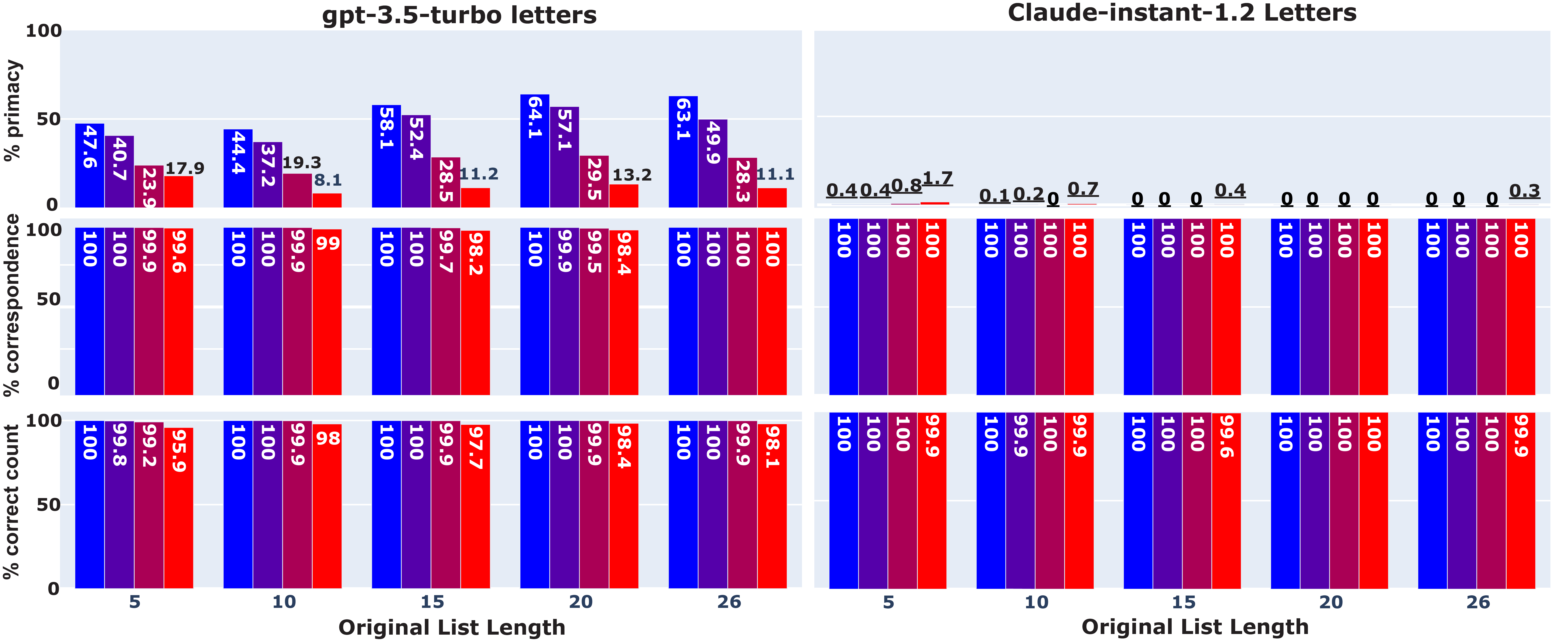}

}

\caption{\label{fig-bias-cot}The percent of responses displaying primacy
bias, correspondence, and correct count The temperature is denoted by
color for gpt-3.5-turbo and claude-instant-1.2 respectively from blue
(0, -1), purple (0.5, 0), dark red (1, 0.5) and red (1.5, 1).}

\end{figure}%

\begin{figure}

\begin{minipage}{0.33\linewidth}
\includegraphics{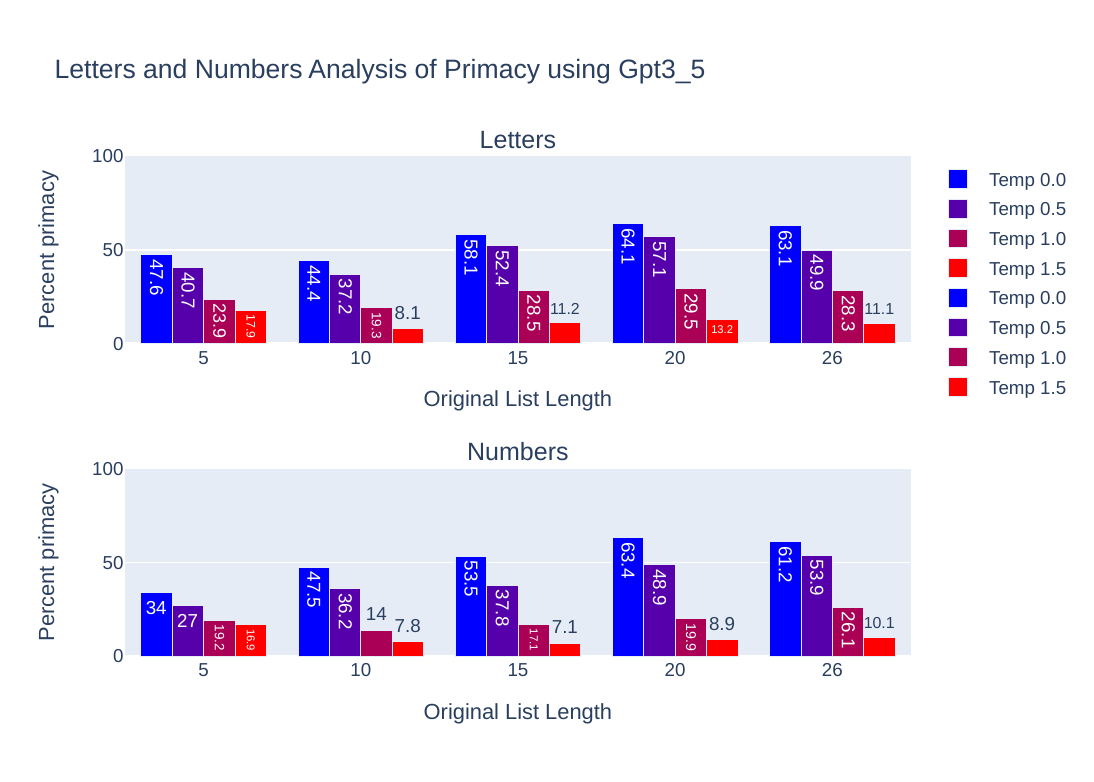}\end{minipage}%
\begin{minipage}{0.33\linewidth}
\includegraphics{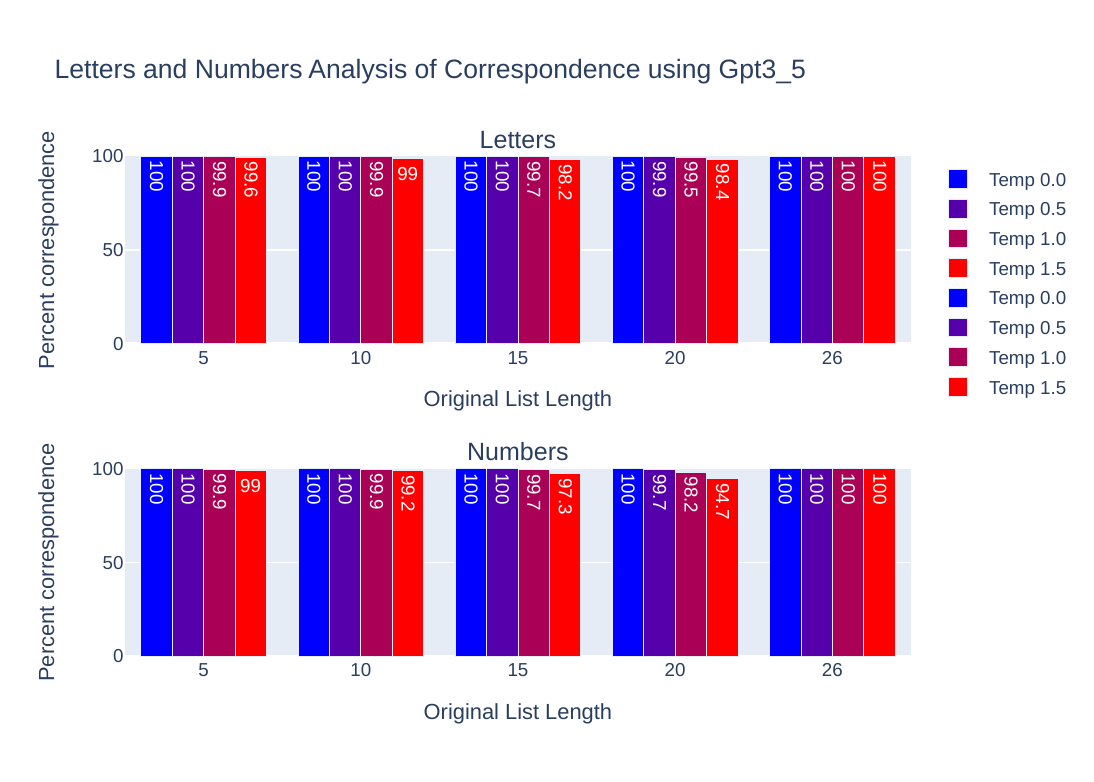}\end{minipage}%
\begin{minipage}{0.33\linewidth}
\includegraphics{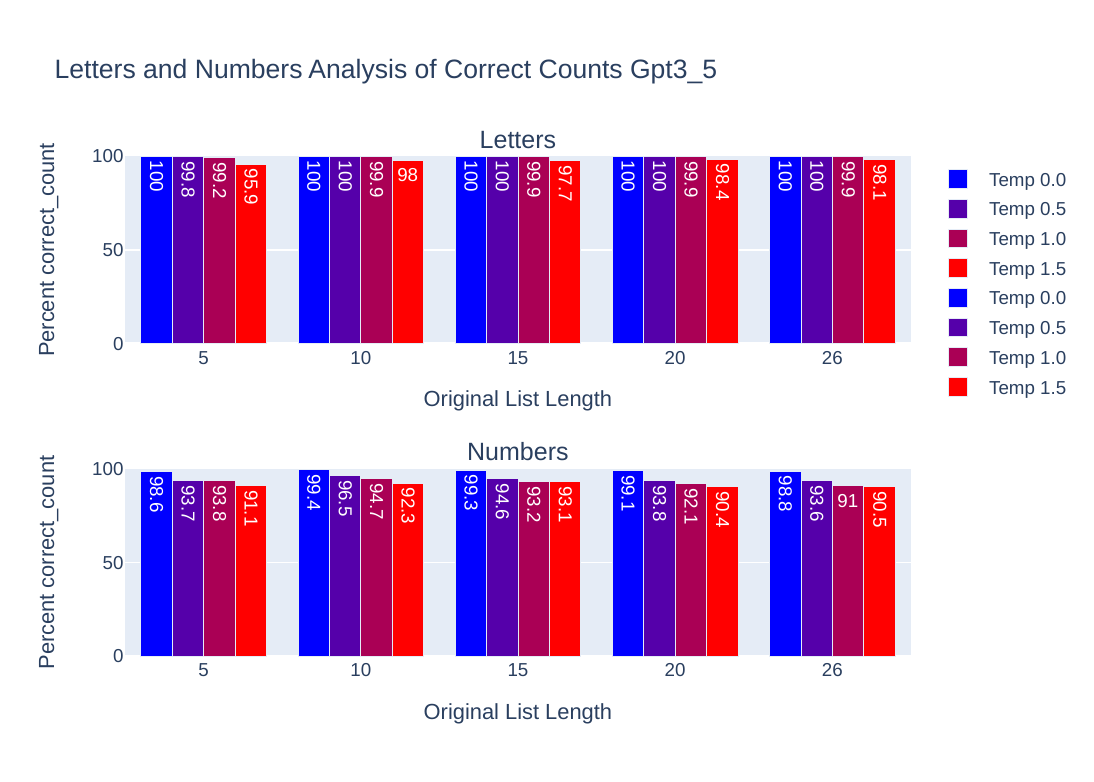}\end{minipage}%

\caption{\label{fig-gpt35-full-bias}The full results of the
gpt-3.5-turbo model with a spacing step marked on primacy,
correspondence, and correct counts. This is compared for all list
lengths analyzed and all temperatures.}

\end{figure}

\begin{figure}

\begin{minipage}{0.33\linewidth}
\includegraphics{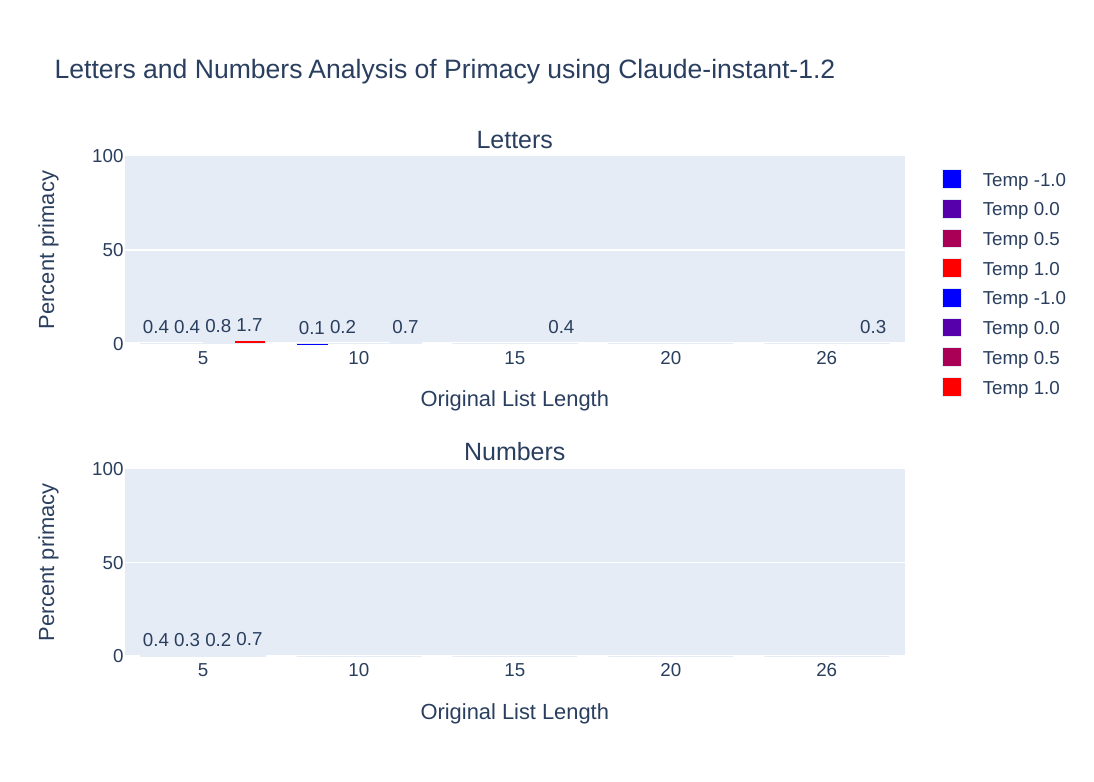}\end{minipage}%
\begin{minipage}{0.33\linewidth}
\includegraphics{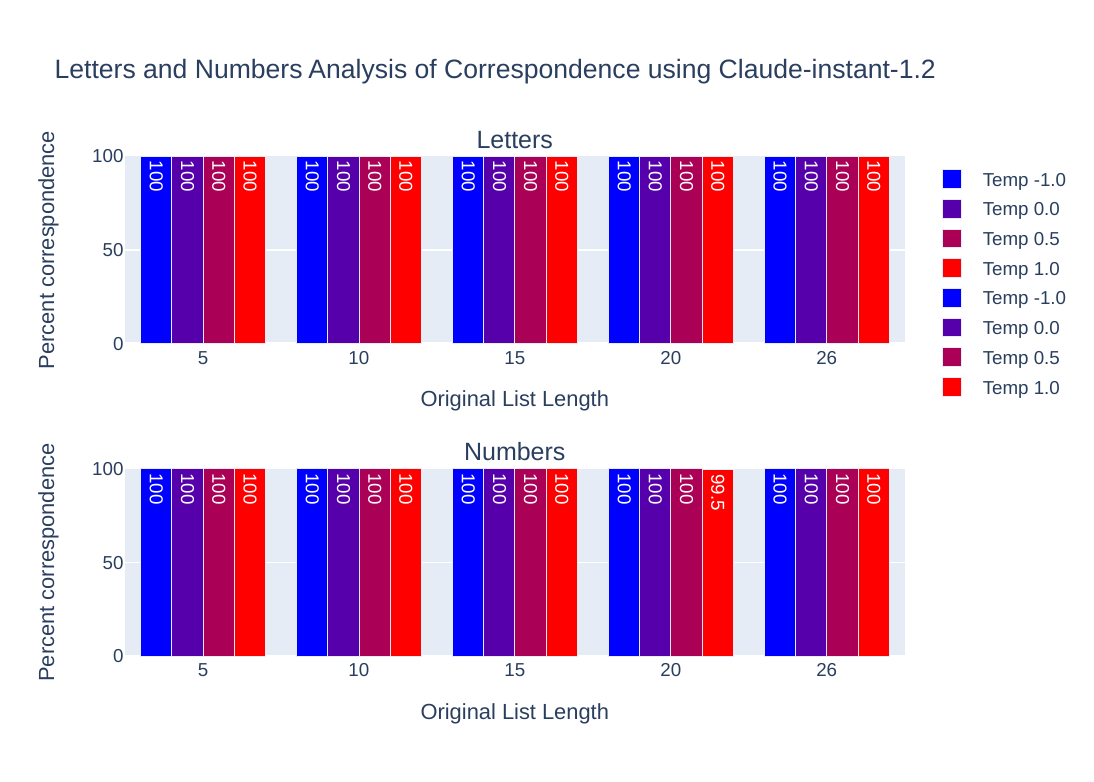}\end{minipage}%
\begin{minipage}{0.33\linewidth}
\includegraphics{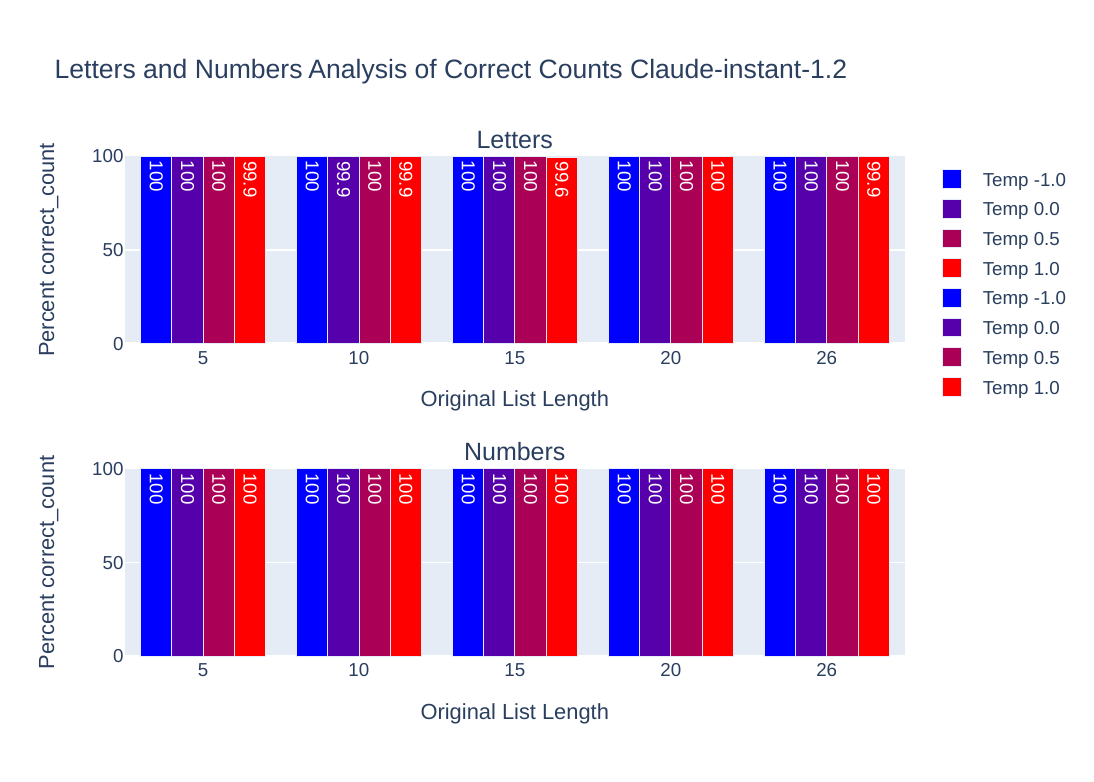}\end{minipage}%

\caption{\label{fig-claude-full-bias}The full results of the claude
model with a spacing step marked on primacy, correspondence, and correct
counts. This is compared for all list lengths analyzed and all
temperatures. For primacy, all empty values are 0 as there was no
occurance of primacy.}

\end{figure}

For all conditions gpt-3.5-turbo showed a significantly increased
primacy bias compared to claude-instant-1.2 (Figure~\ref{fig-bias-cot}).
This trend was consistent across all temperatures and list lengths.
Primacy bias with temperature is correlated negatively for gpt-3.5-turbo
and positively for claude-instant-1.2. Both showed similar levels of
correspondence and correct counts, and gpt-3.5-turbo showed a slightly
lower accuracy in both cases. When altering the object type of a list,
numbers vs letters, the usage of numbers appears to reduce the primacy
bias in most cases while simultaneously reducing instruction following
behavior, correct count, for gpt-3.5-turbo
(Figure~\ref{fig-gpt35-full-bias}). Only in a couple isolated conditions
for claude-instant-1.2 (Figure~\ref{fig-claude-full-bias}) did the list
length and expected probability of primacy correspond with each other
(Table~\ref{tbl-primacy-length}).

\begin{longtable}[]{@{}cc@{}}
\caption{Given a uniform distribution of selecting three objects from a
list the chance that you will return the first three objects in
order.}\label{tbl-primacy-length}\tabularnewline
\toprule\noalign{}
List Length & Probability of Primacy \\
\midrule\noalign{}
\endfirsthead
\toprule\noalign{}
List Length & Probability of Primacy \\
\midrule\noalign{}
\endhead
\bottomrule\noalign{}
\endlastfoot
5 & 1.7\% \\
10 & 0.14\% \\
15 & 0.034\% \\
20 & 0.015\% \\
26 & 0.0064\% \\
\end{longtable}

List length did seem to have some minor effects on primacy bias, such as
at high temperatures the minima were between a list length of
(\(10-15\)) for gpt-3.5-turbo (Figure~\ref{fig-bias-cot}). Which was
observed in both the number and letter condition
(Figure~\ref{fig-gpt35-full-bias}), although low list lengths were
associated with a reduced primacy bias at low temperatures.

\subsection{Positional Selections}\label{positional-selections}

The probability that a position will be selected was calculated with
respect to a variety of temperatures, list lengths, models, and
temperatures. The probability was then split by primacy, so we could
detect differences in the selections. Claude-instant-1.2 had negligible
primacy bias and as such saw little by way of modulation.

\begin{figure}[H]

\centering{

\includegraphics{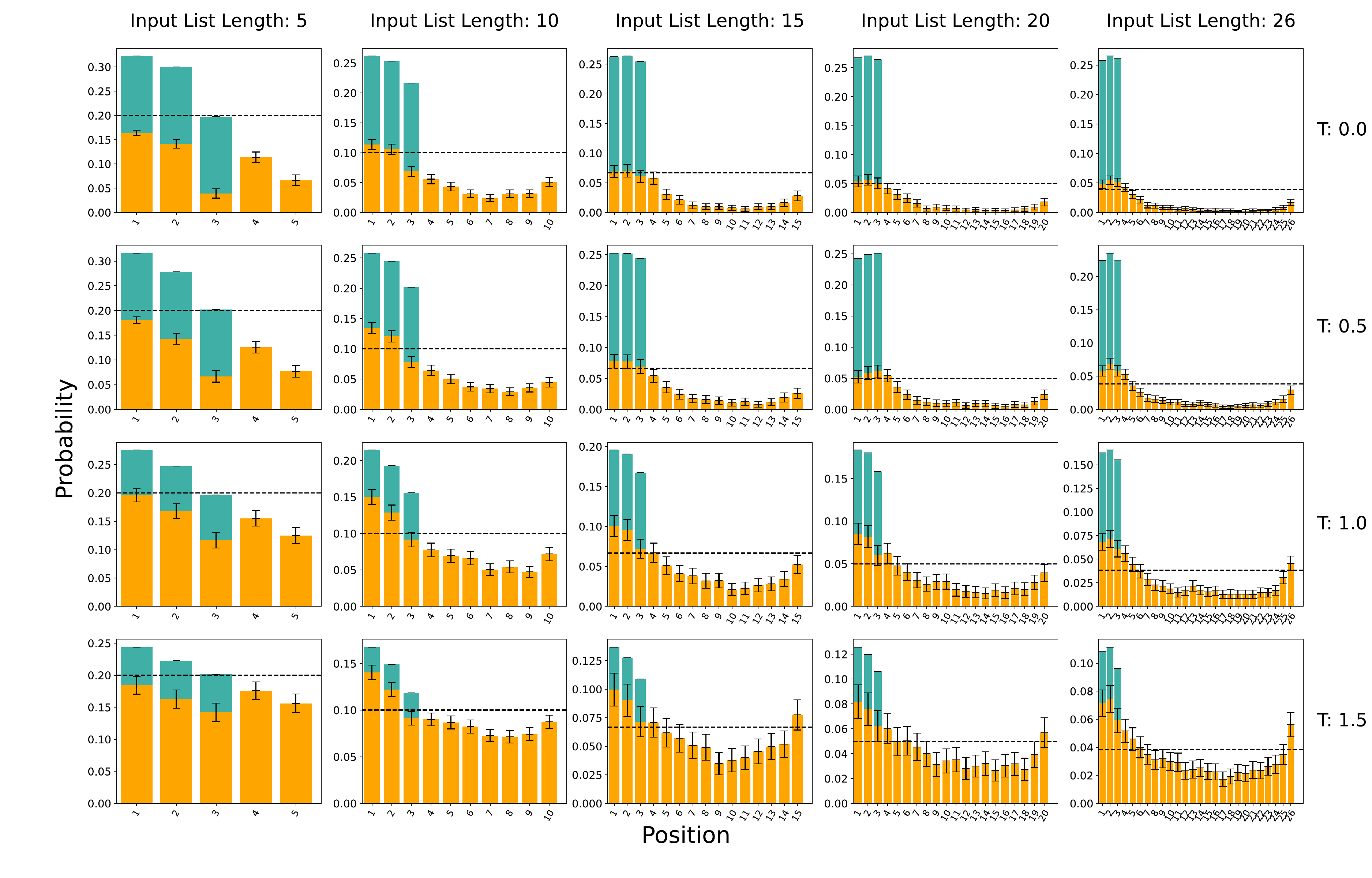}

}

\caption{\label{fig-letter-position-gpt}Using gpt-3.5-turbo, the scaled
probability that a position of a letter will be selected given a
temperature and original list length. The dotted black line is the
expected probability given random sampling of a uniform distribution for
a list length. The orange bars are the probability that a position
without primacy bias will be selected, while blue represents the
probability that a position with primacy bias will be selected. Error
bars are the standard error from \(3000\) bootstrap replicates.}

\end{figure}

As primacy is a feature defined by position, the first three positions
were equally affected by the proportions of the results. In the case of
list length \(5\), the results were restorative at all temperatures to a
seemingly linear decrease in probability that a position will be
selected. In this case, temperature largely affected the steepness of
the linear trend, with \(T:0\) resulting in a probability of \(~0.05\)
for position \(5\) and \(T:1.5\) resulting in a probability of \(0.15\)
(Figure~\ref{fig-letter-position-gpt}).

\begin{figure}

\centering{

\includegraphics{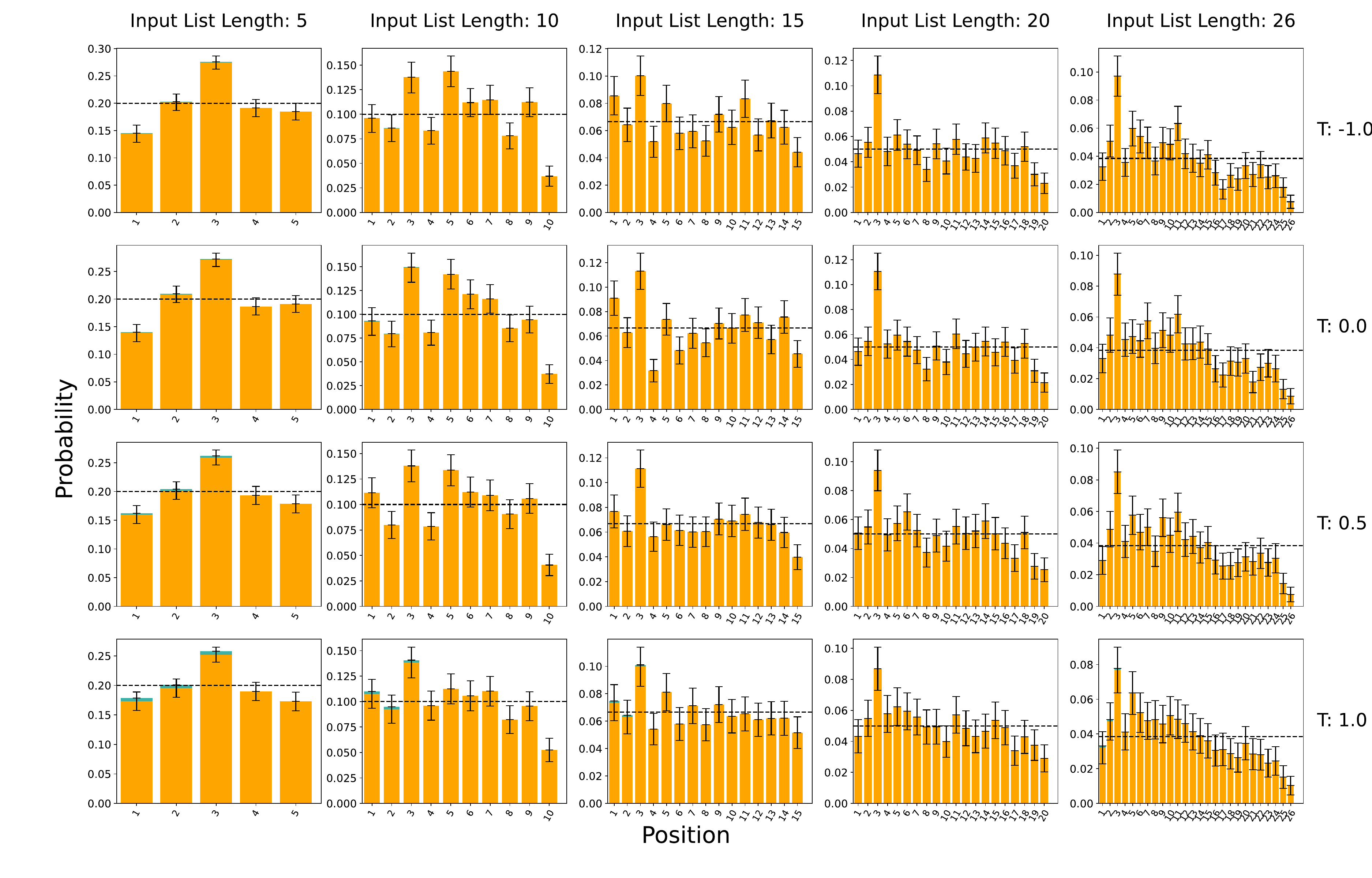}

}

\caption{\label{fig-letter-position-claude}Using claude-instant-1.2, the
scaled probability that a position of a letter will be selected given a
temperature and original list length. The dotted black line is the
expected probability given random sampling of a uniform distribution for
a list length. The orange bars are the probability that a position
without primacy bias will be selected, while blue represents the
probability that a position with primacy bias will be selected. Error
bars are the standard error from \(3000\) bootstrap replicates.}

\end{figure}

As the list length increases, the non-primacy probability of a
positional selection forms a basin where the start and end of a list are
more likely to be selected. Even without primacy bias the most
frequently selected position tends to be the first, second, or third.
Increasing temperature tends to minimize this effect and increases the
probability of the last few positions being selected as well. This
effect was specific to gpt-3.5-turbo, with claude-instant-1.2 showing
only a return to the expected uniform distribution as a function of
temperature (Figure~\ref{fig-letter-position-claude}).

Claude-1.2-instant had a significant bias against the first position at
low list lengths, with a compensatory gain towards the third position.
The relative bias toward the third position was retained, and even
magnified as the input list length increased
(Figure~\ref{fig-letter-position-claude}). There was also a significant
bias against the last position in the list that appears insensitive to
the temperature, while having a complex relationship with list length.

While in the case of gpt-3.5-turbo the probability of a position being
selected was qualitatively invariant of object type (letter vs.~number),
claude-instant-1.2 showed significant modulations in profile
(Figure~\ref{fig-number-position-claude}). The bias against the first
position was higher for numbers than letters, and never reached the
expected uniform distribution. The bias for the third position, while
present, was weaker. Overall, the numbers appear to have more specific
positional preferences that were invariant to temperature.

\subsection{Input Object Selection}\label{input-object-selection}

The probability of an input being selected given that it appeared within
a list was computed. Primacy was considered separately from non-primacy
in order to clarify their effect magnitudes on selection. The
probability of a given input being selected was then compared to the
expected probability of a uniform distribution.

\begin{figure}

\centering{

\includegraphics{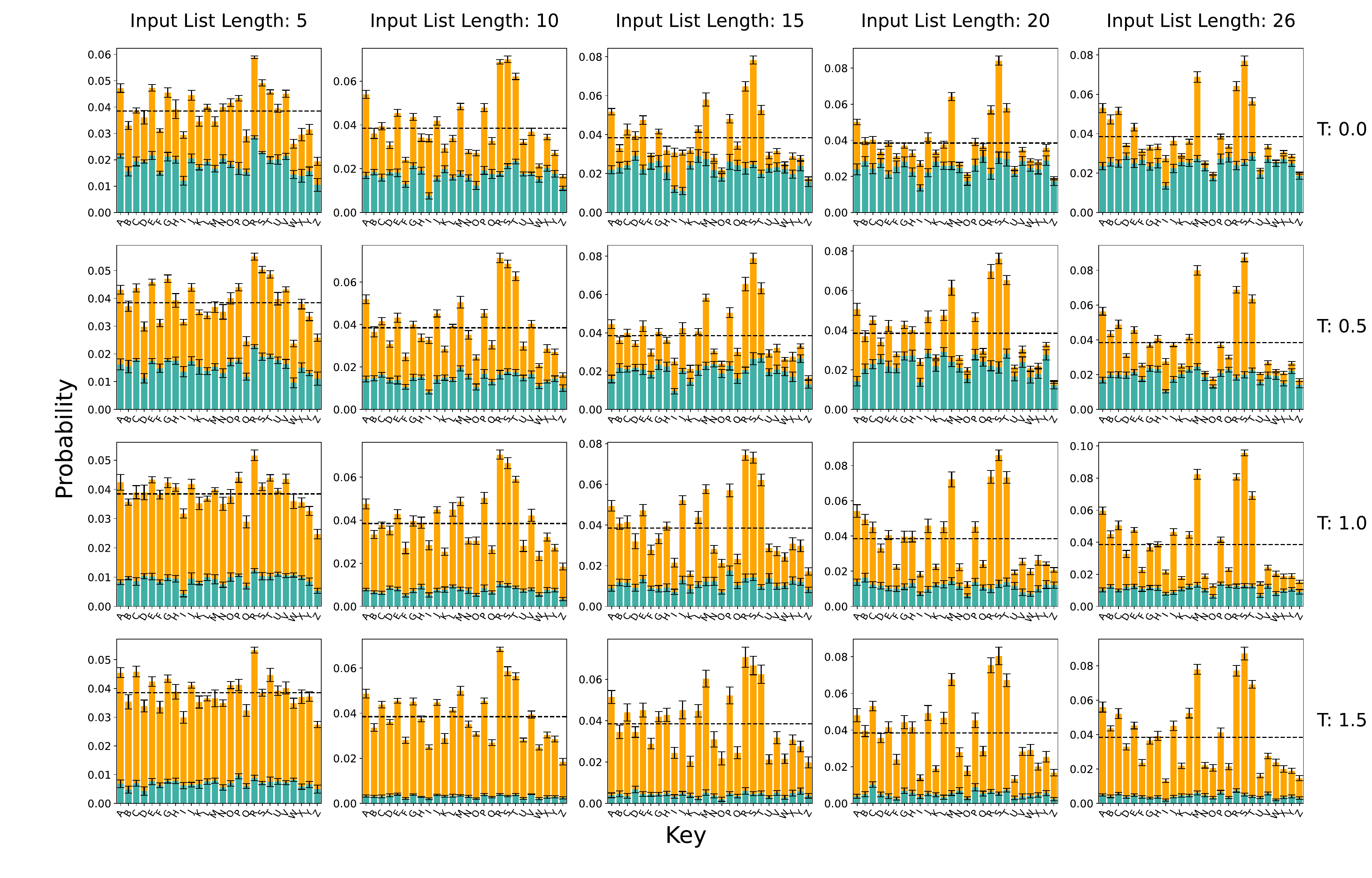}

}

\caption{\label{fig-letter-input-probability-gpt}Using gpt-3.5-turbo,
the scaled probability that a letter will be selected given a
temperature and original list length. The dotted black line is the
expected probability given random sampling of a uniform distribution for
a list length. The orange bars are the probability that a position
without primacy bias will be selected, while blue represents the
probability that a position with primacy bias will be selected. Error
bars are the standard error from \(3000\) bootstrap replicates.}

\end{figure}%

\begin{figure}

\centering{

\includegraphics{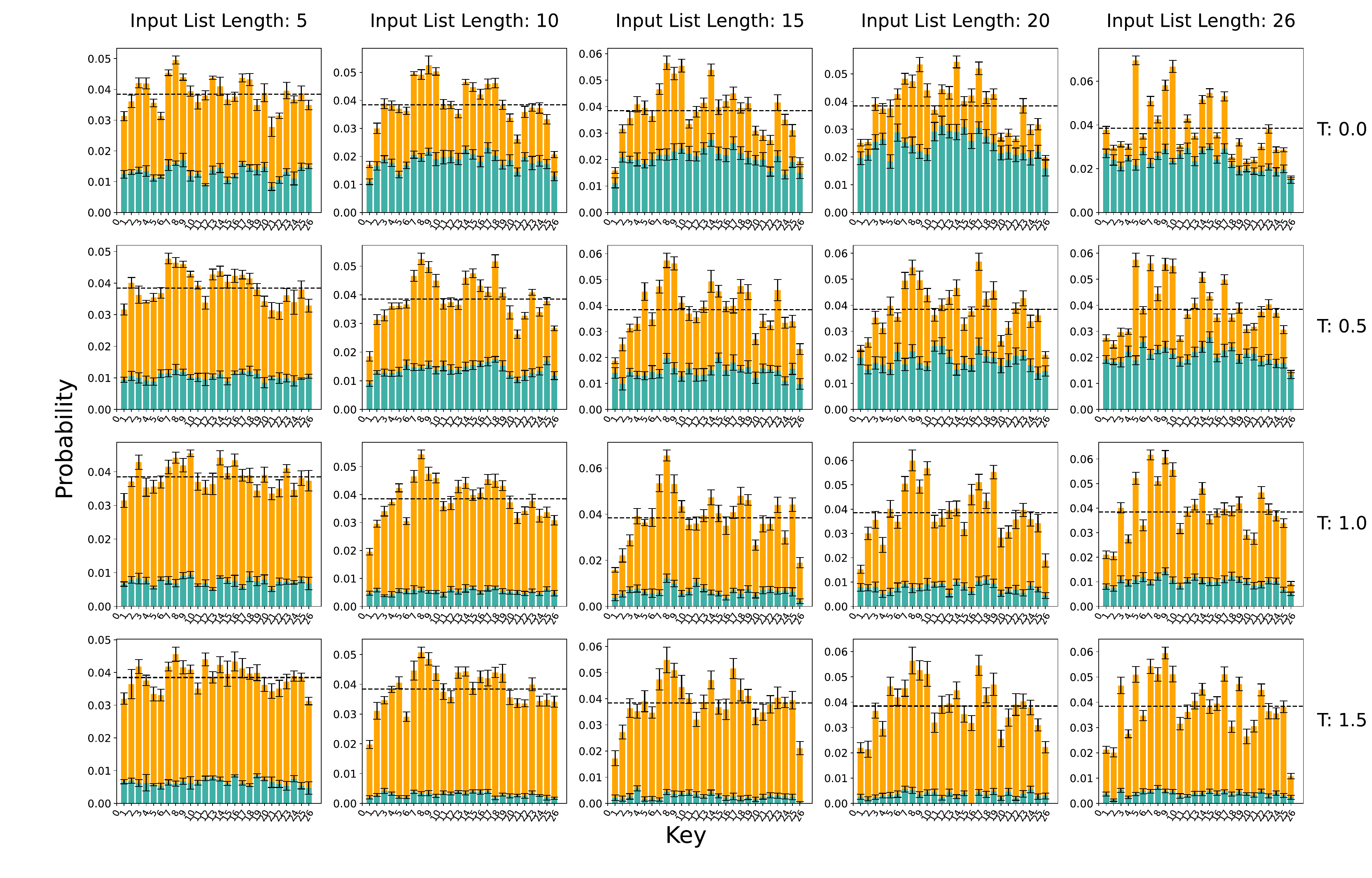}

}

\caption{\label{fig-number-input-probability-gpt}Using gpt-3.5-turbo the
scaled probability that a number will be selected given a temperature
and original list length. The dotted black line is the expected
probability given random sampling of a uniform distribution for a list
length. Orange bars are the probability that a position without primacy
bias will be selected, while blue represents the probability that a
value with primacy bias will be selected. Error bars are standard error
from 3000 bootstrap replicates.}

\end{figure}

Primacy played a significant role in contributing to the total
probability for gpt-3.5-turbo, but showed few sensible effects. Several
letters showed consistently low probability of being selected via
primacy such as the letter I and the letter Z
(Figure~\ref{fig-letter-input-probability-gpt}). What was interesting
was the variation in the non-primacy selection probabilities. There was
significant bias toward or away from certain letters that produced
distributions that grew increasingly non-uniform as the list length
increased. Temperature had a positive correlation with returning toward
the uniform distribution, but was unable to fully restore it within our
study parameters. These effects were qualitatively similar in the case
of number based inputs, but of a smaller magnitude
(Figure~\ref{fig-number-input-probability-gpt}).

\begin{figure}

\centering{

\includegraphics{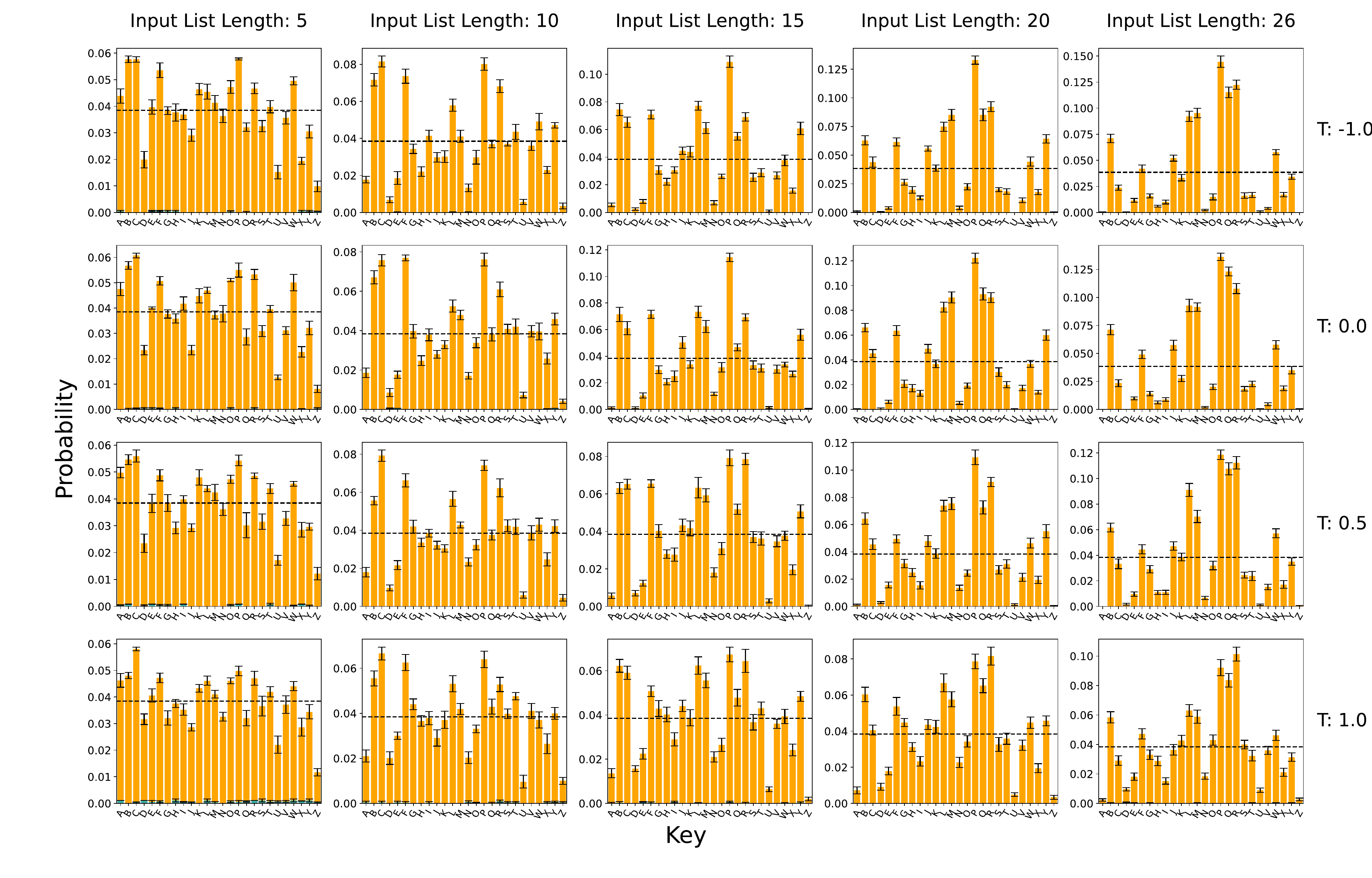}

}

\caption{\label{fig-letter-input-probability-claude}Using
claude-instant-1.2, the scaled probability that a letter will be
selected given a temperature and original list length. The dotted black
line is the expected probability given random sampling of a uniform
distribution for a list length. The orange bars are the probability that
a position without primacy bias will be selected, while blue represents
the probability that a position with primacy bias will be selected.
Error bars are the standard error from \(3000\) bootstrap replicates.}

\end{figure}%

Claude-instant-1.2 was extremely biased when it came to object
selection. At low list lengths the uniformity of the sampling was
relatively high, but showed that several letters were highly selected
against (Figure~\ref{fig-letter-input-probability-claude}). As the list
length increased, the bias against the letters increased in magnitude
until several objects showed effectively zero probability of being
selected. Temperature modulation alleviated the bias, but could not
restore the uniformity of the distribution. When considering numbers,
the bias only grew worse. At low list lengths the probability
distribution resembles a unimodal symmetric form where the first and
last positions are selected against
(Figure~\ref{fig-number-input-probability-claude}). As the list length
increases, multimodality appears and multiple high probability peaks and
valleys occur where certain numbers are heavily selected for or against,
similarly to the letter input selection probabilities.

\begin{figure}

\centering{

\includegraphics{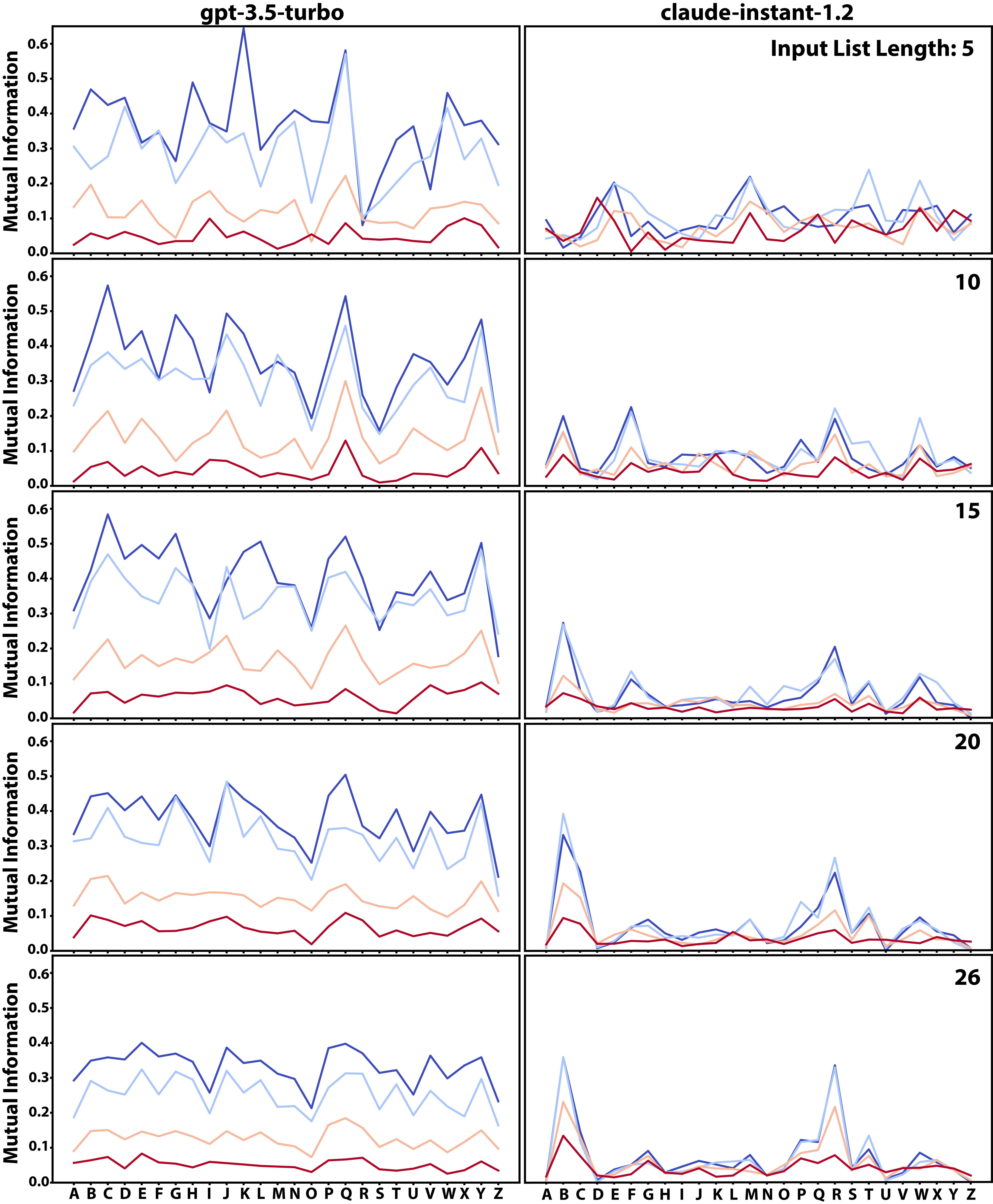}

}

\caption{\label{fig-mi-scores-letters}Mutual information between input
letter and position. The temperature is denoted by color for
gpt-3.5-turbo and claude-instant-1.2, respectively being dark blue
\((0, -1)\), light blue \((0.5, 0)\), light red \((1, 0.5)\) and dark
red \((1.5, 1)\). Each row represents a different initial list length
from 5-26 letters.}

\end{figure}%

\subsection{Mutual Information}\label{mutual-information}

Mutual information allows for the quantification of how two random
variables are linked. If there is a high amount of mutual information,
knowledge of one variable will give significant information on another.

\begin{figure}

\centering{

\includegraphics{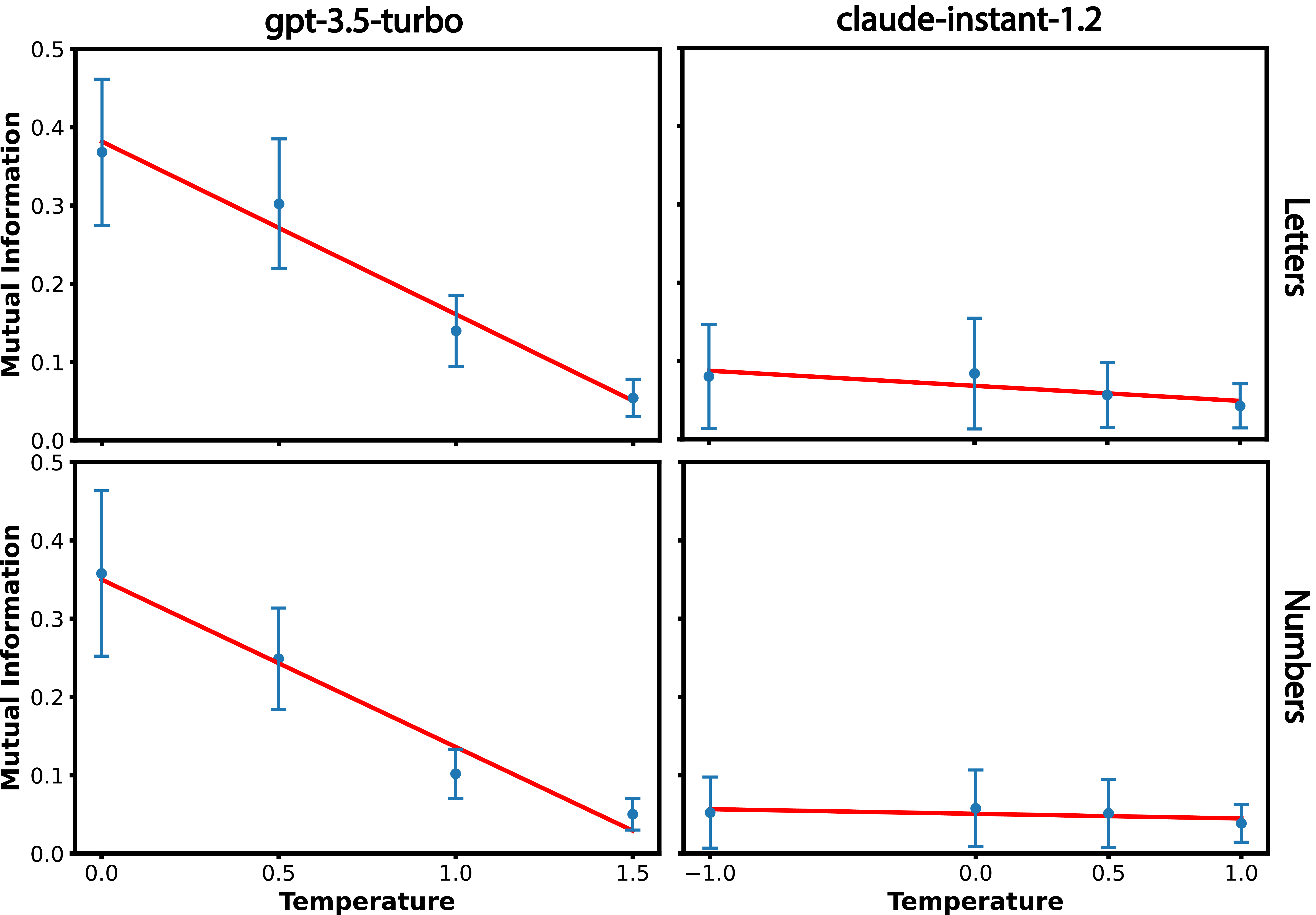}

}

\caption{\label{fig-mutual-regression}Average mutual information between
an input and position with standard deviations. Averages were computed
over all list lengths and inputs of a given type. A linear regression
was computed for letter inputs (A), (gpt-3.5-turbo, \(R^2=0.971\) and
claude-instant-1.2, \(R^2=0.703\)), and number inputs (B),
(gpt-3.5-turbo, \(R^2=0.975\) and claude-instant-1.2, \(R^2=0.394\)).}

\end{figure}

In the case of our work, we computed the mutual information of a
selected object's identity and position on the input list. The higher
the mutual information, the more we learn about the possible position of
a chosen object if we learn the object's identity. Claude-instant-1.2
showed significantly lower mutual information overall when compared to
gpt-3.5-turbo (Figure~\ref{fig-mutual-regression}). Importantly, the
mutual information for gpt-3.5-turbo was negatively correlated with
temperature.

\begin{figure}

\begin{minipage}{0.50\linewidth}

\begin{figure}[H]

{\centering \includegraphics{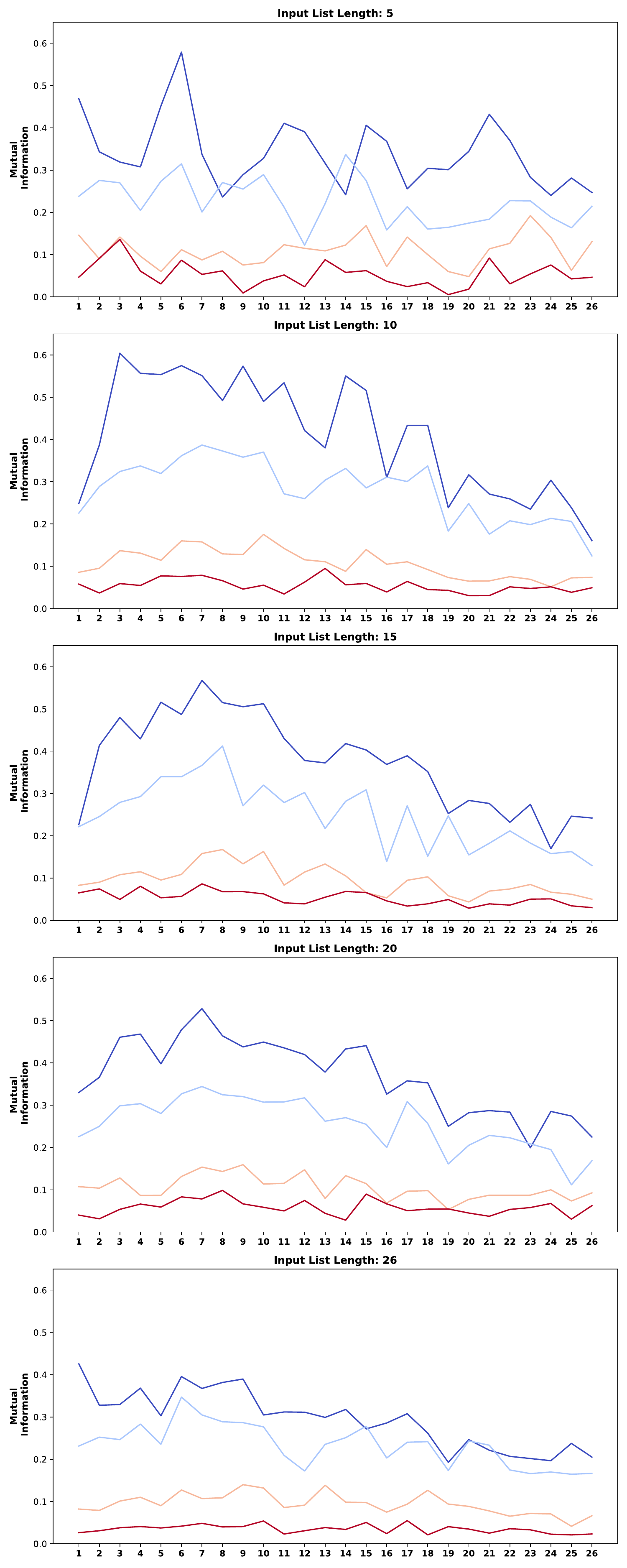}

}

\subcaption{gpt-3.5-instant}

\end{figure}%

\end{minipage}%
\begin{minipage}{0.50\linewidth}

\begin{figure}[H]

{\centering \includegraphics{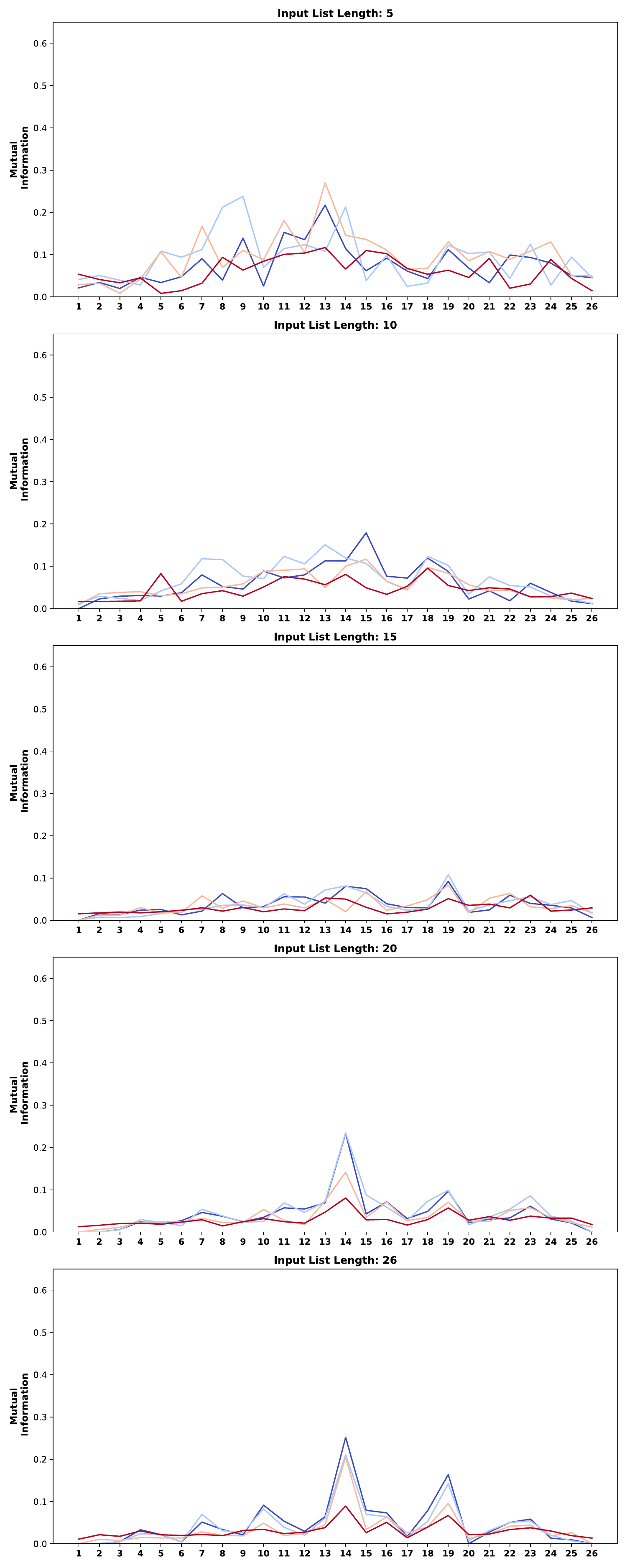}

}

\subcaption{claude-instant-1.2}

\end{figure}%

\end{minipage}%

\caption{\label{fig-mi-scores-numbers}The mutual information between the
input and output of the gpt-3.5-turbo and claude-instant-1.2 models. The
mutual information is computed for each number in the list and then
averaged across all numbers.}

\end{figure}

Mutual information for gpt-3.5-turbo shows a clear temperature
correlation at all list lengths (Figure~\ref{fig-mi-scores-letters}). A
negative correlation between list length and mutual information is
observed for gpt-3.5-turbo, especially at low temperatures.
Claude-instant-1.2 has a positive correlation between the maximum of
served mutual information and list length. While there is a weak
negative correlation between list length and mutual information
(Figure~\ref{fig-mutual-regression}), the effect is only on letters that
do not display mutual information between input identity and position.
Therefore, while we do not observe temperature dependence at low list
lengths, as list length increases, several letters gain a temperature
dependence for mutual information. These effects are invariant to the
object's class (letter or number), with only the specific identity of
the object changing the magnitude of the mutual information
(Figure~\ref{fig-mi-scores-numbers}).

\subsection{Guard Rails With and Without a Sample
Step}\label{guard-rails-with-and-without-a-sample-step}

Guard rails are methods to ensure a LLM returns output in a requested
format, which is useful for quantifying the results of our study. When
sampling for experimental purposes we used a sample step between the
sampling query and the guard rail. To verify our methodology we included
a direct guard rail method for study, where we do not separate the two
steps (Figure~\ref{fig-graphical-abstract}).

The largest change affected by a sample step is a significant reduction
in the primacy bias (Figure~\ref{fig-guard-rails}), with all conditions
displaying a minimum of a 37\% reduction in primacy bias up to 81\%. The
direct method helped ensure correspondence between the input and output
list, although this only proved to be a minor problem at higher
temperatures and intermediate list lengths for the sample step.
Following instructions by responding with the correct number of objects
from the list was greatly increased across the board via the
introduction of a sample step. These effects were seen similarly in
letter selection, with slight differences in effect magnitude
(Figure~\ref{fig-gpt35-full-direct-bias}).

Claude-instant-1.2 was more robust against the application of guard
rails (Figure~\ref{fig-claude-full-direct-bias}), with the largest
change being in primacy, as the direct primacy values were similar to
gpt-3.5-turbo despite claude-1.2-turbo having significantly lower
incidence of primacy initially. There is a weaker effect on the correct
count metric for numbers, with high list lengths fully ablating the
effect of the direct guard rail. For letters the effect on correct count
was negligible.

When considering the mutual information, we see that the direct
application of guard rails causes a large spike for all studied
conditions (Figure~\ref{fig-mi-direct-linear-regression}); to be
expected given the high primacy of direct guard rail applications
reducing the number of possible positions
(Figure~\ref{fig-guard-rails}). The effect of temperature on mutual
information is reduced for claude-instant-1.2 as compared to
gpt-3.5-turbo, with an extremely chaotic profile for input identities
(Figure~\ref{fig-mi-scores-letters-direct}). The effect of list length
on mutual information is consistently negative, with larger list lengths
having lower mutual information across models and temperatures
(Figure~\ref{fig-mi-scores-numbers-direct}). That said, there does
appear to be a lower limit to the mutual information, dependent on the
object identity and model.

\section{Discussion}\label{discussion}

\subsection{Comparisons Between
Models}\label{comparisons-between-models}

For the task of list selection, claude-instant-1.2 outperforms
gpt-3.5-turbo-0613 in terms of low bias (Figure~\ref{fig-bias-cot}) and
robustness under the application of a guard rail
(Figure~\ref{fig-claude-full-direct-bias}). The mutual information
between position and object identity highlights this disparity, where,
except at high temperatures, claude-instant-1.2 has lower mutual
information across the board (Figure~\ref{fig-mi-scores-letters}).

There is little public information about the internal structure of the
two models, our results highlight an important consideration: even
similarly performing models may have wildly divergent bias profiles. For
example, gpt-3.5-turbo-0613 shows a positional bias structure,
Figure~\ref{fig-letter-position-gpt}, similar to the that reported by
Wang et al. (\citeproc{ref-wangPrimacyEffectChatGPT2023}{2023}), while
the positional bias structure of claude-instant-1.2 is inverted, with
the central positions being the most common at low list lengths. This is
a trend that, as list length increases, morphs into a favoring of the
third position and a disfavoring of the first two and last positions.
This result is at odds with the predicted behavior derived from Liu et
al. (\citeproc{ref-liuLostMiddleHow2023}{2023}), which would point to a
favoring of information in the start and end of a context. When dealing
with bias, generalizing behavior is risky and may not be appropriate
even for similar tasks and models.

\begin{figure}

\centering{

\includegraphics{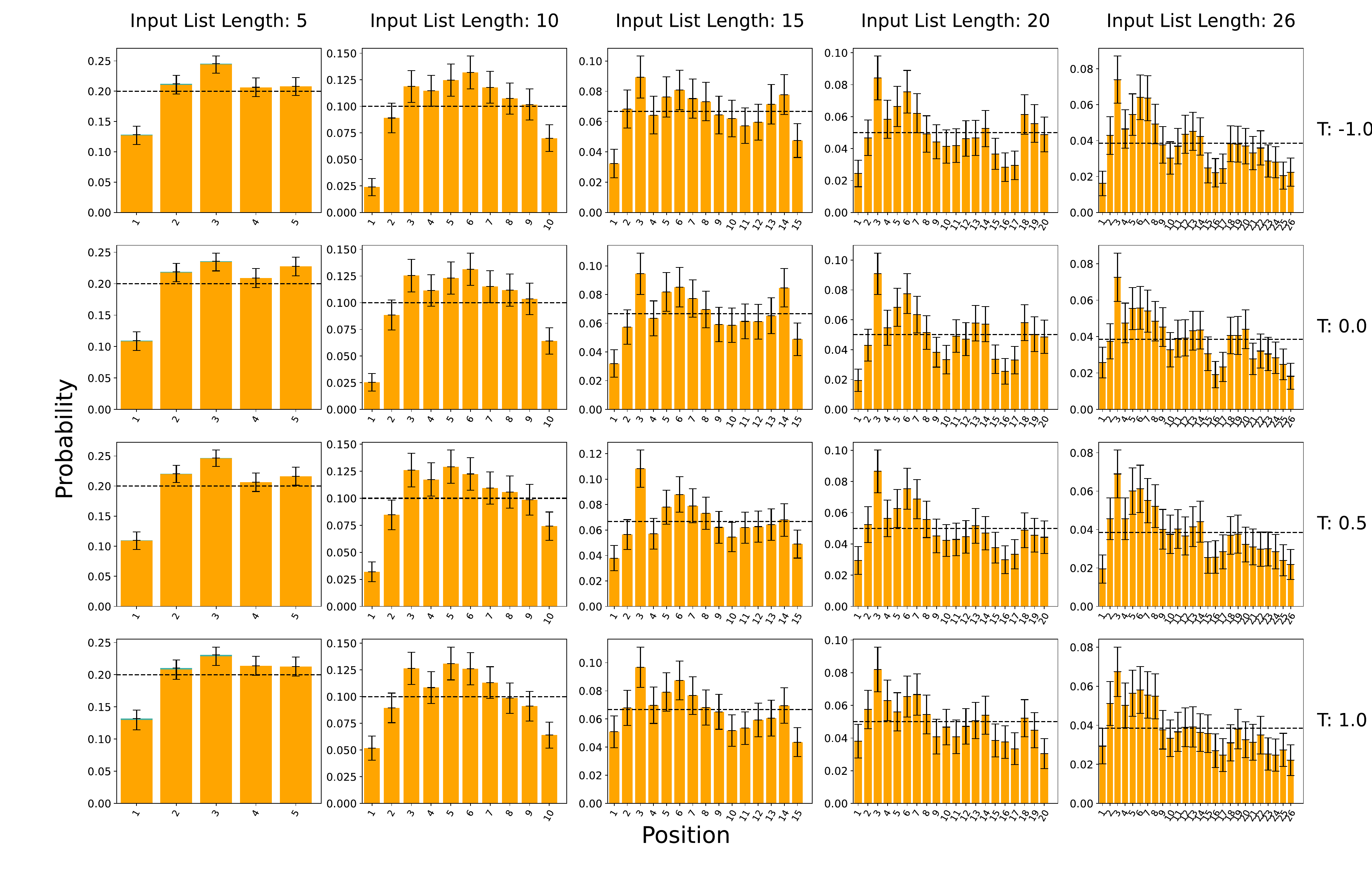}

}

\caption{\label{fig-number-position-claude}Using claude-instant-1.2 the
scaled probability that a position of a number will be selected given a
temperature and original list length was computed. The dotted black line
is the expected probability given random sampling of a uniform
distribution for a list length. Orange bars are the probability that a
position without primacy bias will be selected, while blue represents
the probability that a value with primacy bias will be selected. Error
bars are standard error from 3000 bootstrap replicates.}

\end{figure}

\begin{figure}

\centering{

\includegraphics{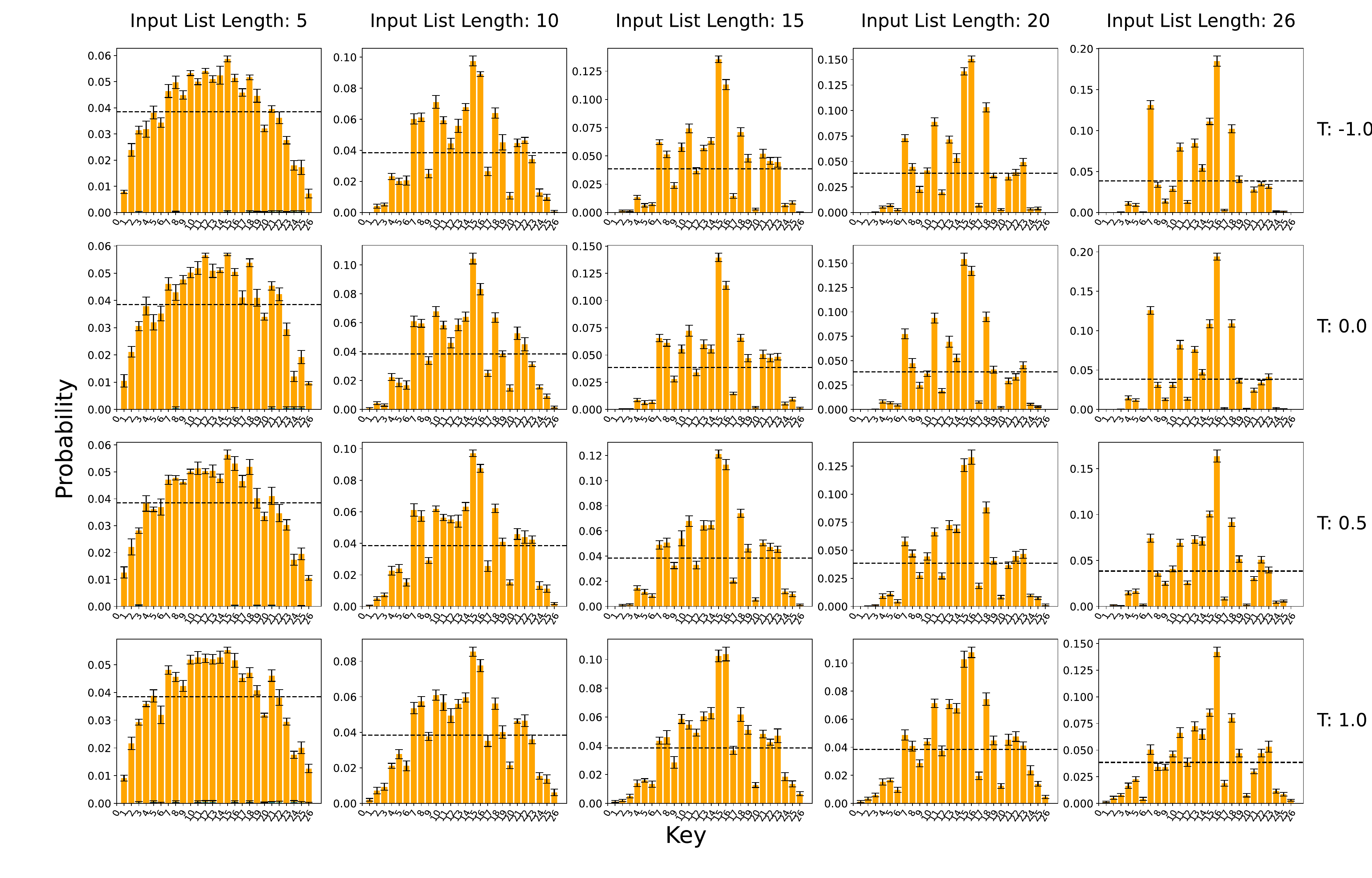}

}

\caption{\label{fig-number-input-probability-claude}Using
claude-instant-1.2 the scaled probability that a number will be selected
given a temperature and original list length. The dotted black line is
the expected probability given random sampling of a uniform distribution
for a list length. Orange bars are the probability that a position
without primacy bias will be selected, while blue represents the
probability that a value with primacy bias will be selected. Error bars
are standard error from 3000 bootstrap replicates.}

\end{figure}

Despite claude-instant-1.2 having significantly lowered incidence of
primacy bias, it has its own bias which is relatively insensitive to
temperature (Figure~\ref{fig-number-position-claude}). Moreover there
are a number of conditions where certain objects are not selected at
all. Furthermore there are a variety of biases we may not be
appropriately detecting. One such candidate would be a bias towards the
selection of the average value of a given list of numbers
(Figure~\ref{fig-number-input-probability-claude}). Unfortunately we did
not study this in enough detail to comment extensively, but this is
indicative of emergent bias patterns that blindly trusting any given LLM
may blindside us with.

\subsection{Guard Rails and Cognitive
Load}\label{guard-rails-and-cognitive-load}

\begin{figure}

\begin{minipage}{0.33\linewidth}
\includegraphics{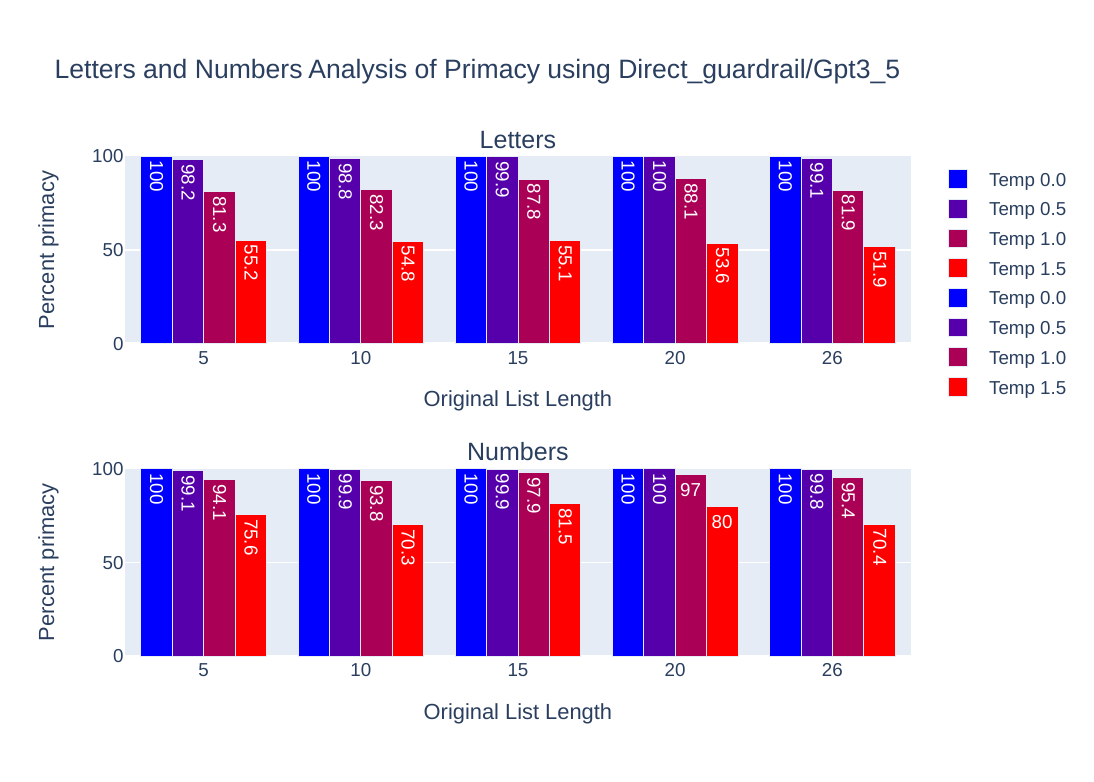}\end{minipage}%
\begin{minipage}{0.33\linewidth}
\includegraphics{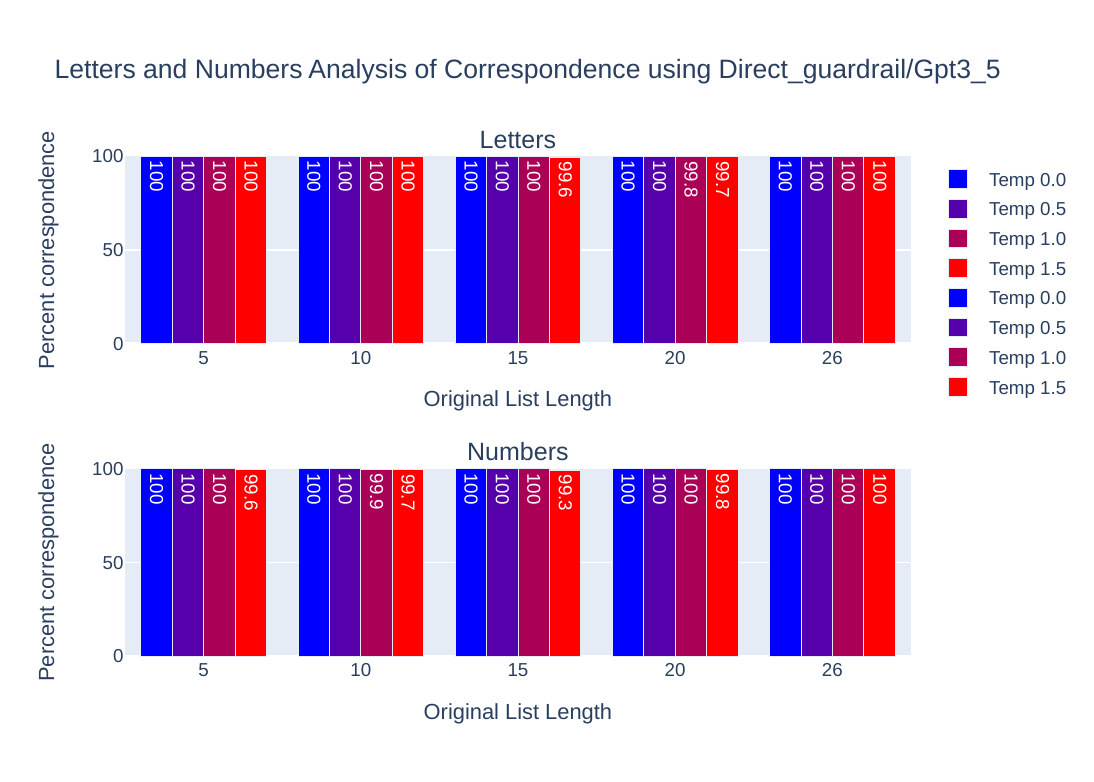}\end{minipage}%
\begin{minipage}{0.33\linewidth}
\includegraphics{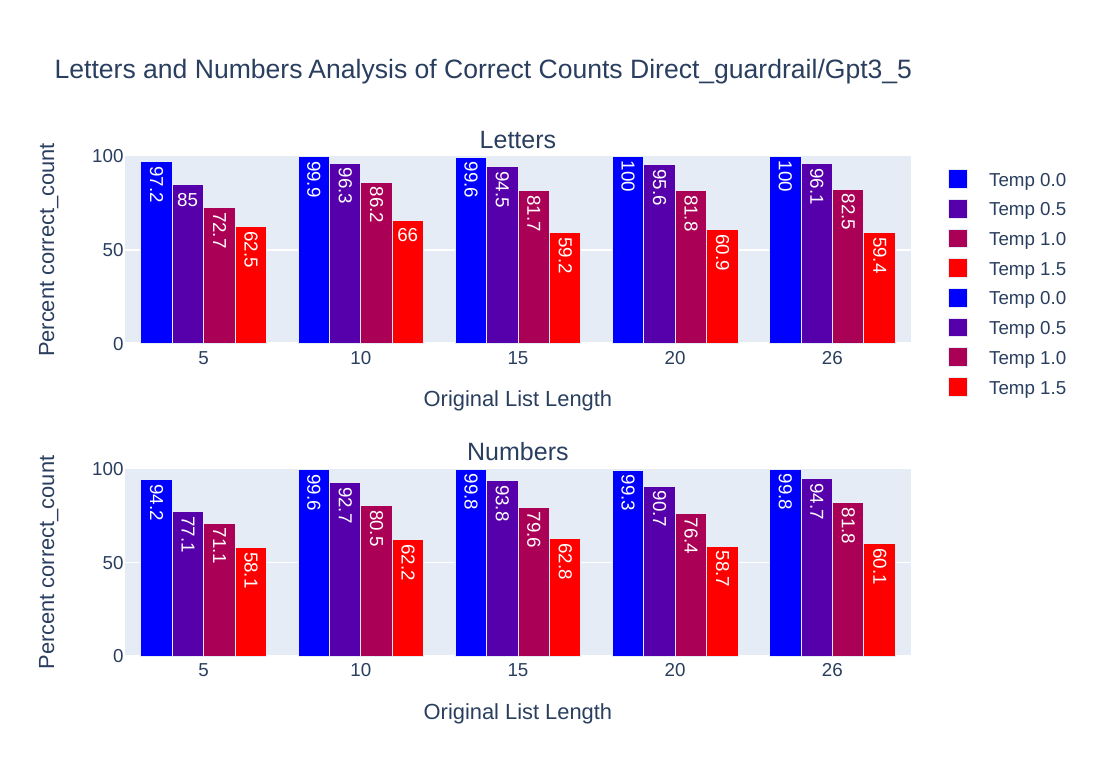}\end{minipage}%

\caption{\label{fig-gpt35-full-direct-bias}The full results of the
gpt-3.5-turbo model without a spacing step marked on primacy,
correspondence, and correct counts. This is compared for all list
lengths analyzed and all temperatures.}

\end{figure}

\begin{figure}

\begin{minipage}{0.33\linewidth}
\includegraphics{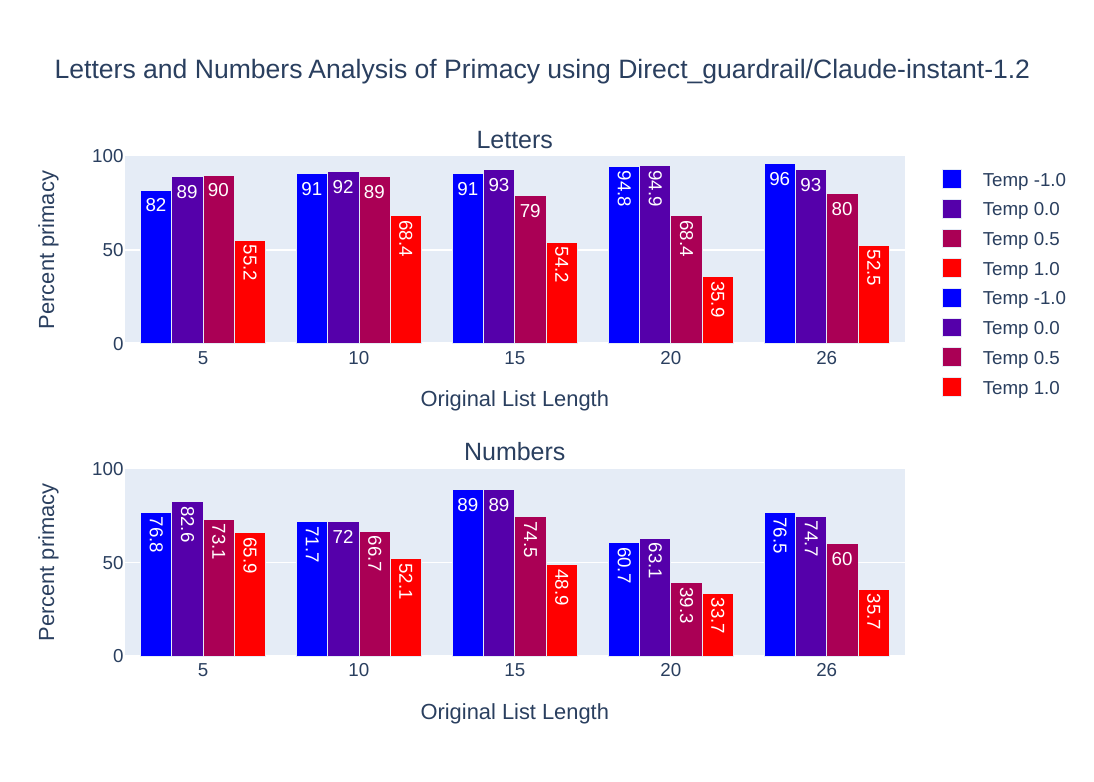}\end{minipage}%
\begin{minipage}{0.33\linewidth}
\includegraphics{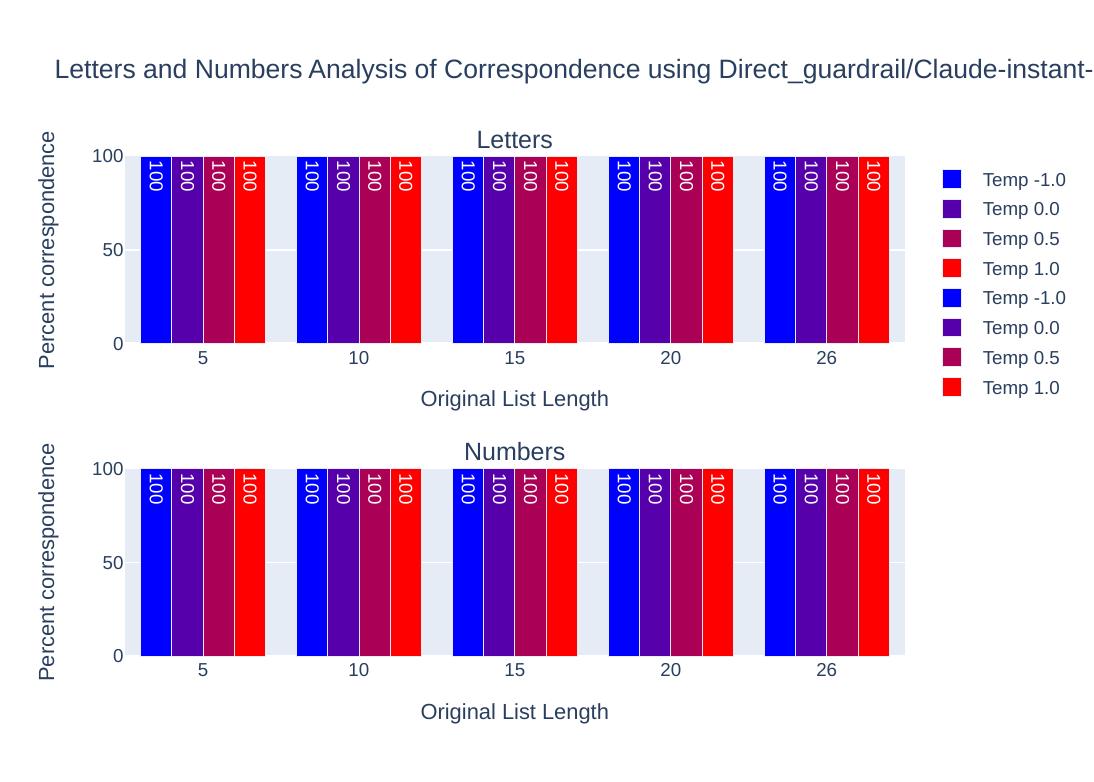}\end{minipage}%
\begin{minipage}{0.33\linewidth}
\includegraphics{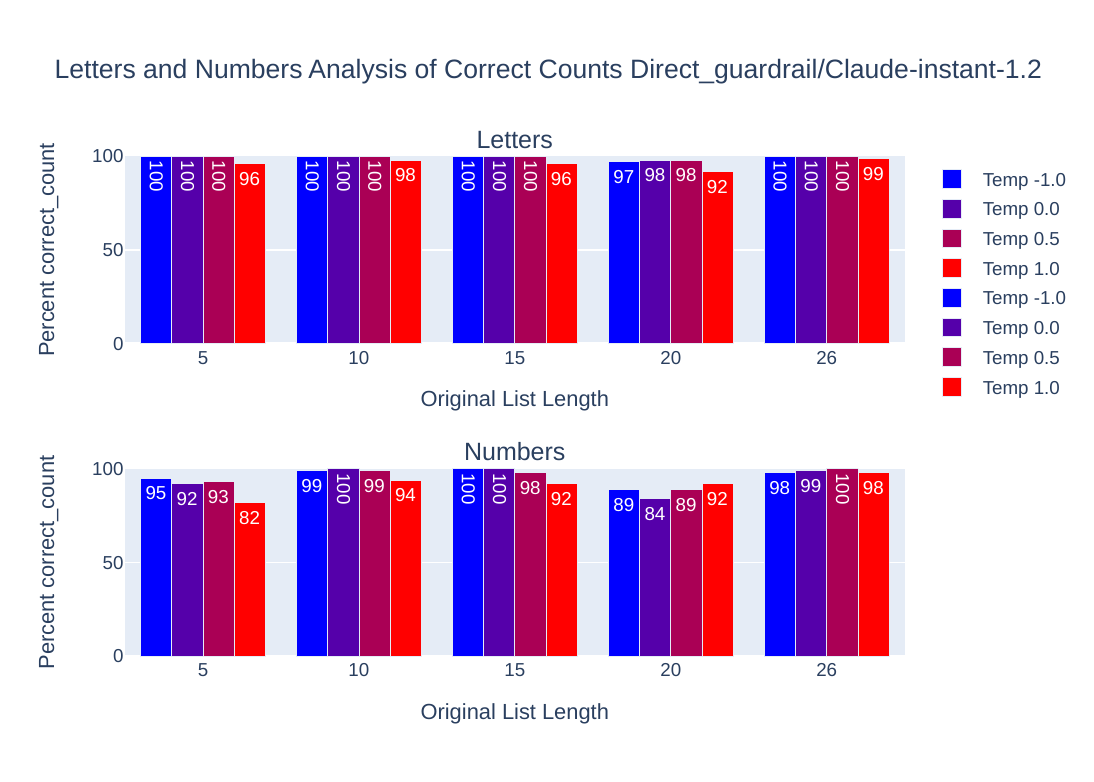}\end{minipage}%

\caption{\label{fig-claude-full-direct-bias}The full results of the
claude model without a spacing step marked on primacy, correspondence,
and correct counts. This is compared for all list lengths analyzed and
all temperatures. For primacy, all empty values are 0 as there was no
occurrence of primacy.}

\end{figure}

\begin{figure}

\begin{minipage}{0.50\linewidth}

\begin{figure}[H]

{\centering \includegraphics{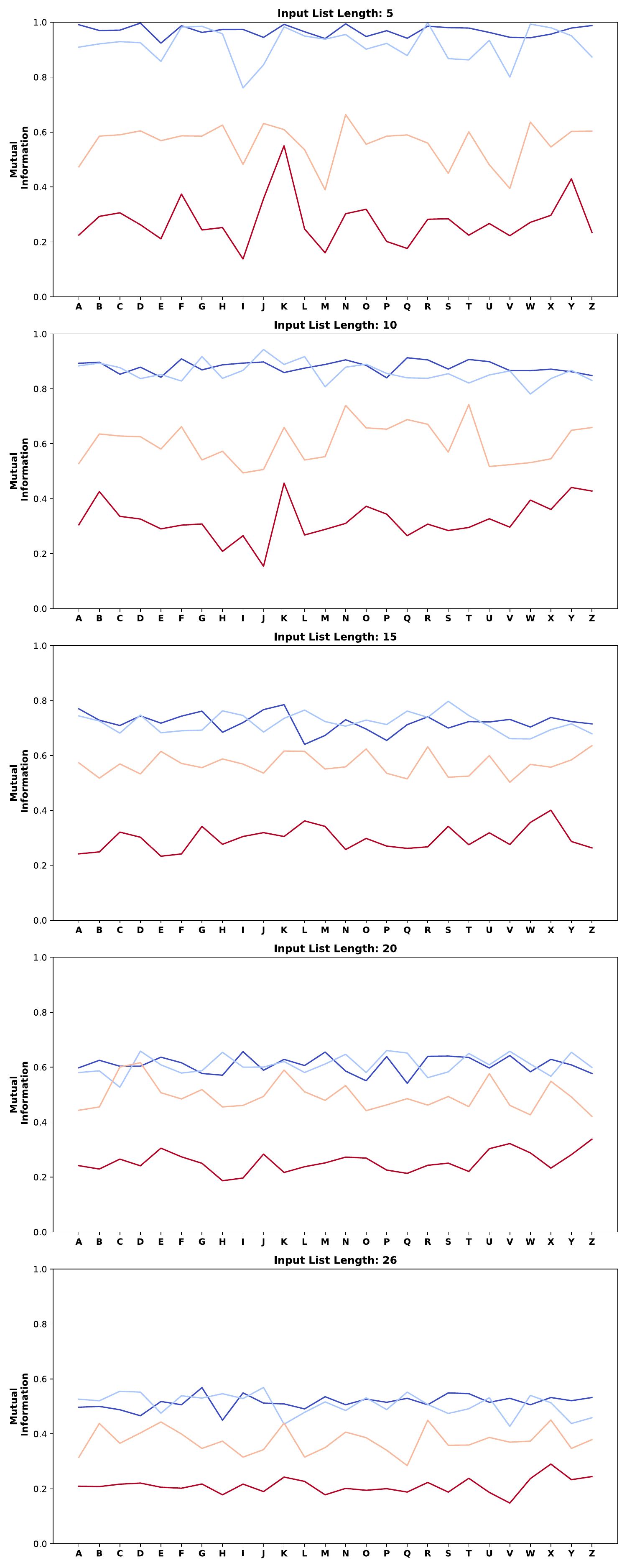}

}

\subcaption{gpt-3.5-instant}

\end{figure}%

\end{minipage}%
\begin{minipage}{0.50\linewidth}

\begin{figure}[H]

{\centering \includegraphics{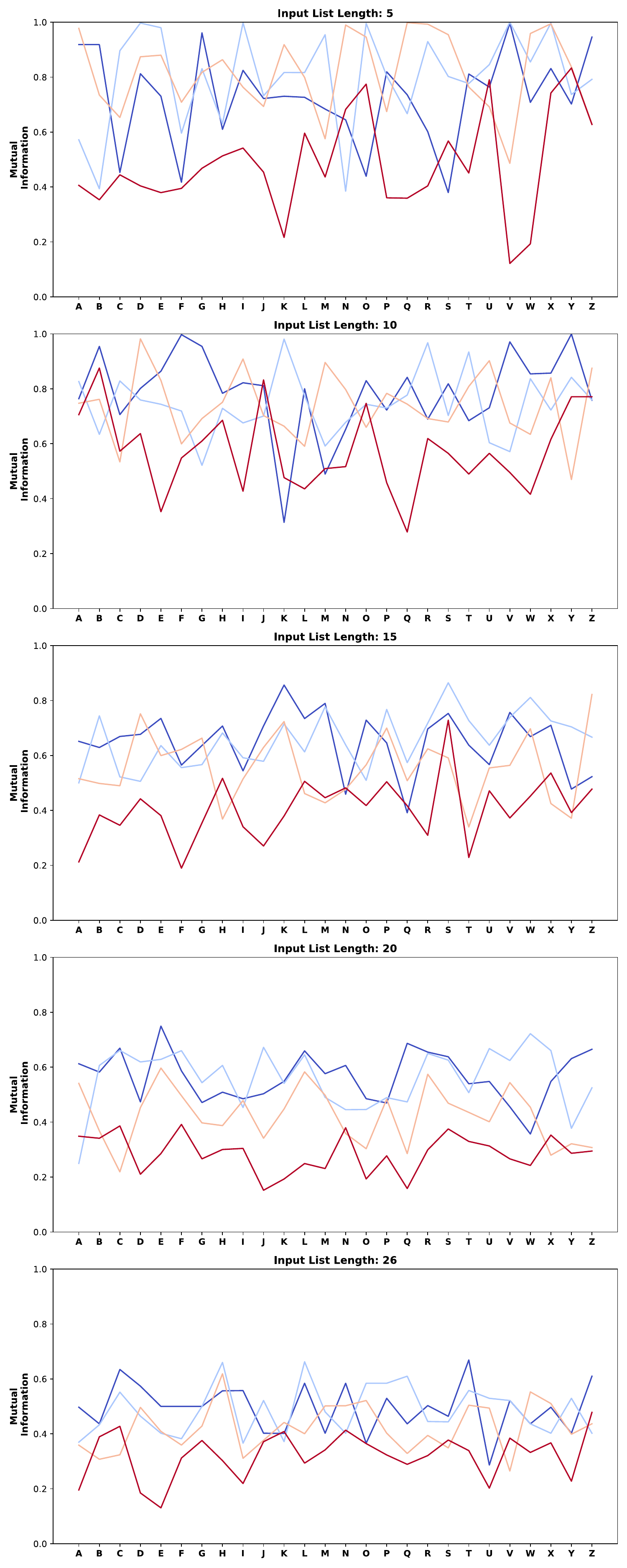}

}

\subcaption{claude-instant-1.2}

\end{figure}%

\end{minipage}%

\caption{\label{fig-mi-scores-letters-direct}The mutual information
between the input and output of the direct guard rails method for
gpt-3.5-turbo and claude-instant-1.2 models. The mutual information is
computed for each letter in the list and then averaged across all
letters.}

\end{figure}

\begin{figure}

\begin{minipage}{0.50\linewidth}

\begin{figure}[H]

{\centering \includegraphics{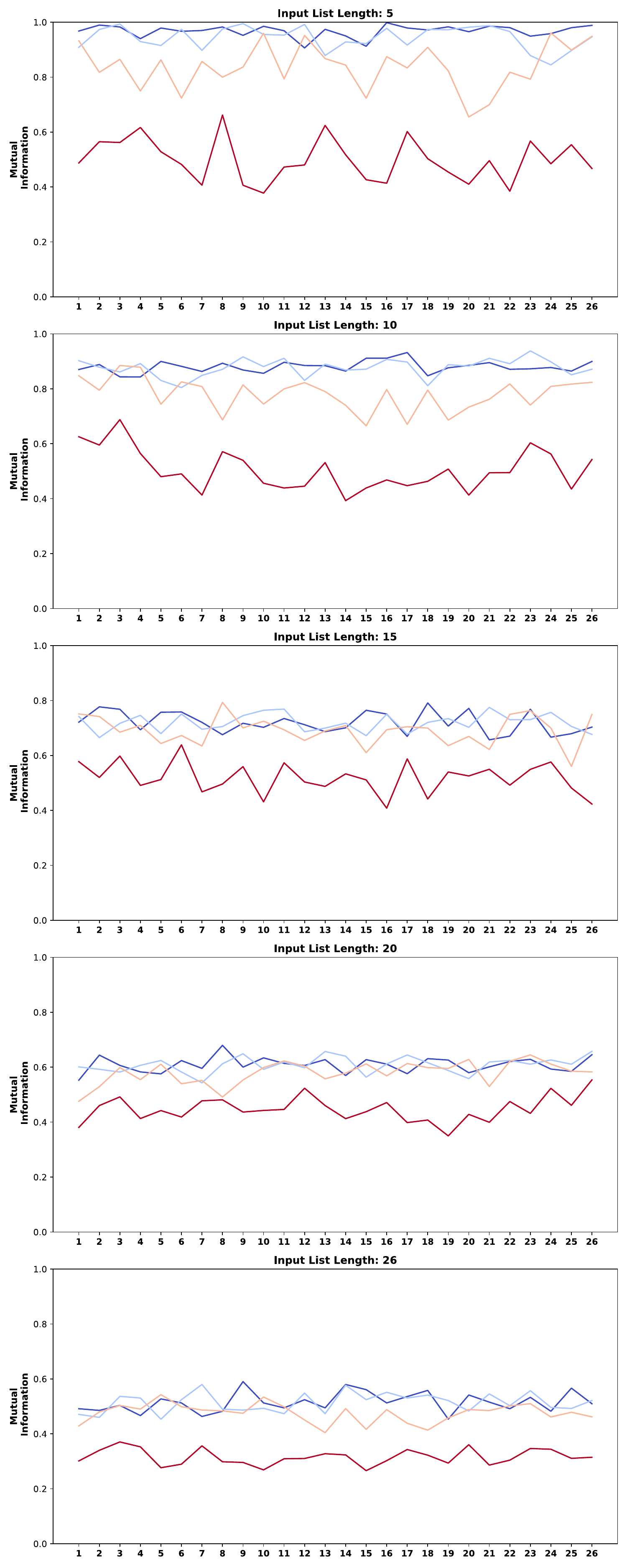}

}

\subcaption{gpt-3.5-instant}

\end{figure}%

\end{minipage}%
\begin{minipage}{0.50\linewidth}

\begin{figure}[H]

{\centering \includegraphics{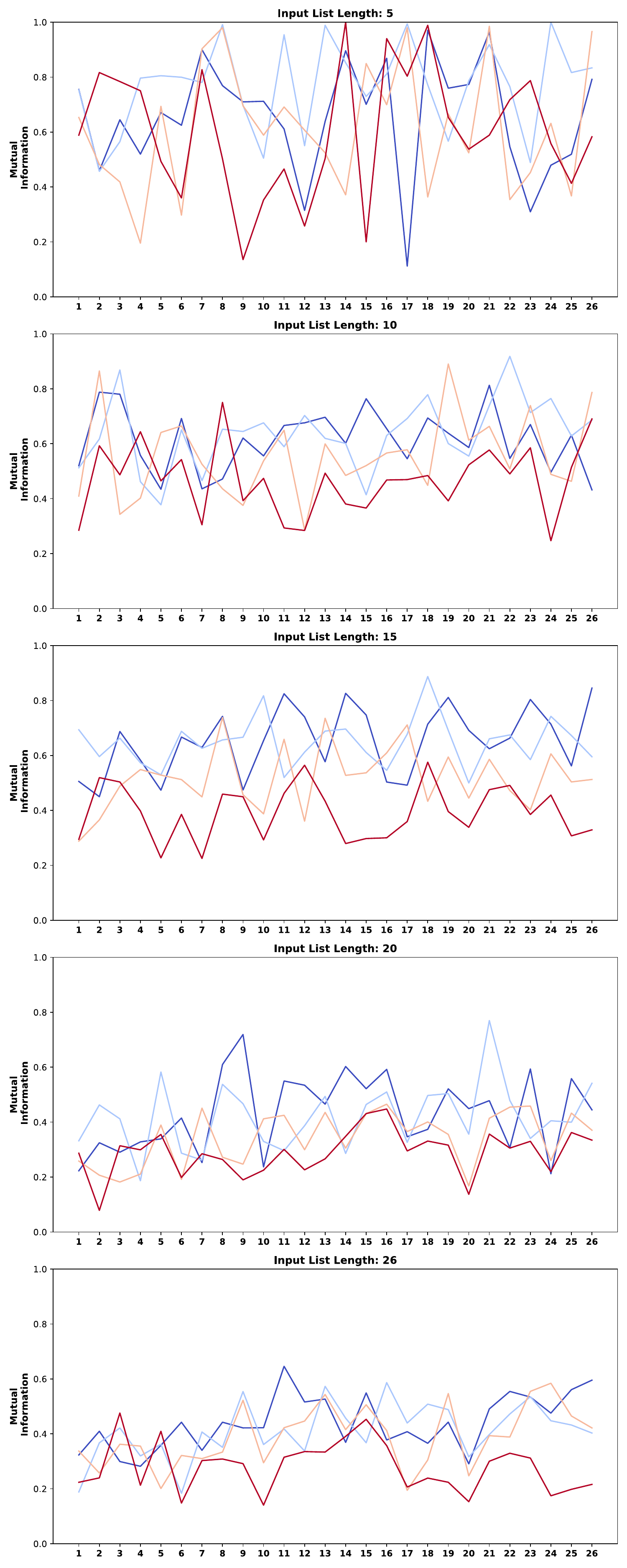}

}

\subcaption{claude-instant-1.2}

\end{figure}%

\end{minipage}%

\caption{\label{fig-mi-scores-numbers-direct}The mutual information
between the input and output of the direct guard rails method for
gpt-3.5-turbo and claude-instant-1.2 models. The mutual information is
computed for each number in the list and then averaged across all
numbers.}

\end{figure}

Guard rails are potent methods for assuring the output of a LLM complies
with user defined parameters (Rebedea et al.
(\citeproc{ref-rebedeaNeMoGuardrailsToolkit2023}{2023})). While this
technology constitutes a great tool for many situations, the effect of
these guard rails on LLM decision-making has not been fully examined
(Shankar et al.
(\citeproc{ref-shankarSPADESynthesizingAssertions2024}{2024}),
Abdelkader et al.
(\citeproc{ref-abdelkaderMLOnRailsSafeguardingMachine2024}{2024})). The
introduction of a guard rail resulted in a model agnostic gain in
primacy bias (Figure~\ref{fig-gpt35-full-direct-bias},
Figure~\ref{fig-claude-full-direct-bias}), while decreasing the correct
count, a metric of instruction-following behavior. Without guard rails,
only gpt-3.5-turbo-0613, while selecting number objects, showed any
issues with instruction-following (Figure~\ref{fig-gpt35-full-bias}).
When guard rails are applied, both letters and numbers experienced a
similar magnitude increase in primacy and failure to follow
instructions. There is model specificity to guard rail induced bias, as
while there was still some effect on instruction adherence, it was
qualitatively different from claude-instant-1.2
(Figure~\ref{fig-claude-full-direct-bias}). Furthermore, the direct
guard rail had a significant effect on the mutual information between
position and object identity for gpt-3.5-turbo-0613 and
claude-instant-1.2 both (Figure~\ref{fig-mi-scores-letters-direct},
Figure~\ref{fig-mi-scores-numbers-direct}), with the averages for both
models being significantly higher in the direct guard rail application
than the two-step method (Figure~\ref{fig-mi-direct-linear-regression}).

Asking a human to select three objects from a list and write them down
is a trivial task. Asking them to format the response in a JSON, a task
guard rails have been developed to tackle in LLMs, will result in a much
more significant expenditure of effort. To handle the extra cognitive
load, humans will compensate by allowing bias to reduce the effort (de
Jong (\citeproc{ref-dejongCognitiveLoadTheory2010}{2010})). In the same
way, LLMs follow similar trends by increasing primacy bias
(Figure~\ref{fig-guard-rails}) and the mutual information between
objects and positions (Figure~\ref{fig-mi-scores-letters},
Figure~\ref{fig-mi-scores-letters-direct}). The relative semantic
complexity of the task may indicate that LLMs experience a form of
cognitive load that causes a compensatory raise in bias when a guard
rail is applied. This would align with a novel prompt injection
technique termed ``cognitive overload'' (Xu et al.
(\citeproc{ref-xuCognitiveOverloadJailbreaking2023}{2023})). In our
study, the LLM compensates by increasing the primacy bias and reducing
instruction adherence. This compensation implies that the use of any
sort of structural assurance in output for a LLM will introduce some
form of compensation. While in our case this involved the adherence to
instruction and uncreative list selection, the exact nature may vary
greatly depending on the model and task just as a human might (Paas and
van Merriënboer
(\citeproc{ref-paasCognitiveLoadTheoryMethods2020}{2020})). Therefore
the distribution of outputs of a cognitively taxing structure for a LLM
should be empirically analyzed given our current theoretical grounding.

The proposed theory of LLM cognition implies an interesting future
direction of research, where the cognitive load of a LLM is measured in
order to determine the appropriate compensation for a given task. If
done properly models for predicting the effect of cognitive load can be
made, allowing for rapid deployment of complex structures in LLMs.

\section{Limitations and Future
Directions}\label{limitations-and-future-directions}

The scope of this study focused exclusively on the patterns inherent in
selecting naive numbers and letters from lists. Conceptually this is
identical to the usage of more complex objects such as words, phrases,
or randomized tokens. In common use these effects may show great
mitigation or enhancement due to the semantic content of the objects
being selected. This connection between the semantic content of the
objects and the selection patterns is a proposed area of future
research. Future research should delve into how LLMs bias more
meaningful use cases, such as real-world agentic systems and code
generation.

We only selected three objects from the list at a time, other works
should expand on this to see if the patterns hold for larger or smaller
selection pools. We also noted some patterns in the objects that were
selected together, this is an area of future research and should be
expanded to help understand LLM list construction tasks.

The prompt we used was a simple request to ``Please select 3 of the
following:''; our goal was to minimize prompt bias while still retaining
the spirit of directed object selection. There are numerous options for
prompt construction such as including ``randomly'' or even attempting to
instruct the LLM to select things in an unbiased manner. Although from
our preliminary work, we believe this will just lead to new probability
distributions rather than ablating them fully.

Of note, this is not accounting for the possibility that the presence of
other objects in a list may affect the probability of an object being
selected:
\(P(\ell_{i} \in \mathbb{x}_{s} | \ell_{j} \in \mathbb{x}_{s}) \neq P(\ell_{i} \in \mathbb{x}_{s})\).
Looking into how the presence of other objects in the list affect the
selection probabilities will allow for better list debiasing methods.

The models used are closed source, using open source models such as
Pythia would allow for close inspection of bias evolution as a function
of training and size (Biderman et al.
(\citeproc{ref-bidermanPythiaSuiteAnalyzing2023}{2023})). This would
also allow us to more accurately analyze the modulation in bias as the
functionality of a model changes to see if phase transitions in ability
are associated with bias modulation (Chang
(\citeproc{ref-changSimpleExplanationPhase2023}{2023})).

\section{Conclusion}\label{conclusion}

\begin{figure}

\begin{minipage}{0.50\linewidth}
\includegraphics{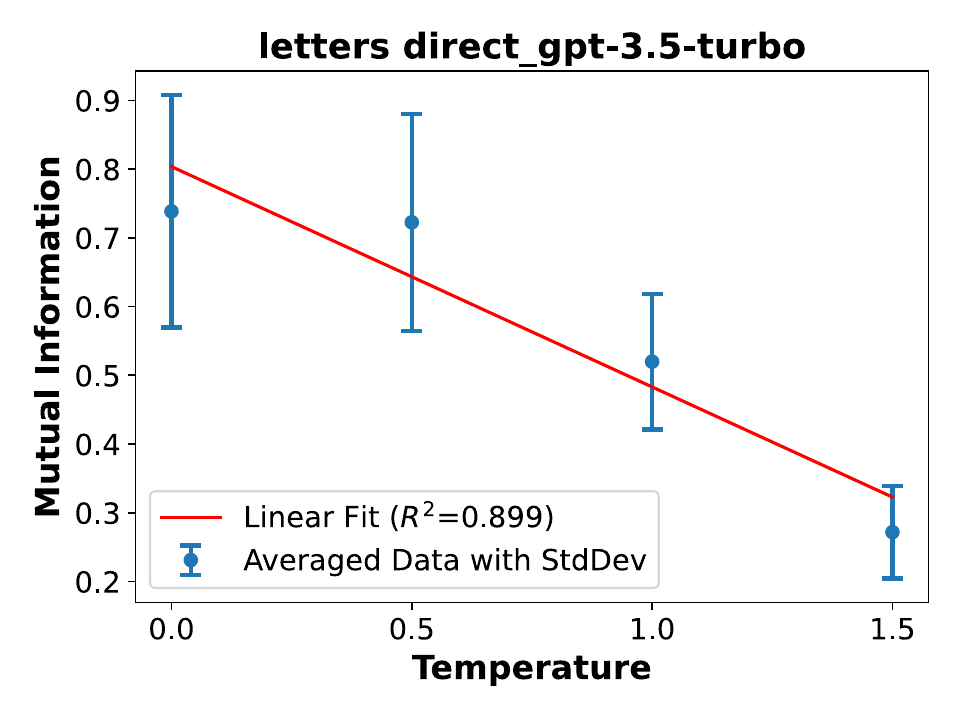}\end{minipage}%
\begin{minipage}{0.50\linewidth}
\includegraphics{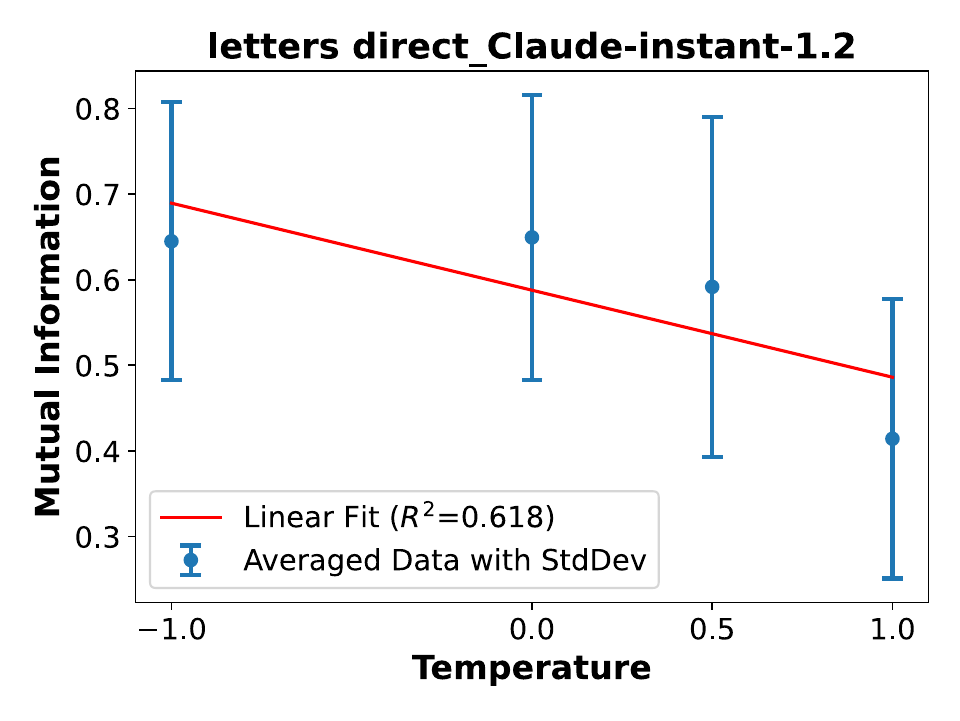}\end{minipage}%
\newline
\begin{minipage}{0.50\linewidth}
\includegraphics{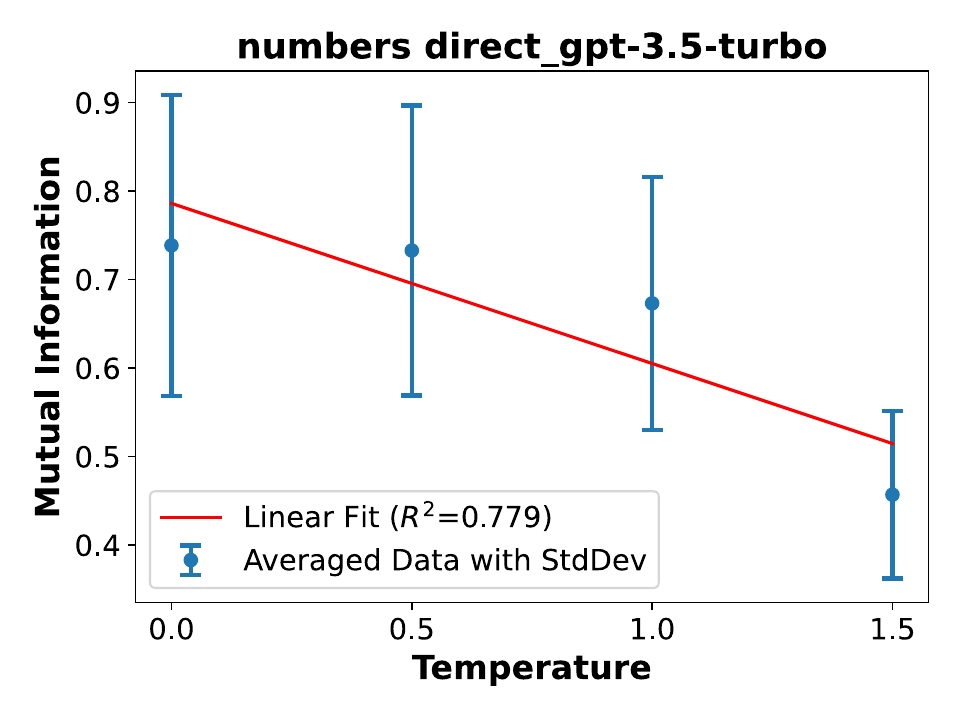}\end{minipage}%
\begin{minipage}{0.50\linewidth}
\includegraphics{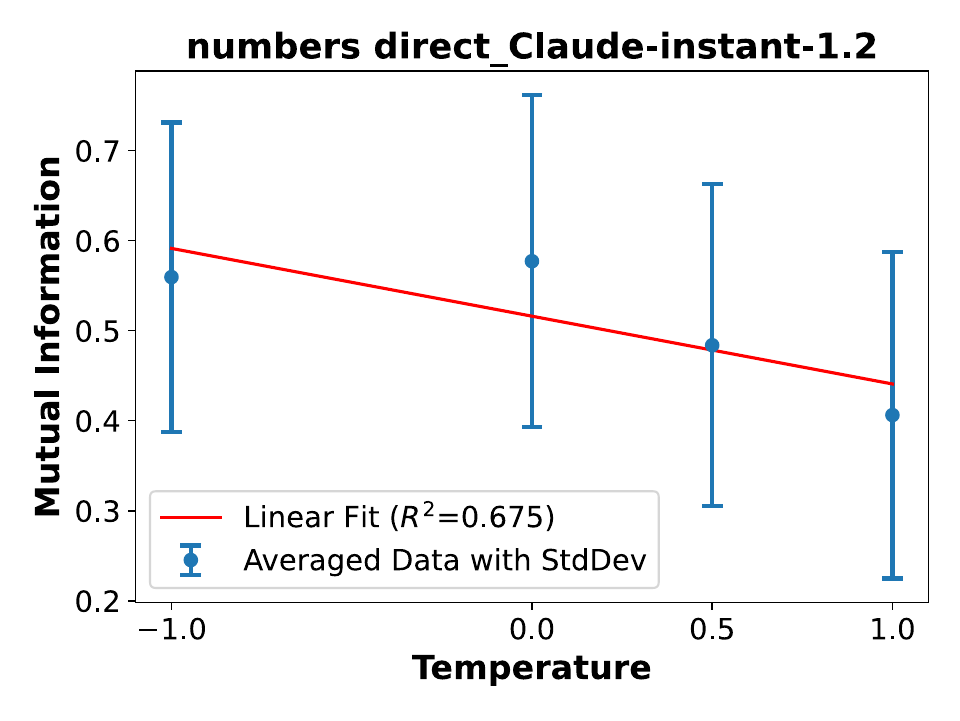}\end{minipage}%

\caption{\label{fig-mi-direct-linear-regression}Average mutual
information between an input and position with standard deviations.
Averages were computed over all list lengths and inputs of a given type.
A linear regression was computed for letter inputs (gpt-3.5-turbo
\(R^2 = 0.899\), Claude-instant-1.2 \(R^2 = 0.618\)) and number inputs
(gpt-3.5-turbo \(R^2 = 0.779\), Claude-instant-1.2 \(R^2 = 0.675\)).}

\end{figure}

\begin{figure}

\begin{minipage}{0.20\linewidth}

\begin{figure}[H]

{\centering \includegraphics{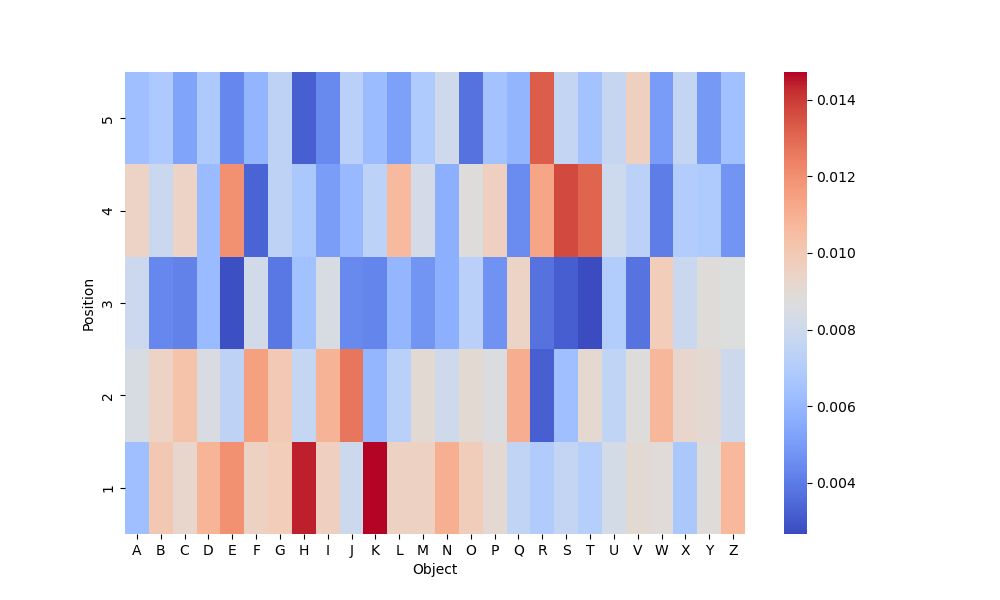}

}

\subcaption{T = 0, Count = 5}

\end{figure}%

\end{minipage}%
\begin{minipage}{0.20\linewidth}

\begin{figure}[H]

{\centering \includegraphics{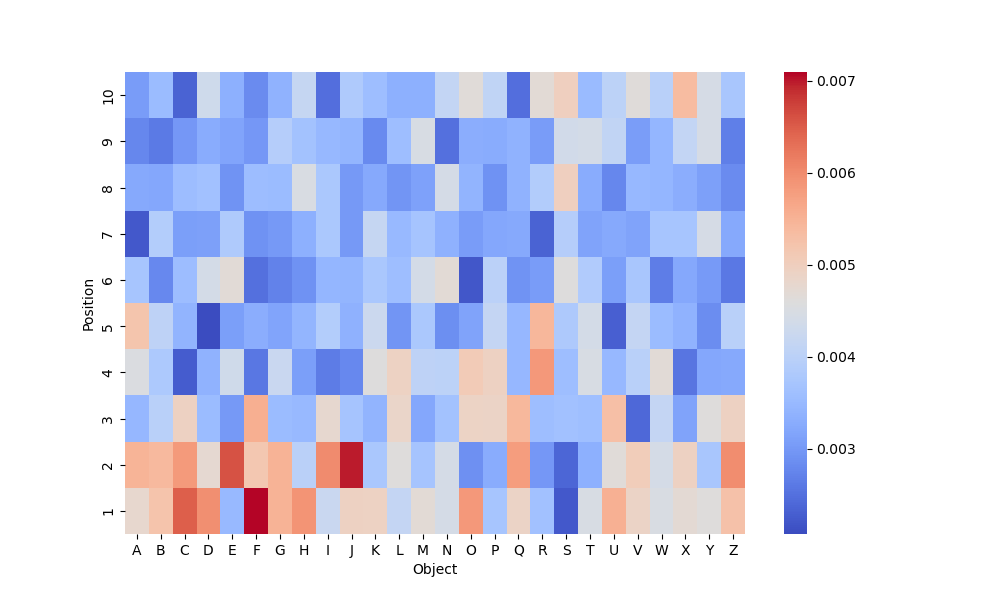}

}

\subcaption{T = 0, Count = 10}

\end{figure}%

\end{minipage}%
\begin{minipage}{0.20\linewidth}

\begin{figure}[H]

{\centering \includegraphics{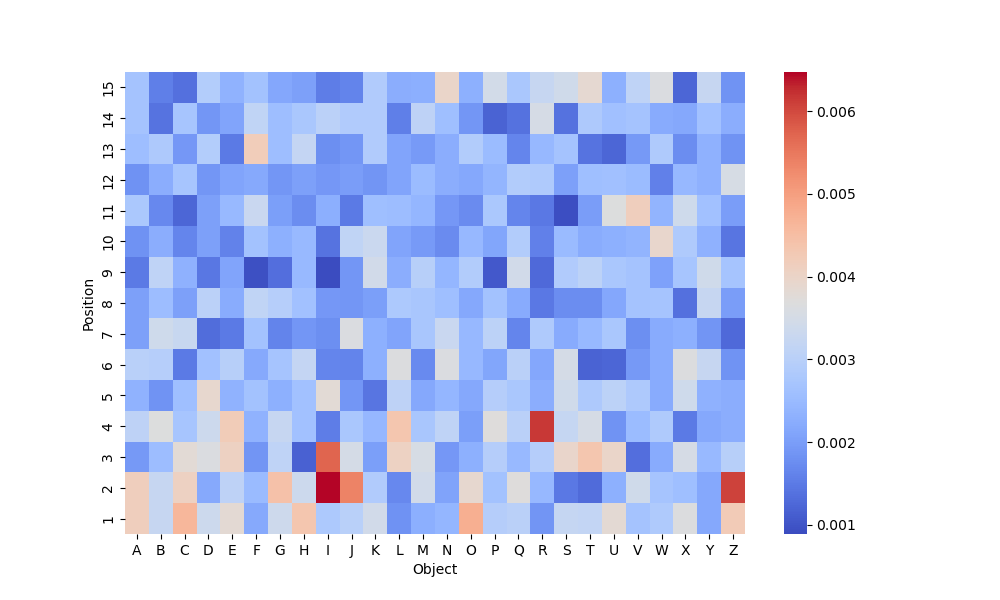}

}

\subcaption{T = 0, Count = 15}

\end{figure}%

\end{minipage}%
\begin{minipage}{0.20\linewidth}

\begin{figure}[H]

{\centering \includegraphics{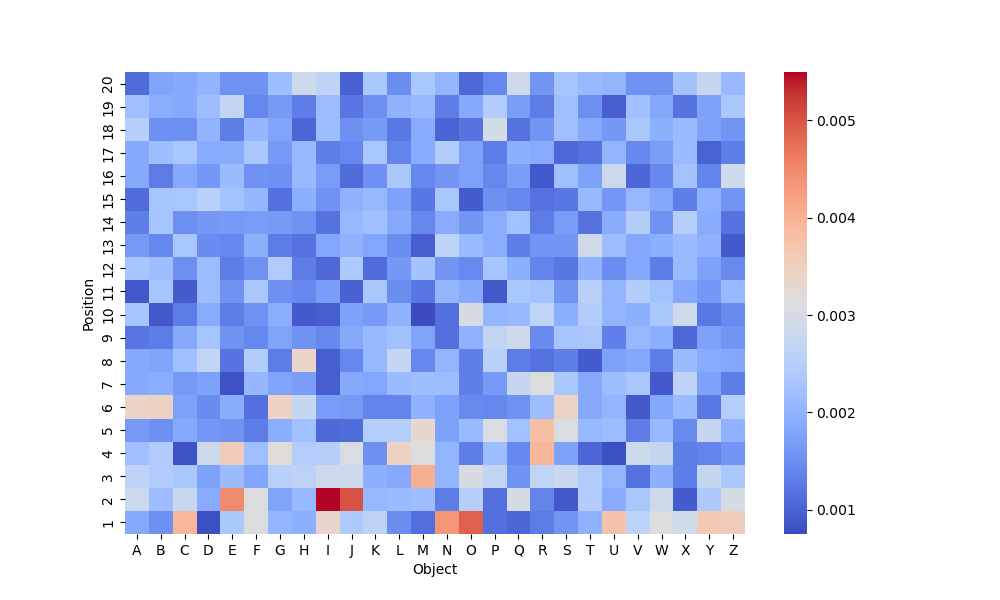}

}

\subcaption{T = 0, Count = 20}

\end{figure}%

\end{minipage}%
\begin{minipage}{0.20\linewidth}

\begin{figure}[H]

{\centering \includegraphics{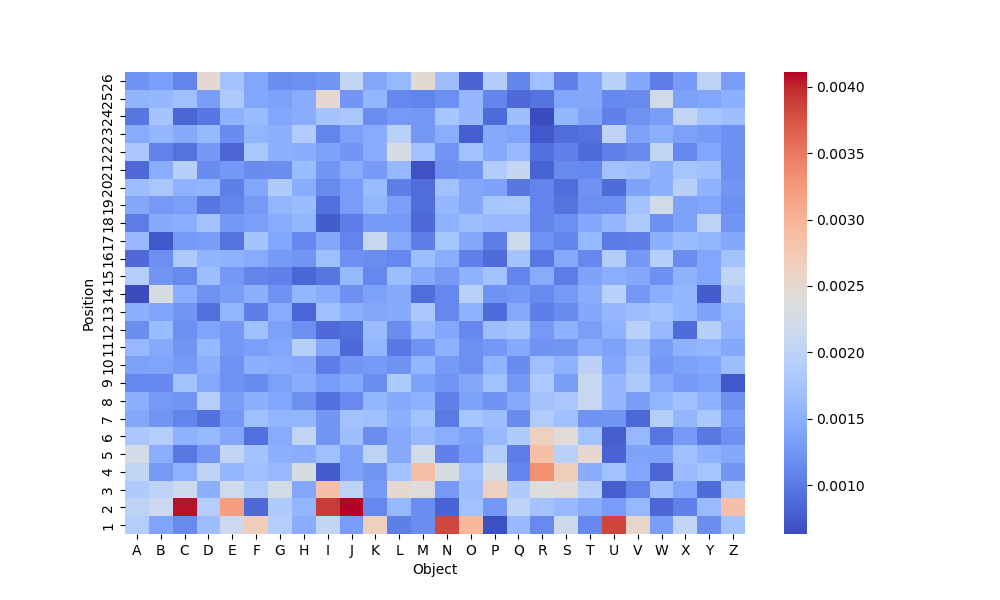}

}

\subcaption{T = 0, Count = 26}

\end{figure}%

\end{minipage}%
\newline
\begin{minipage}{0.20\linewidth}

\begin{figure}[H]

{\centering \includegraphics{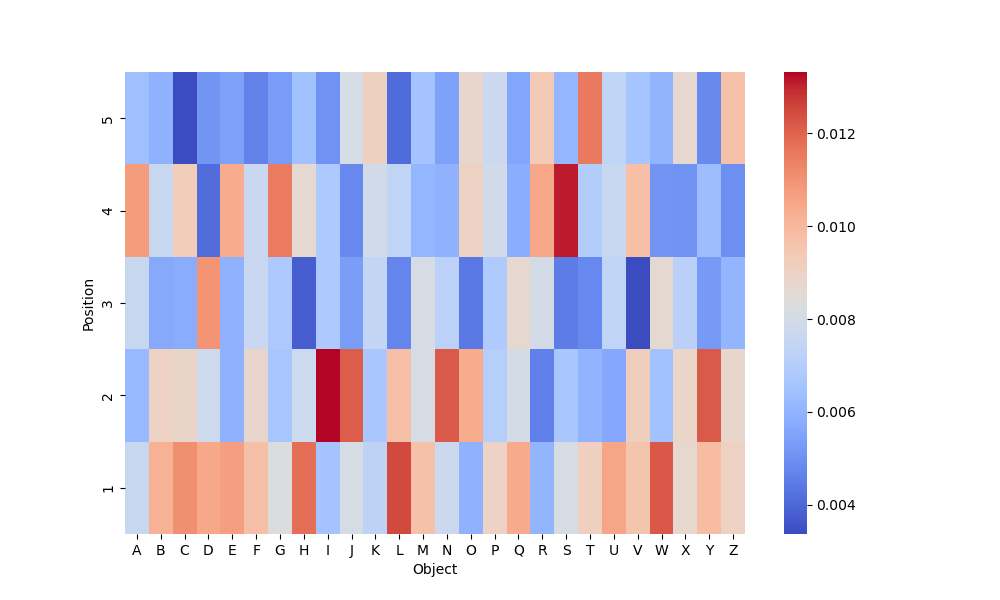}

}

\subcaption{T = 0.5, Count = 5}

\end{figure}%

\end{minipage}%
\begin{minipage}{0.20\linewidth}

\begin{figure}[H]

{\centering \includegraphics{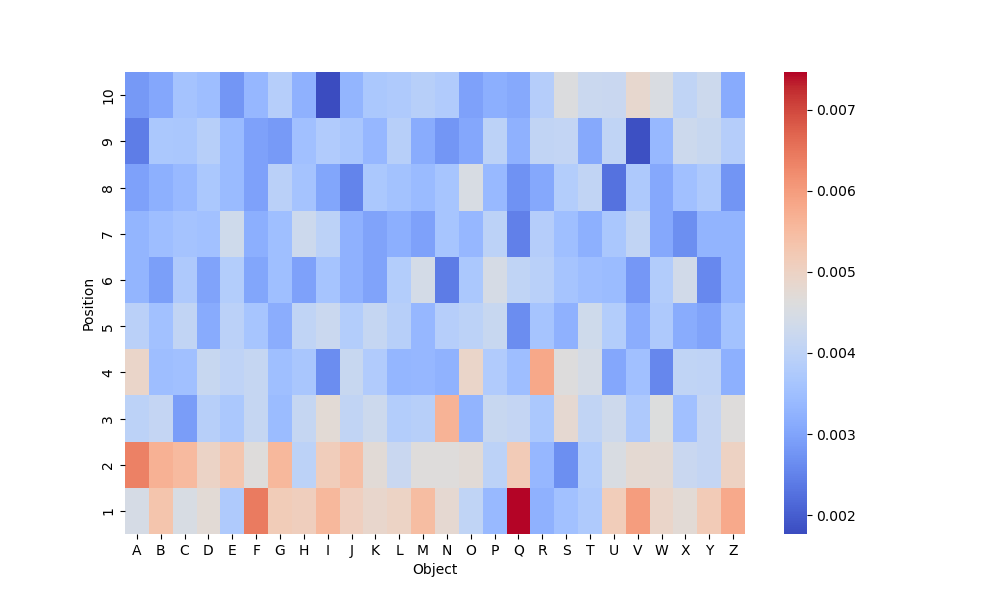}

}

\subcaption{T = 0.5, Count = 10}

\end{figure}%

\end{minipage}%
\begin{minipage}{0.20\linewidth}

\begin{figure}[H]

{\centering \includegraphics{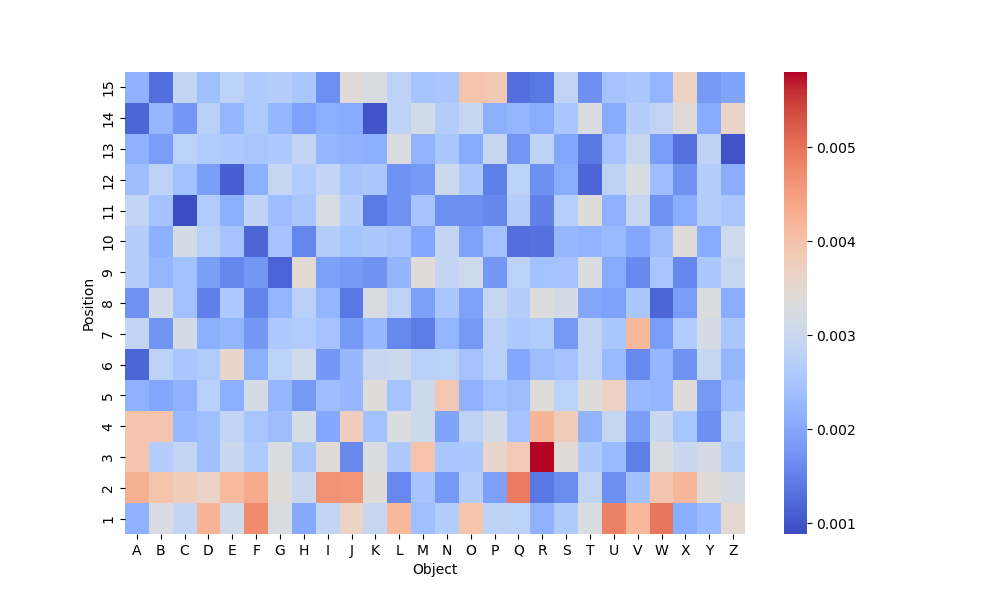}

}

\subcaption{T = 0.5, Count = 15}

\end{figure}%

\end{minipage}%
\begin{minipage}{0.20\linewidth}

\begin{figure}[H]

{\centering \includegraphics{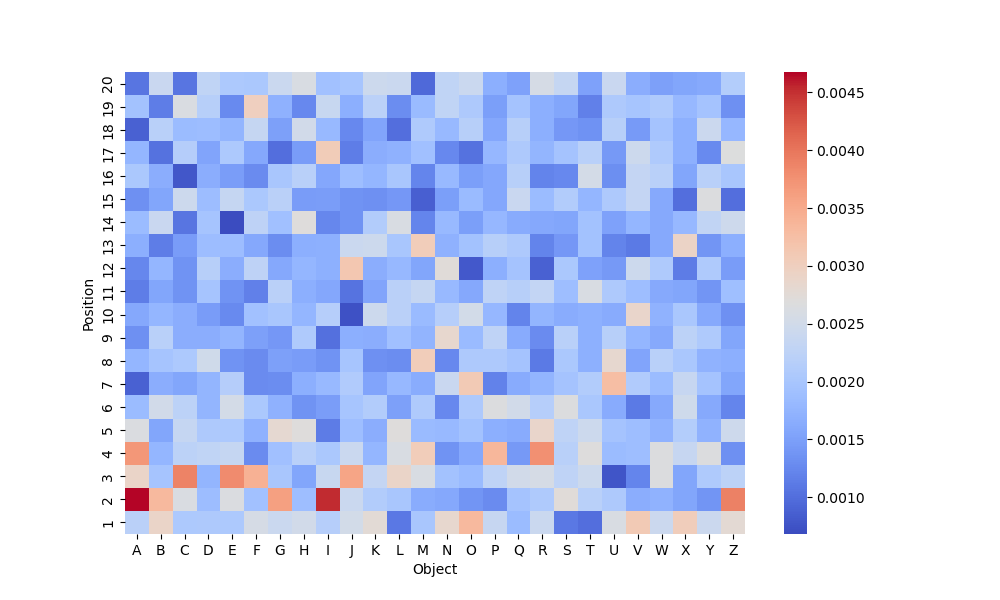}

}

\subcaption{T = 0.5, Count = 20}

\end{figure}%

\end{minipage}%
\begin{minipage}{0.20\linewidth}

\begin{figure}[H]

{\centering \includegraphics{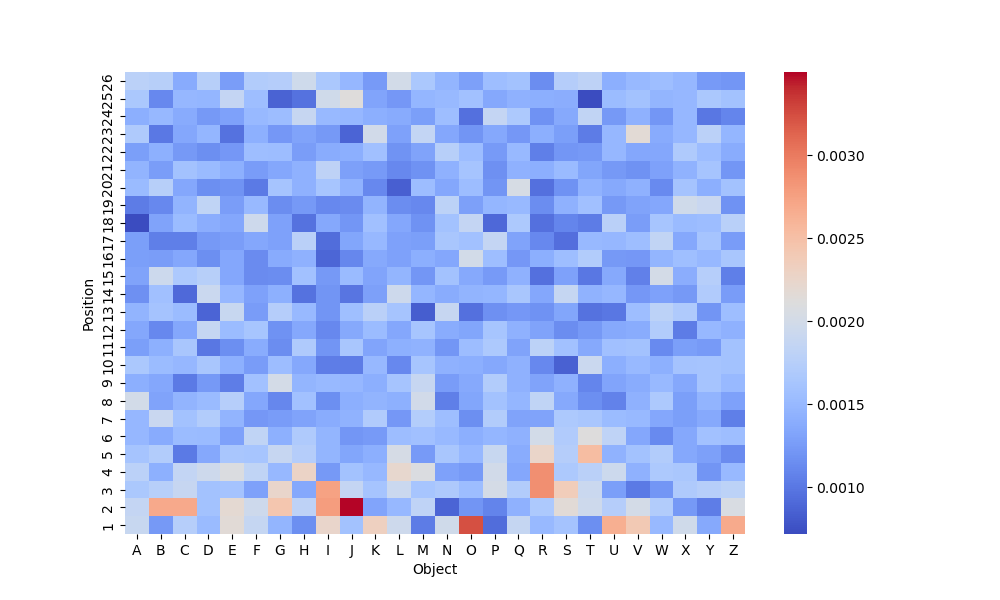}

}

\subcaption{T = 0.5, Count = 26}

\end{figure}%

\end{minipage}%
\newline
\begin{minipage}{0.20\linewidth}

\begin{figure}[H]

{\centering \includegraphics{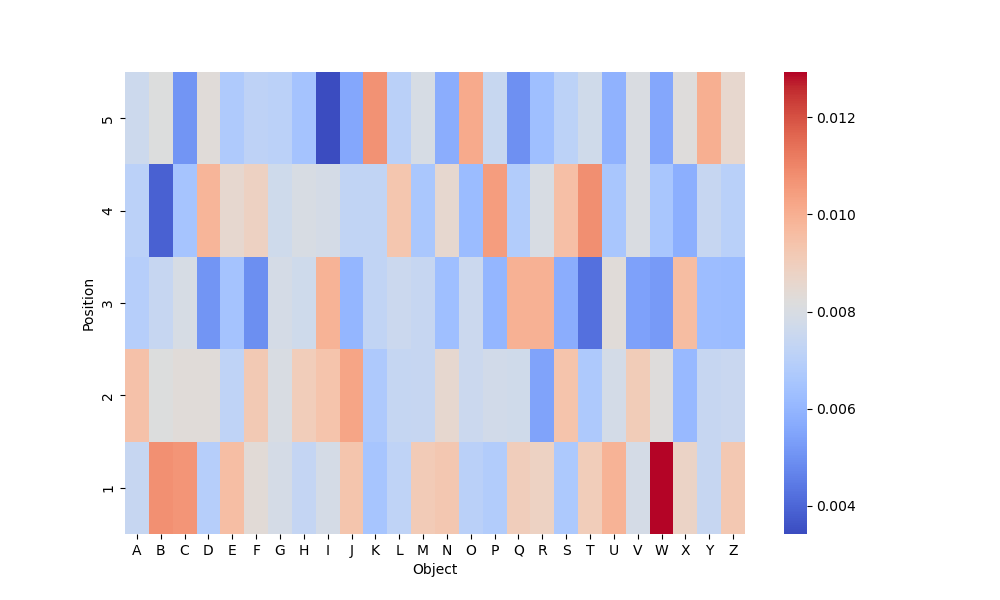}

}

\subcaption{T = 1, Count = 5}

\end{figure}%

\end{minipage}%
\begin{minipage}{0.20\linewidth}

\begin{figure}[H]

{\centering \includegraphics{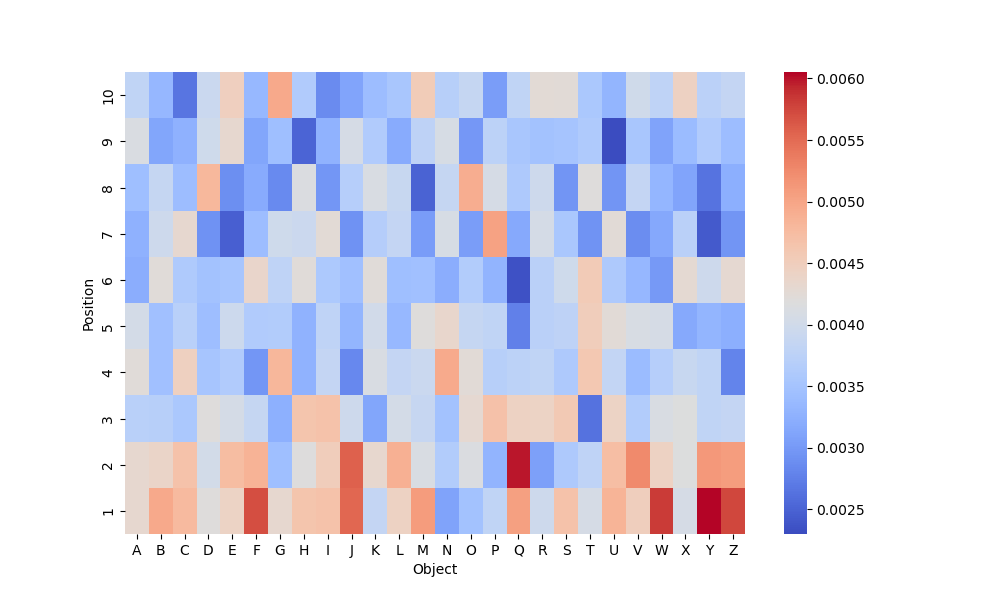}

}

\subcaption{T = 1, Count = 10}

\end{figure}%

\end{minipage}%
\begin{minipage}{0.20\linewidth}

\begin{figure}[H]

{\centering \includegraphics{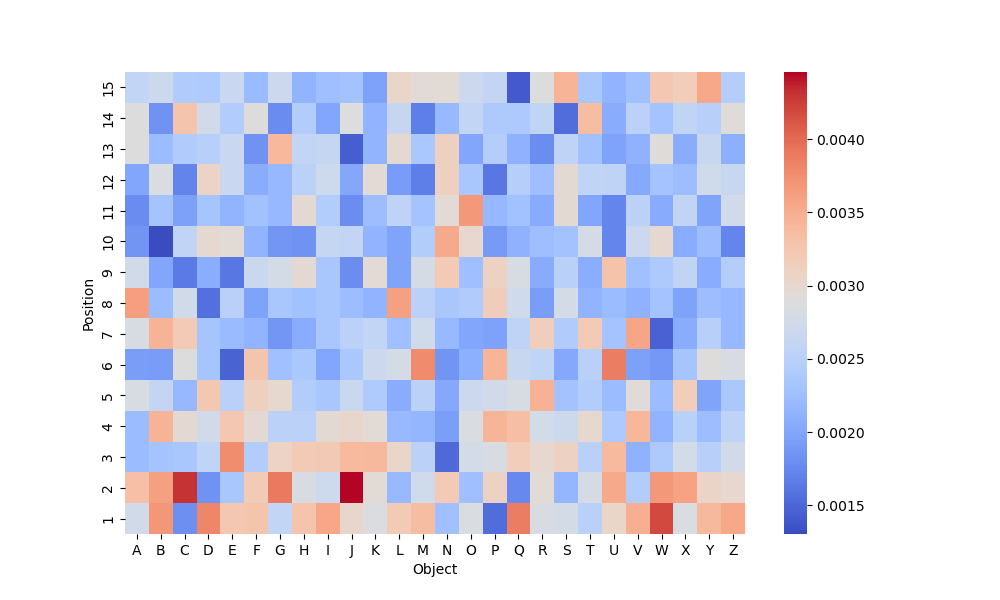}

}

\subcaption{T = 1, Count = 15}

\end{figure}%

\end{minipage}%
\begin{minipage}{0.20\linewidth}

\begin{figure}[H]

{\centering \includegraphics{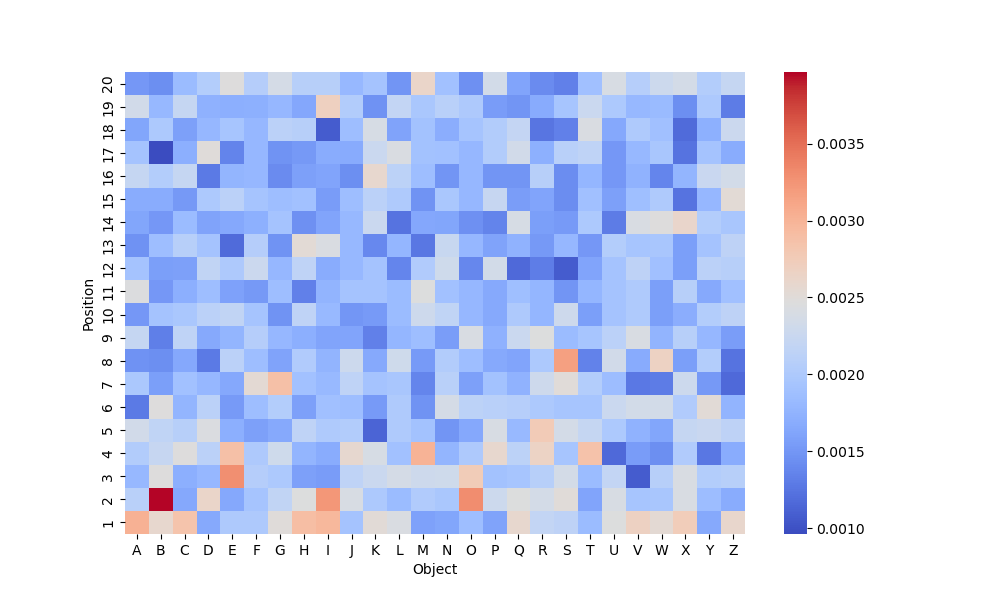}

}

\subcaption{T = 1, Count = 20}

\end{figure}%

\end{minipage}%
\begin{minipage}{0.20\linewidth}

\begin{figure}[H]

{\centering \includegraphics{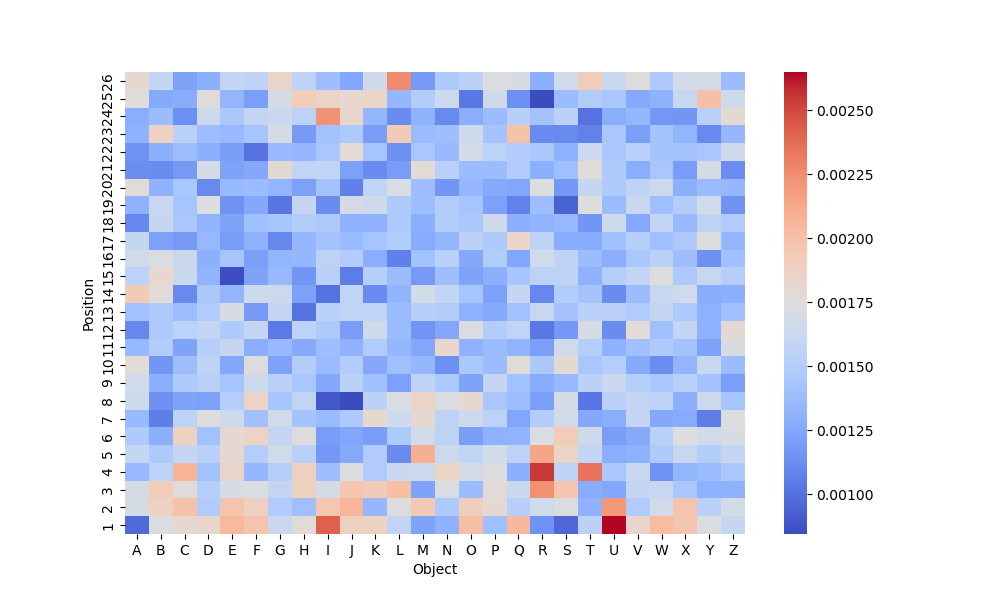}

}

\subcaption{T = 1, Count = 26}

\end{figure}%

\end{minipage}%
\newline
\begin{minipage}{0.20\linewidth}

\begin{figure}[H]

{\centering \includegraphics{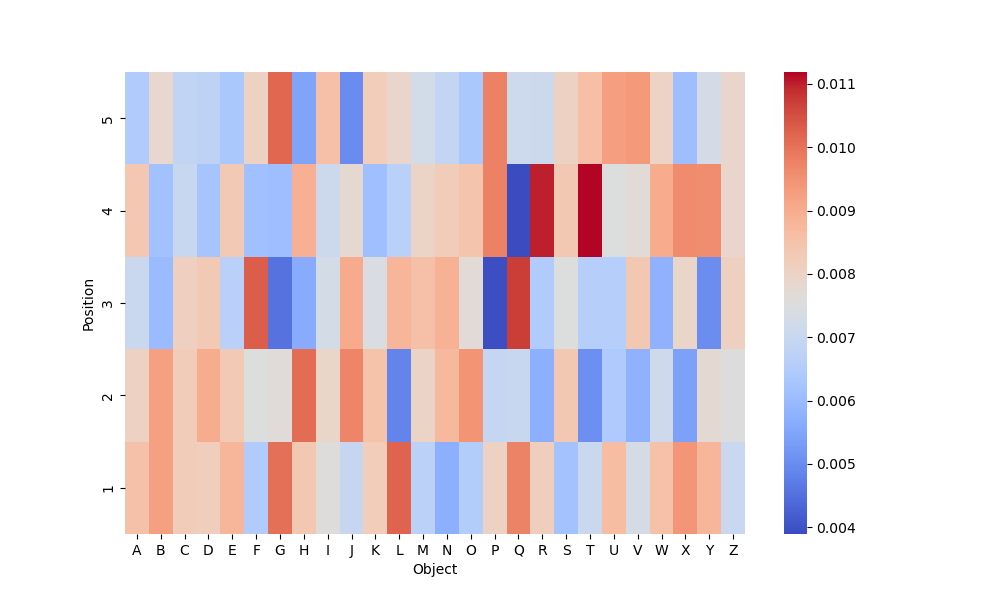}

}

\subcaption{T = 1.5, Count = 5}

\end{figure}%

\end{minipage}%
\begin{minipage}{0.20\linewidth}

\begin{figure}[H]

{\centering \includegraphics{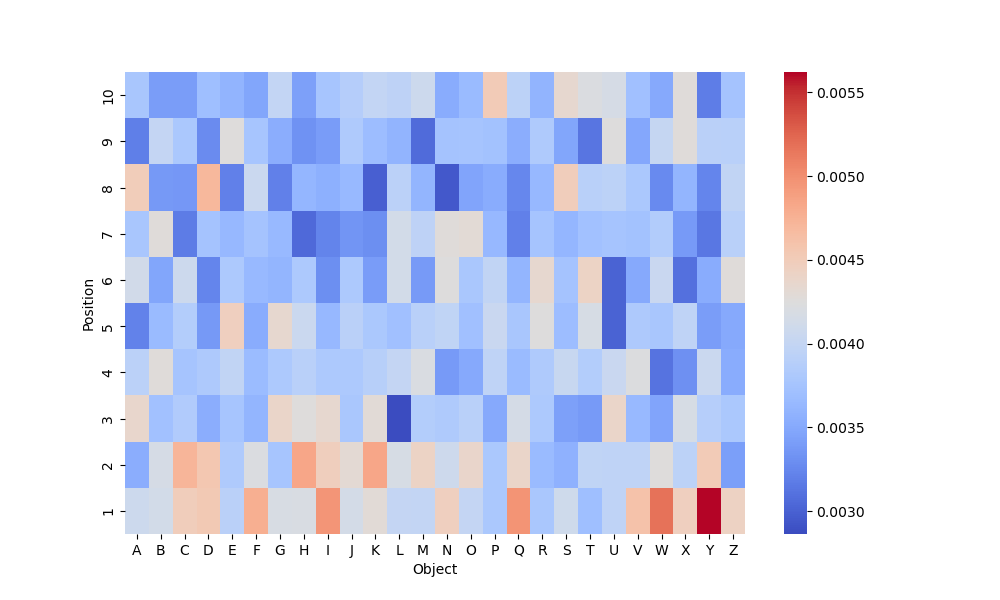}

}

\subcaption{T = 1.5, Count = 10}

\end{figure}%

\end{minipage}%
\begin{minipage}{0.20\linewidth}

\begin{figure}[H]

{\centering \includegraphics{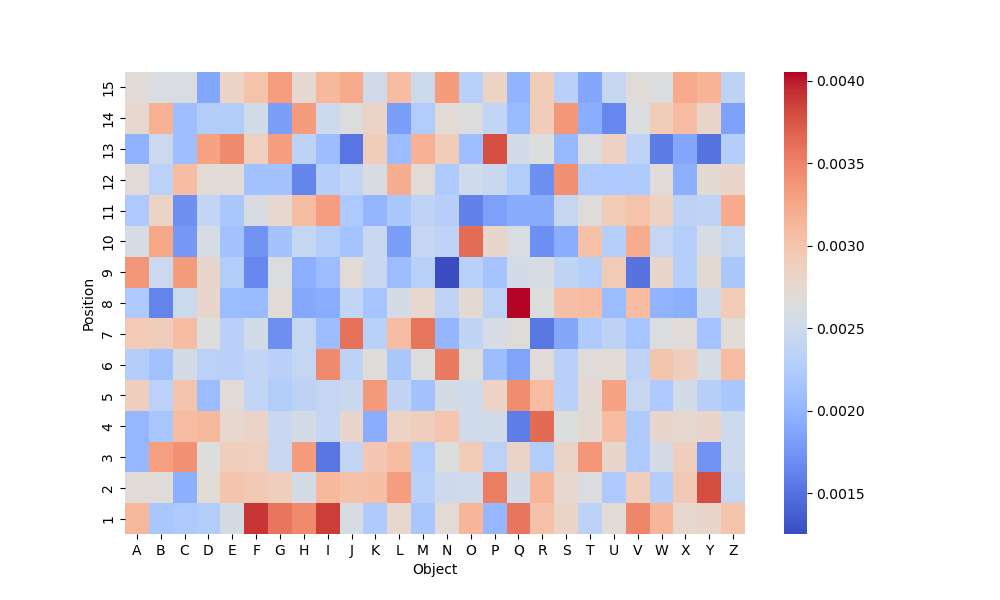}

}

\subcaption{T = 1.5, Count = 15}

\end{figure}%

\end{minipage}%
\begin{minipage}{0.20\linewidth}

\begin{figure}[H]

{\centering \includegraphics{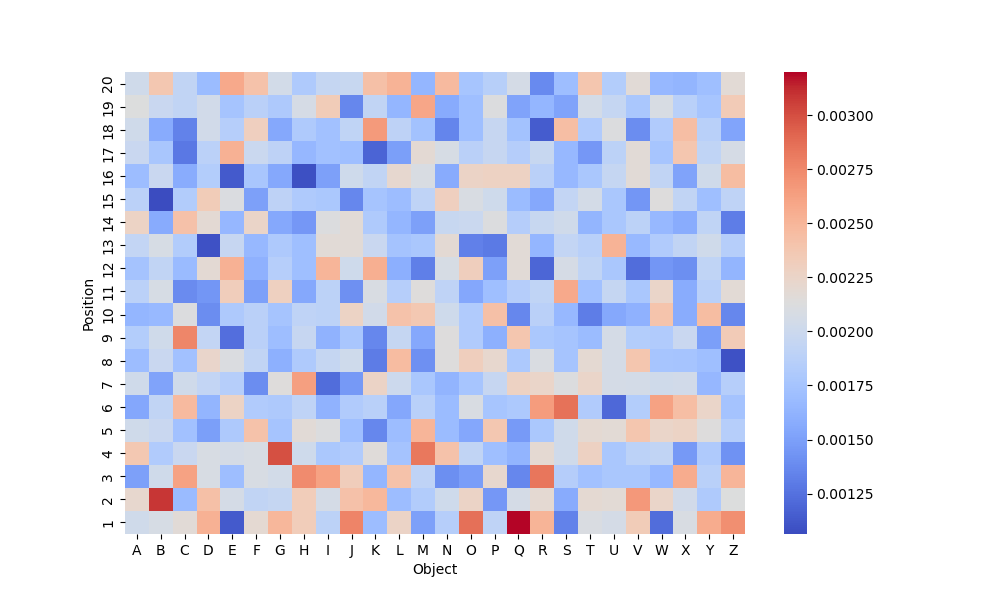}

}

\subcaption{T = 1.5, Count = 20}

\end{figure}%

\end{minipage}%
\begin{minipage}{0.20\linewidth}

\begin{figure}[H]

{\centering \includegraphics{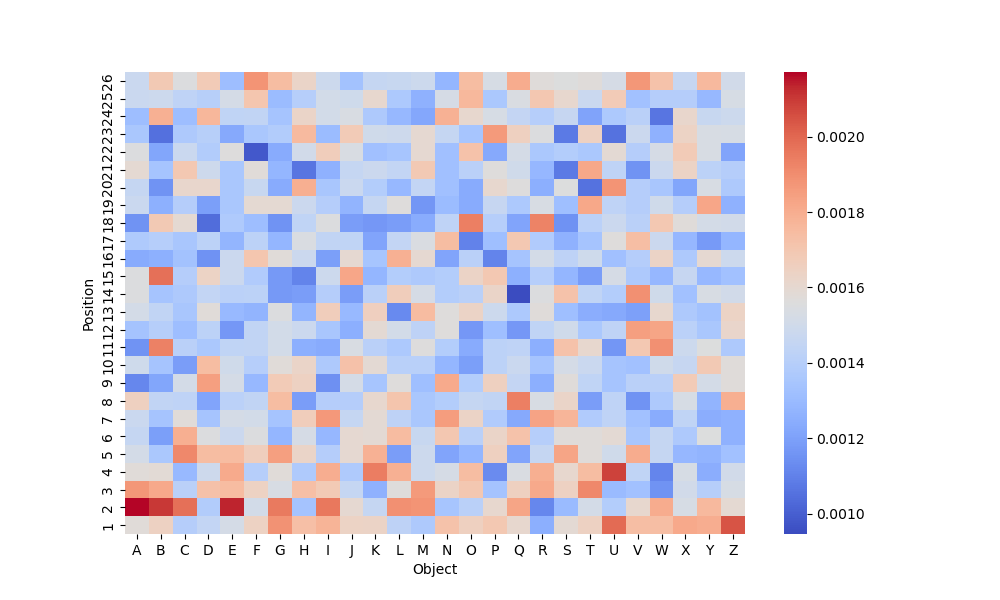}

}

\subcaption{T = 1.5, Count = 26}

\end{figure}%

\end{minipage}%

\caption{\label{fig-gpt35-joint-probability-letters}The joint
probability of gpt-3.5-turbo for letters and positions}

\end{figure}

The application of LLMs requires a careful eye to the prompt structure
and implementation. If, for example, one were to construct a list of
actions a LLM agent could undertake in response to some contextual data,
the appropriate generated answer would be in a tug-of-war with the
object-position bias (Figure~\ref{fig-gpt35-joint-probability-letters}).
Ideally, the model used will have a low mutual information between the
position and object, such as in claude-instant-1.2, allowing for
agnostic object order. That is not always the case
(Figure~\ref{fig-mutual-regression}), and needs to be tested empirically
for the use case at hand, especially due to the object identities and
hyperparameters creating wildly different mutual information
(Figure~\ref{fig-mi-direct-linear-regression}). A variety of methods
could be employed to handle this, the simplest being to sample the
positional bias and object bias of the LLM, and optimizing the order
such that it approaches the uniform distribution when no context is
provided. Such a method would also be amenable to the use of a thesaurus
to retain semantic meaning of actions while minimizing the bias from
selection.

When implementing an LLM, one must consider what tools and structures
are being imposed on the output and how that will bias the final outputs
(Figure~\ref{fig-guard-rails}). While the intrinsic bias for a model may
be low, complex prompt structures will strongly alter the probability
distribution of the output. In our study we were able to circumvent this
by inserting a sample step between the selection task and the JSON
structuring task. We propose a model of cognitive load to interpret this
result, with the two-step method reducing the total cognitive load of
our initial prompt, which fosters creativity and reduces bias while
taking advantage of the functionality of a guard rail for our work. Our
work was significantly sped up via the application of a guard rail over
attempting to perform arbitrary text matching. We achieved a promising
result for the use of guard rails in LLMs, but more work is needed to
understand the cognitive load of guard rails and other prompt
structures.

\section{Competing Interests}\label{competing-interests}

Rafael F. Irgoli\v{c} has contributed to Guardrails AI's open-source
project, and has been involved in contractual work with Guardrails AI.
The other authors declare that they have no competing interests.

\section{Authors' Contributions}\label{authors-contributions}

J.E. forumlated the problem, designed the experiments, analyzed the
data, and wrote the manuscript. R.F.I. designed and implemented the
experiments. All authors read, edited, and approved the final
manuscript.

\newpage{}

\section*{References}\label{references}
\addcontentsline{toc}{section}{References}

\phantomsection\label{refs}

\newpage{}

\section{Supplemental Information}\label{supplemental-information}

\begin{figure}

\centering{

\includegraphics{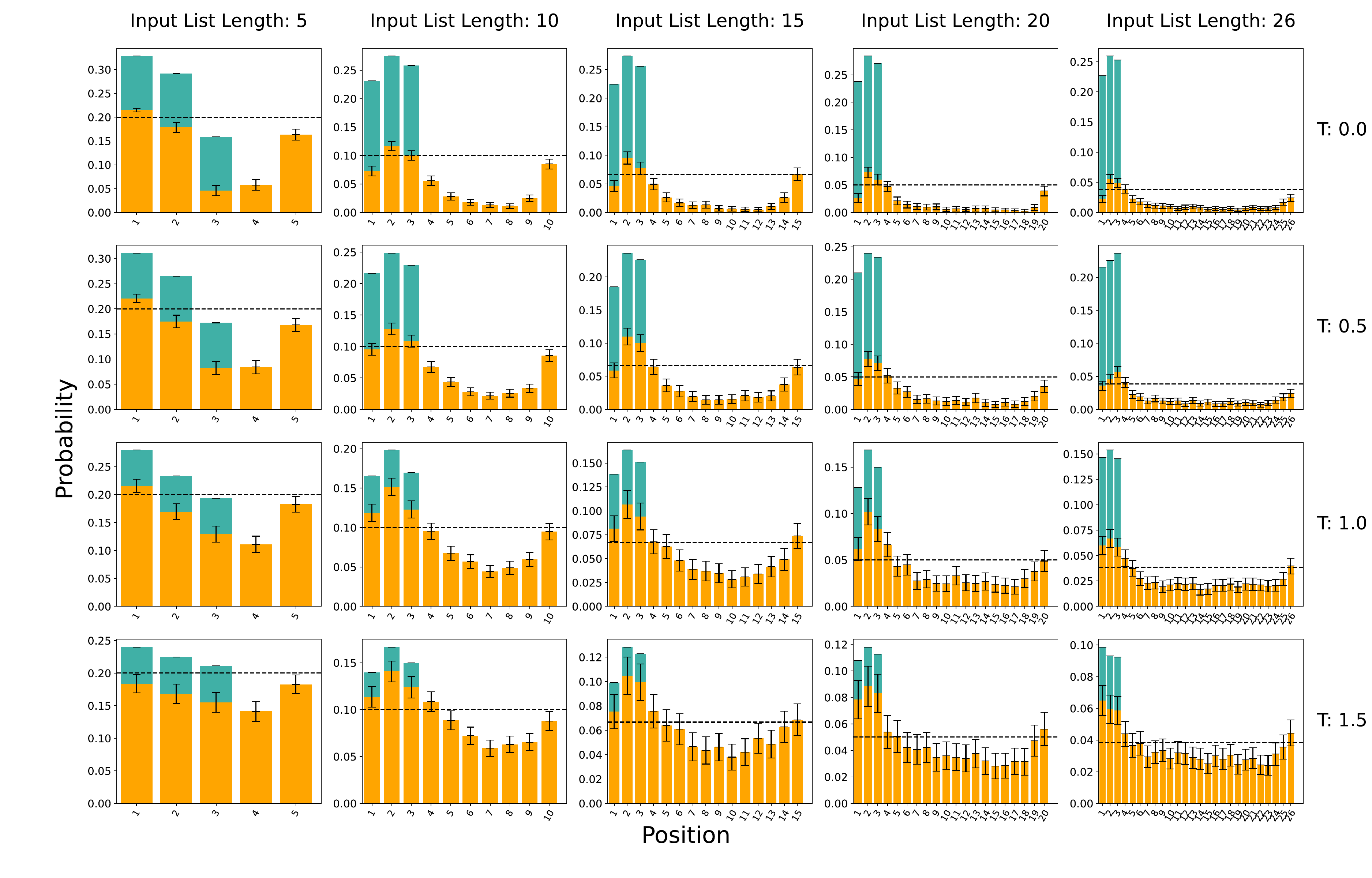}

}

\caption{\label{fig-number-position-gpt35}Using gpt-3.5-turbo the scaled
probability that a position of a number will be selected given a
temperature and original list length was computed. The dotted black line
is the expected probability given random sampling of a uniform
distribution for a list length. Orange bars are the probability that a
position without primacy bias will be selected, while blue represents
the probability that a value with primacy bias will be selected. Error
bars are standard error from 3000 bootstrap replicates.}

\end{figure}%

\begin{figure}

\begin{minipage}{0.20\linewidth}

\begin{figure}[H]

{\centering \includegraphics{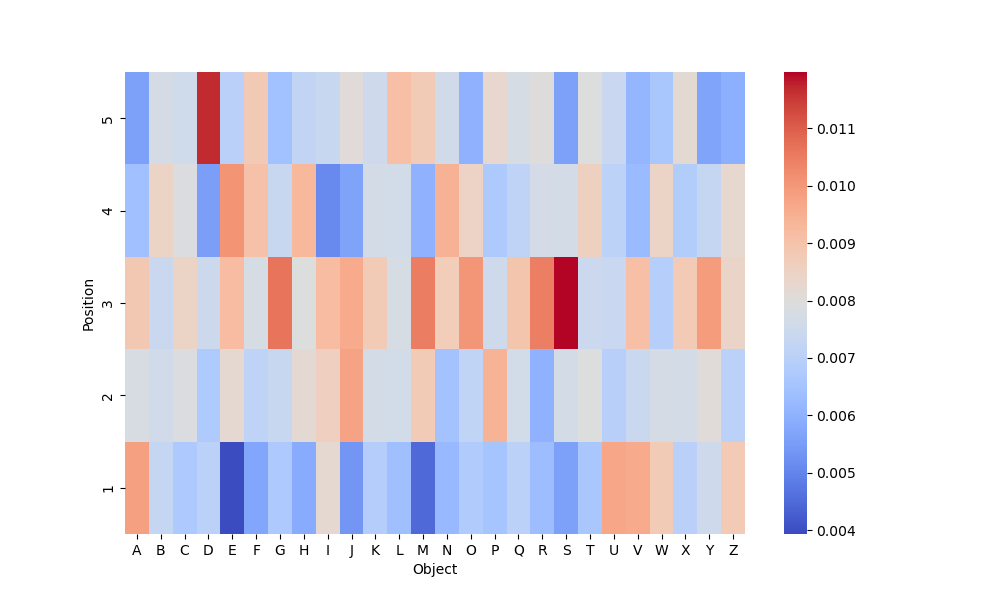}

}

\subcaption{T = -1, Count = 5}

\end{figure}%

\end{minipage}%
\begin{minipage}{0.20\linewidth}

\begin{figure}[H]

{\centering \includegraphics{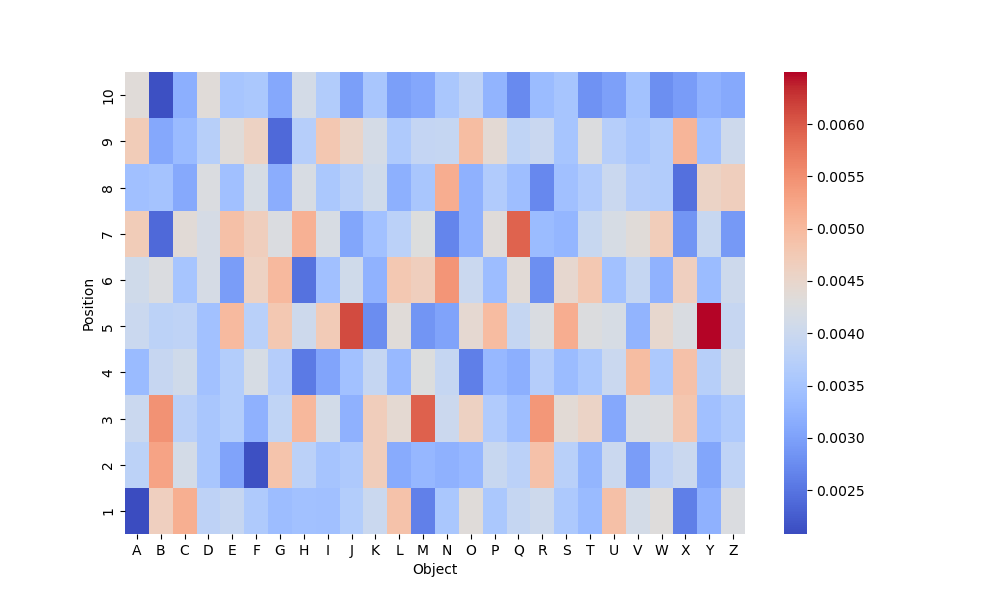}

}

\subcaption{T = -1, Count = 10}

\end{figure}%

\end{minipage}%
\begin{minipage}{0.20\linewidth}

\begin{figure}[H]

{\centering \includegraphics{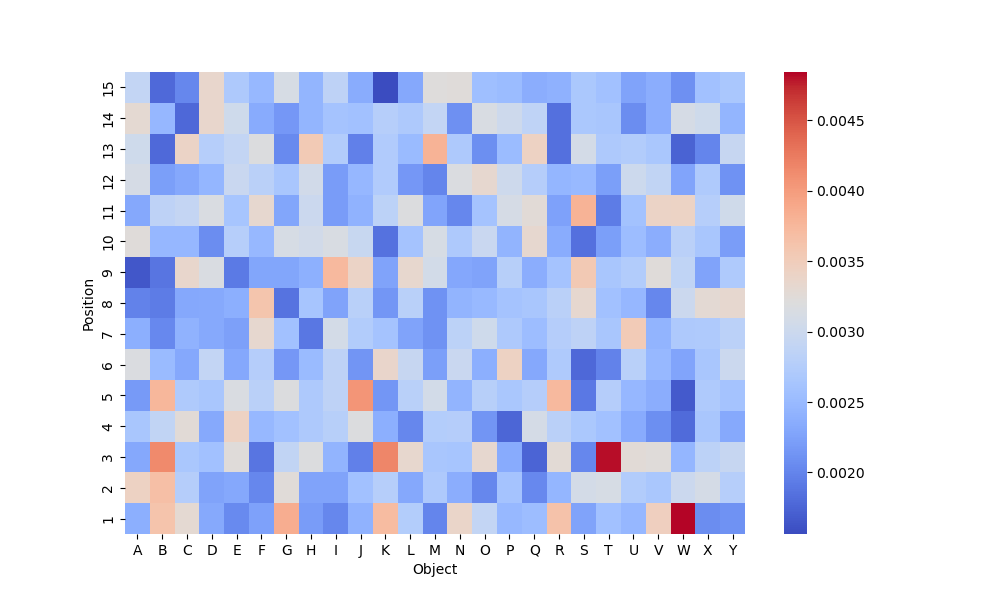}

}

\subcaption{T = -1, Count = 15}

\end{figure}%

\end{minipage}%
\begin{minipage}{0.20\linewidth}

\begin{figure}[H]

{\centering \includegraphics{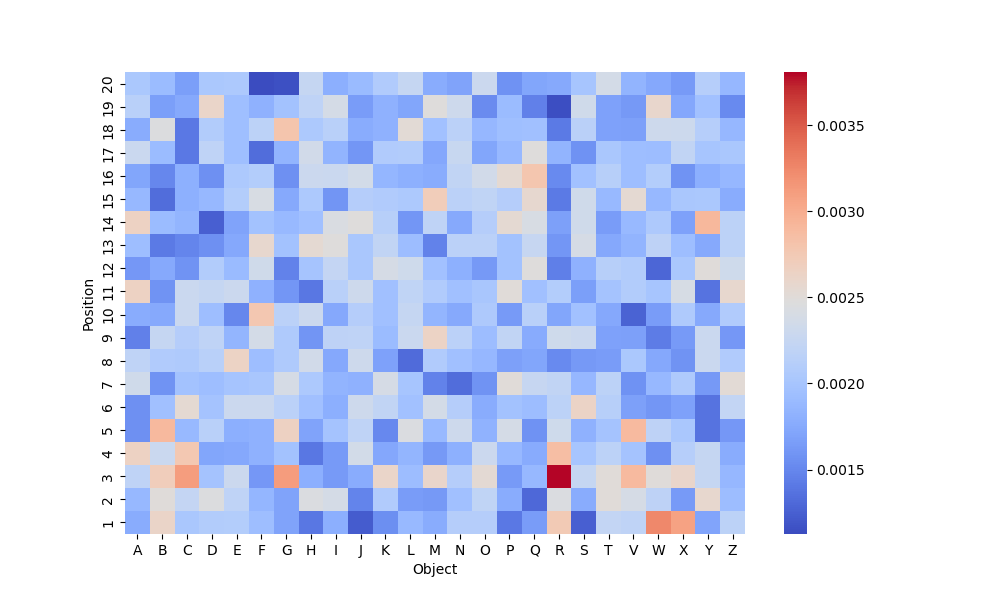}

}

\subcaption{T = -1, Count = 20}

\end{figure}%

\end{minipage}%
\begin{minipage}{0.20\linewidth}

\begin{figure}[H]

{\centering \includegraphics{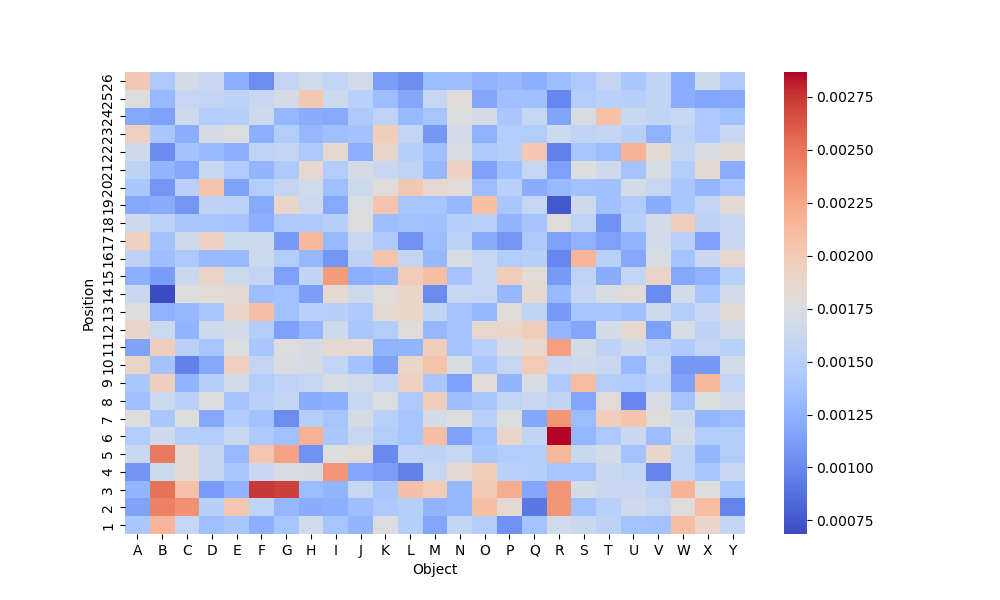}

}

\subcaption{T = -1, Count = 26}

\end{figure}%

\end{minipage}%
\newline
\begin{minipage}{0.20\linewidth}

\begin{figure}[H]

{\centering \includegraphics{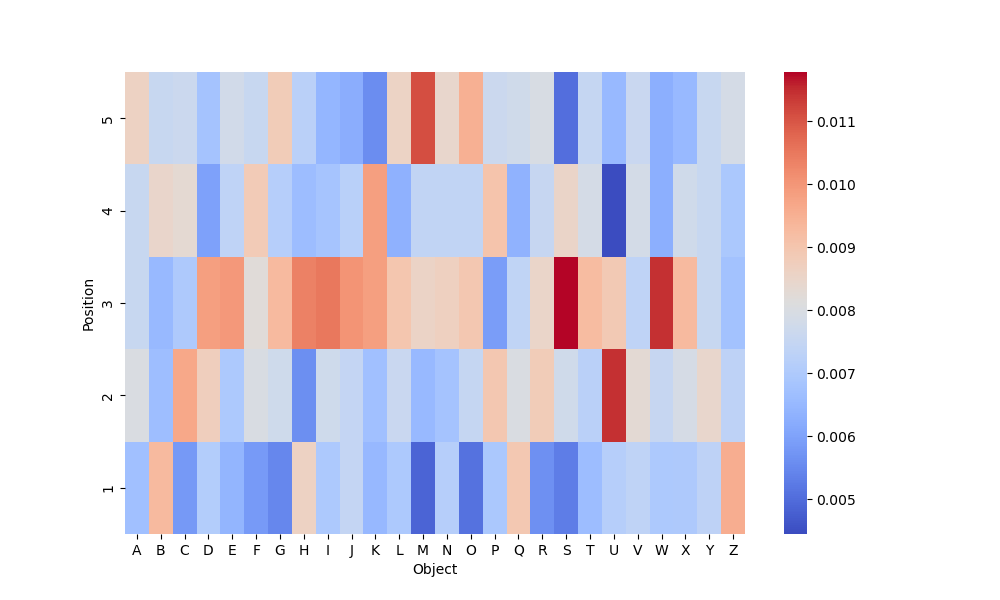}

}

\subcaption{T = 0, Count = 5}

\end{figure}%

\end{minipage}%
\begin{minipage}{0.20\linewidth}

\begin{figure}[H]

{\centering \includegraphics{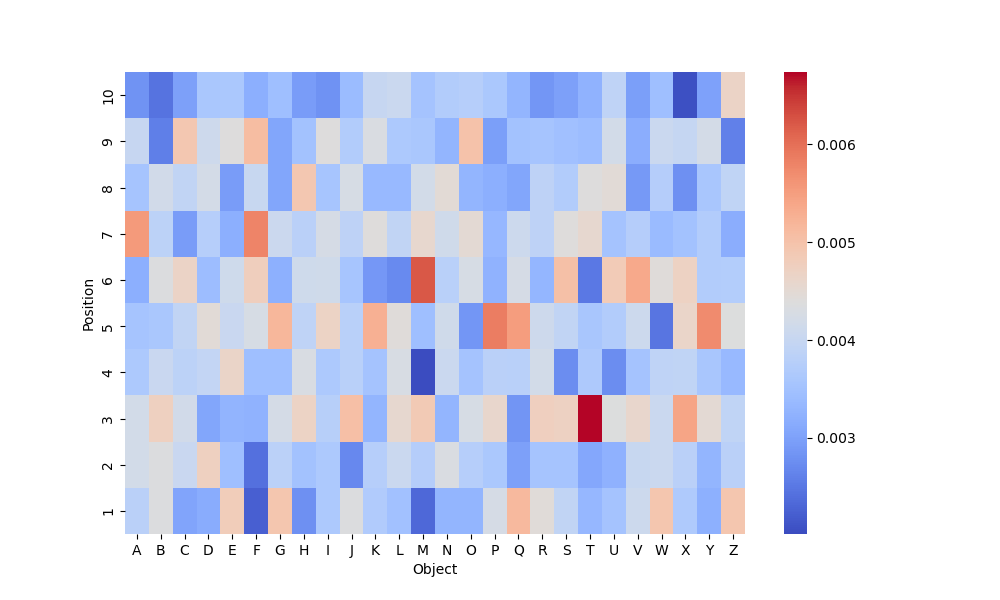}

}

\subcaption{T = 0, Count = 10}

\end{figure}%

\end{minipage}%
\begin{minipage}{0.20\linewidth}

\begin{figure}[H]

{\centering \includegraphics{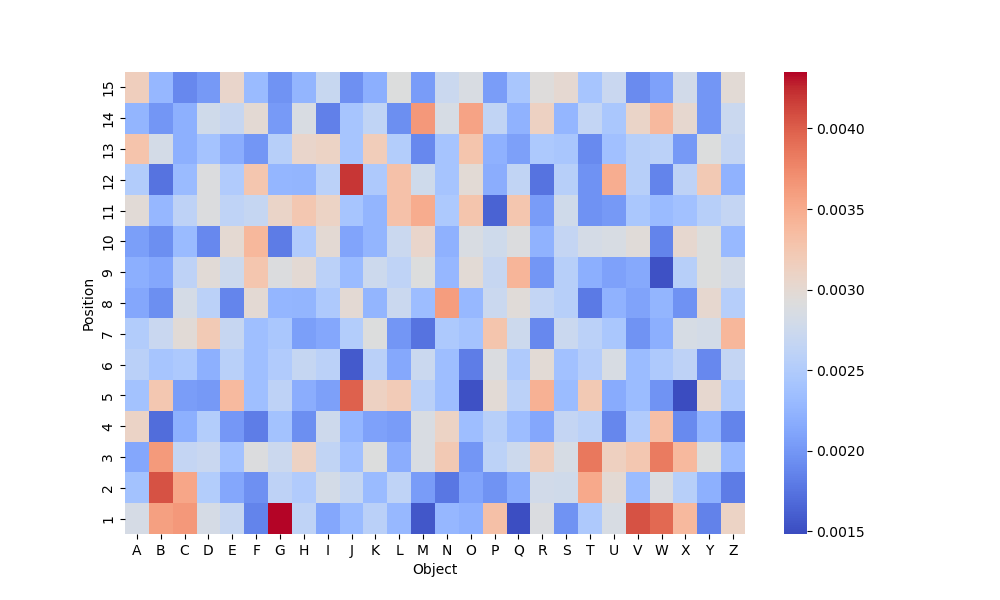}

}

\subcaption{T = 0, Count = 15}

\end{figure}%

\end{minipage}%
\begin{minipage}{0.20\linewidth}

\begin{figure}[H]

{\centering \includegraphics{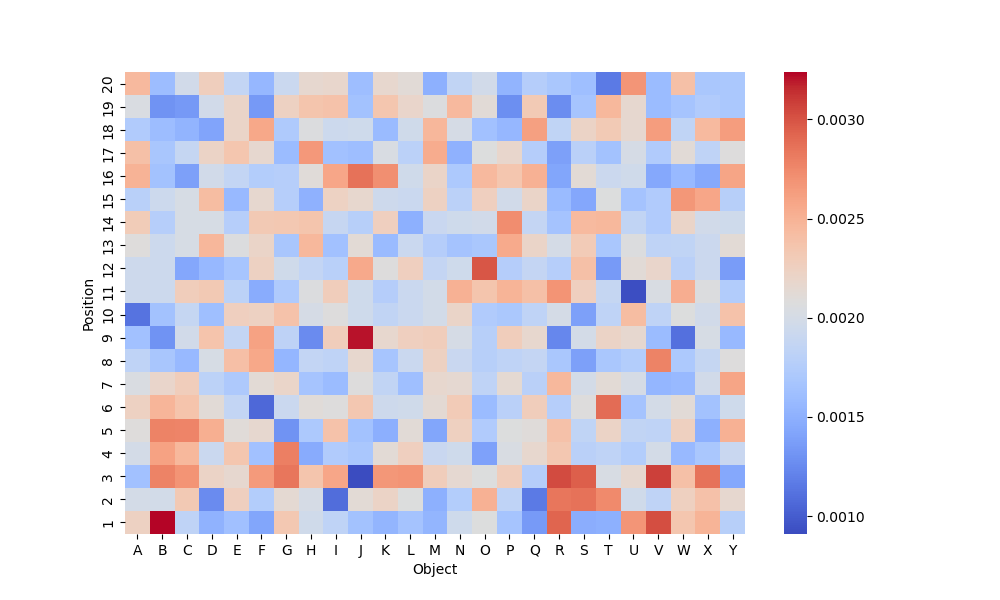}

}

\subcaption{T = 0, Count = 20}

\end{figure}%

\end{minipage}%
\begin{minipage}{0.20\linewidth}

\begin{figure}[H]

{\centering \includegraphics{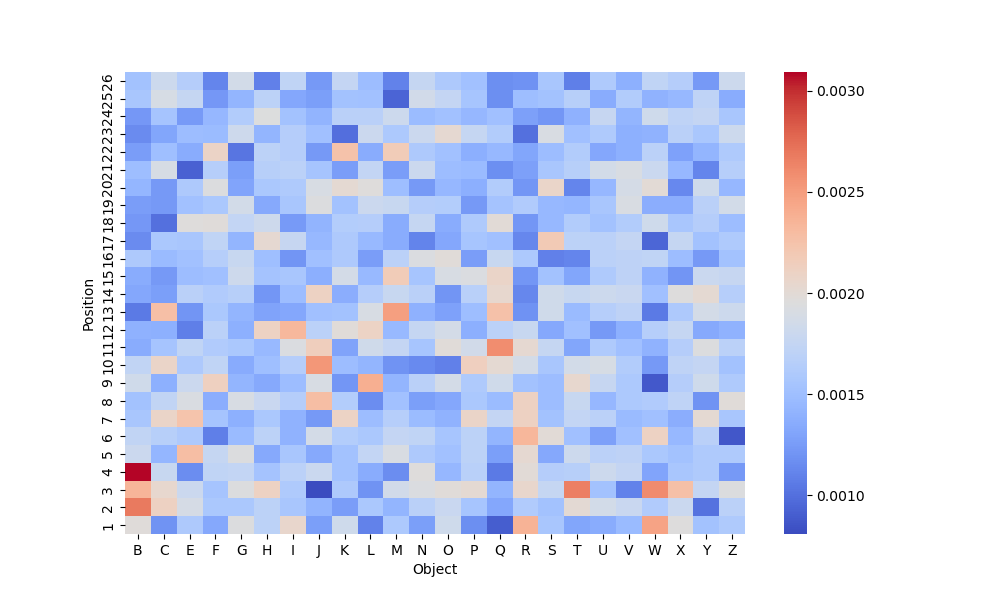}

}

\subcaption{T = 0, Count = 26}

\end{figure}%

\end{minipage}%
\newline
\begin{minipage}{0.20\linewidth}

\begin{figure}[H]

{\centering \includegraphics{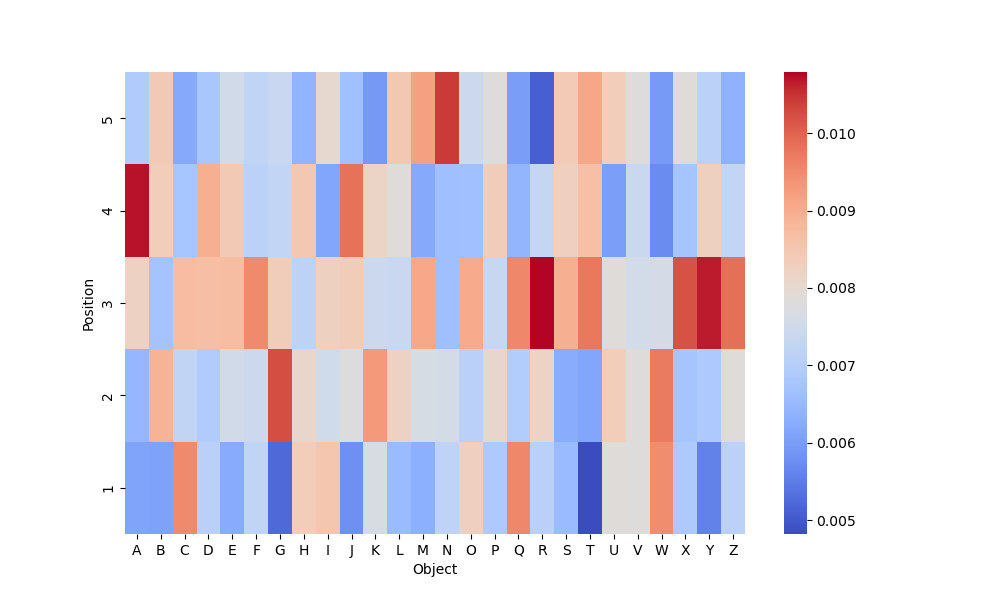}

}

\subcaption{T = 0.5, Count = 5}

\end{figure}%

\end{minipage}%
\begin{minipage}{0.20\linewidth}

\begin{figure}[H]

{\centering \includegraphics{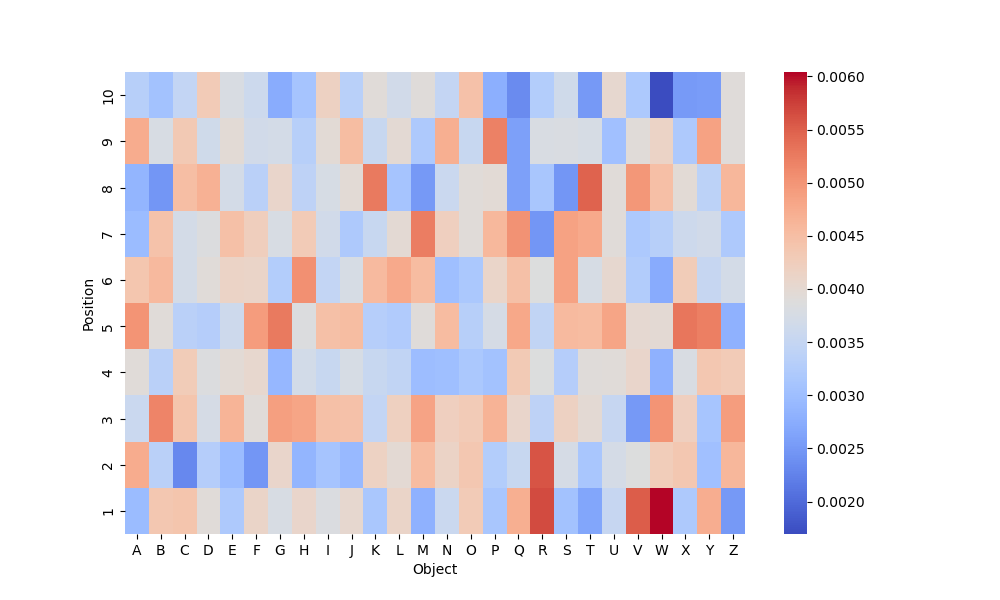}

}

\subcaption{T = 0.5, Count = 10}

\end{figure}%

\end{minipage}%
\begin{minipage}{0.20\linewidth}

\begin{figure}[H]

{\centering \includegraphics{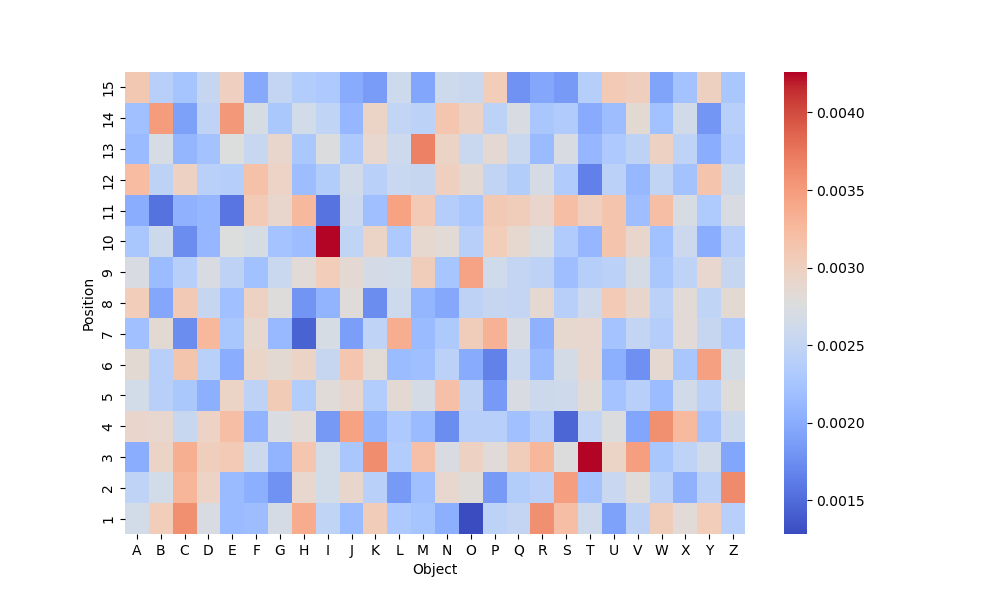}

}

\subcaption{T = 0.5, Count = 15}

\end{figure}%

\end{minipage}%
\begin{minipage}{0.20\linewidth}

\begin{figure}[H]

{\centering \includegraphics{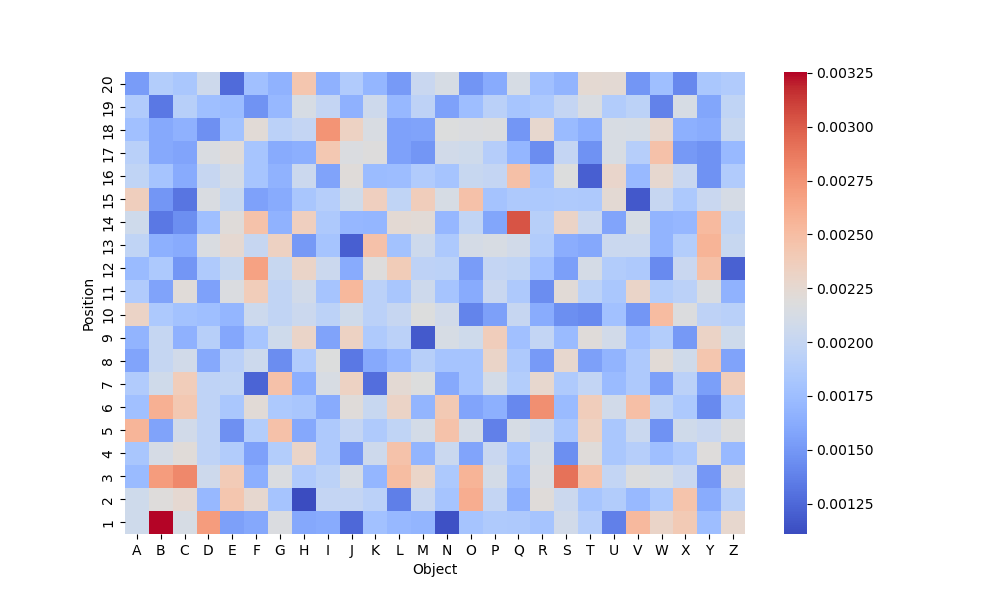}

}

\subcaption{T = 0.5, Count = 20}

\end{figure}%

\end{minipage}%
\begin{minipage}{0.20\linewidth}

\begin{figure}[H]

{\centering \includegraphics{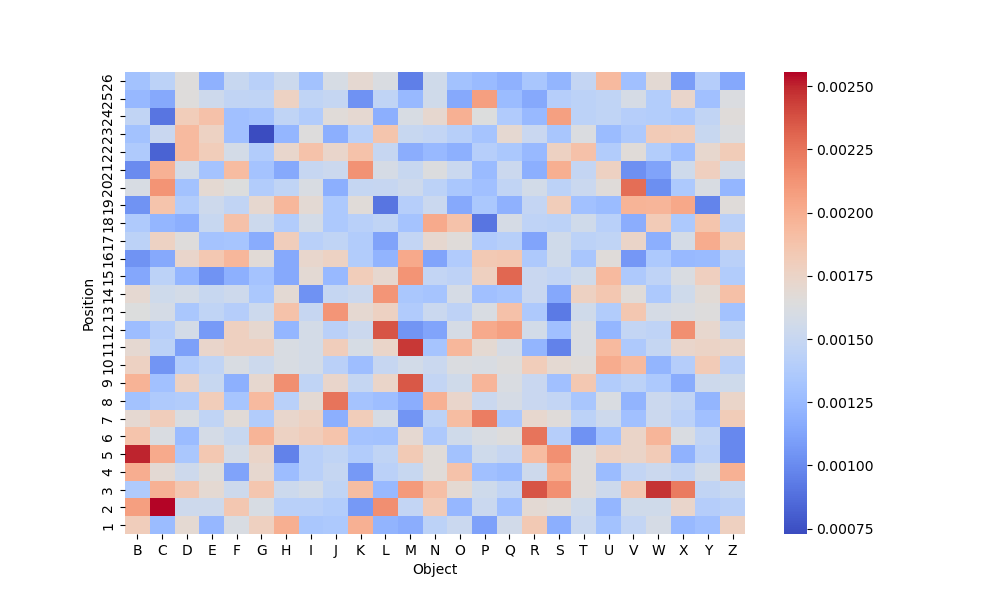}

}

\subcaption{T = 0.5, Count = 26}

\end{figure}%

\end{minipage}%
\newline
\begin{minipage}{0.20\linewidth}

\begin{figure}[H]

{\centering \includegraphics{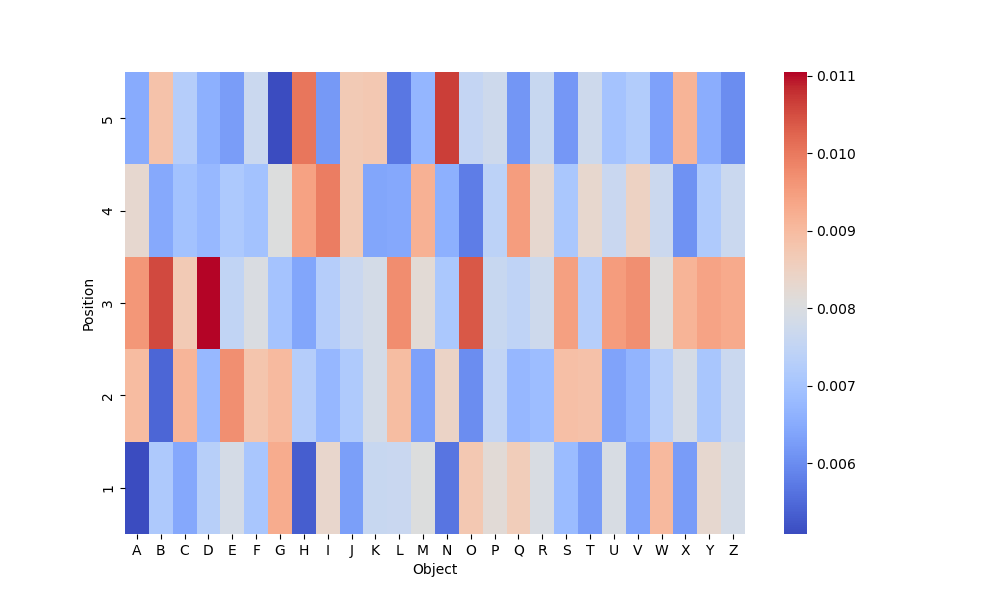}

}

\subcaption{T = 1, Count = 5}

\end{figure}%

\end{minipage}%
\begin{minipage}{0.20\linewidth}

\begin{figure}[H]

{\centering \includegraphics{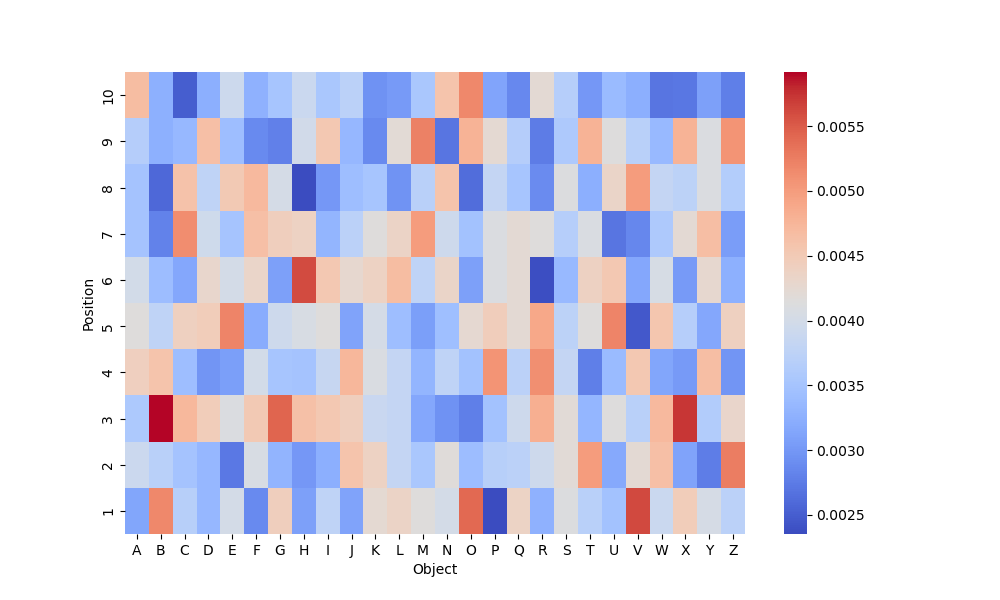}

}

\subcaption{T = 1, Count = 10}

\end{figure}%

\end{minipage}%
\begin{minipage}{0.20\linewidth}

\begin{figure}[H]

{\centering \includegraphics{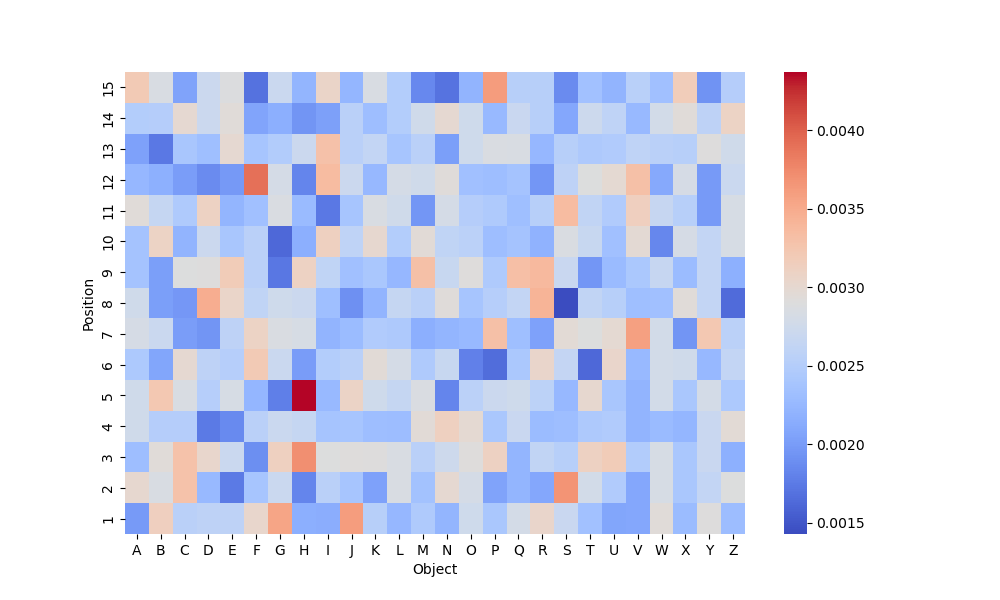}

}

\subcaption{T = 1, Count = 15}

\end{figure}%

\end{minipage}%
\begin{minipage}{0.20\linewidth}

\begin{figure}[H]

{\centering \includegraphics{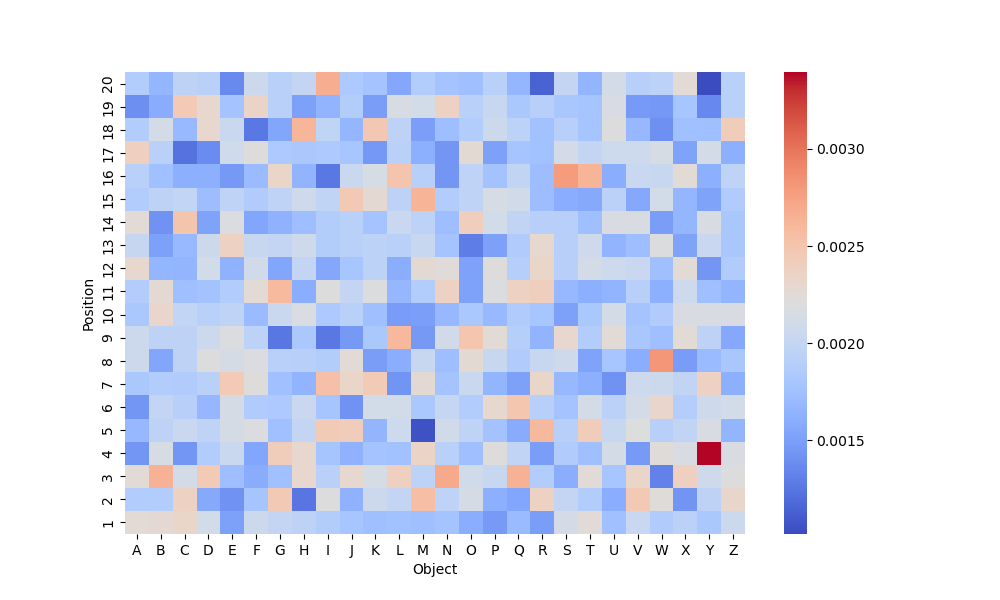}

}

\subcaption{T = 1, Count = 20}

\end{figure}%

\end{minipage}%
\begin{minipage}{0.20\linewidth}

\begin{figure}[H]

{\centering \includegraphics{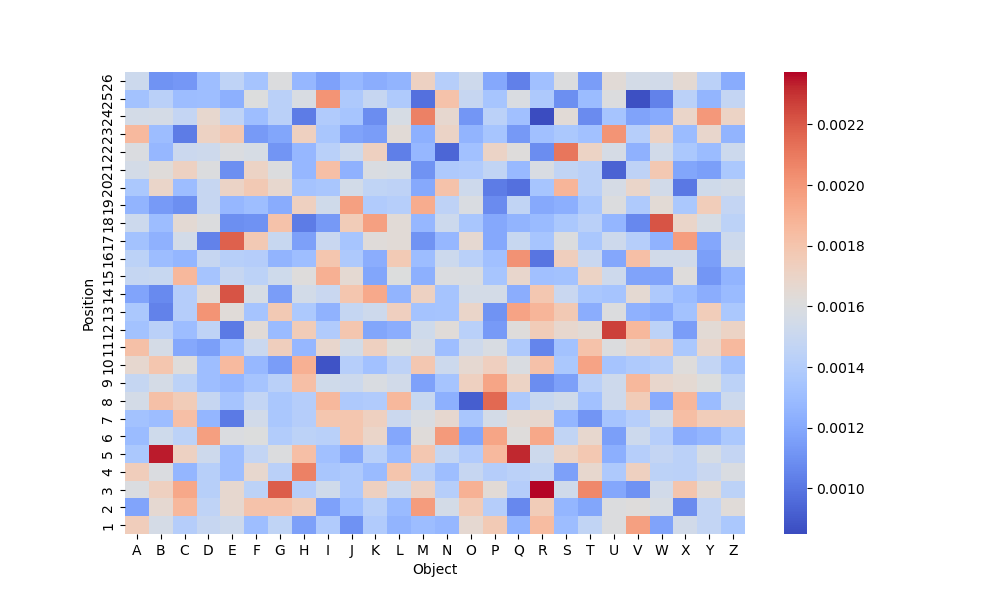}

}

\subcaption{T = 1, Count = 26}

\end{figure}%

\end{minipage}%

\caption{\label{fig-claude-joint-probability-letters}The joint
probability of claude-instant-1.2 for letters and positions}

\end{figure}%

\begin{codelisting}

\caption{\label{lst-prompt-construct}Code to construct initial prompt
for data collection}

\centering{

\begin{verbatim}
def construct_prompt(choices: list[str], choice_count: int):
    prompt = f"""Please select {choice_count} of the following:"""
    for i, choice in enumerate(choices):
        prompt += f"\n- {choice}"

    return prompt
\end{verbatim}

}

\end{codelisting}%

\begin{codelisting}

\caption{\label{lst-spacing-step-collection}Code to perform spacing step
data collection via LLM call, here \{prompt\} refers to the constructed
prompt from Listing~\ref{lst-prompt-construct}, while
\{initial\_response\} refers to the result of the first prompt.}

\centering{

\begin{verbatim}

<rail version="0.1">

<output>
<list 
    name="choices"
>
</list>
</output>

<instructions>
You are a helpful assistant only capable of communicating with valid JSON, 
and no other text.

${{gr.json_suffix_prompt_examples}}
</instructions>

<prompt>
+++
{initial_response}
+++

${{gr.xml_prefix_prompt}}

${{output_schema}}

Your returned value should be a dictionary with a single "choices" key, 
whose value contains a list of values chosen in the above response enclosed in +++.

</prompt>


</rail>
\end{verbatim}

}

\end{codelisting}%

\begin{codelisting}

\caption{\label{lst-direct-data-collection}Code to perform direct data
collection via LLM call, here \{prompt\} refers to the constructed
prompt from Listing~\ref{lst-prompt-construct}}

\centering{

\begin{verbatim}
<rail version="0.1">

<output>
<list 
    name="choices"
    format="length: {choice_count} {choice_count}"
    on-fail-format="noop"
>
<choice>
{"".join(f'''<case name="{choice}">
</case>''' for choice in choices)}
</choice>
</list>
</output>

<instructions>
You are a helpful assistant only capable of communicating with valid JSON, 
and no other text.

${{gr.json_suffix_prompt_examples}}
</instructions>

<prompt>
{prompt}

${{gr.xml_prefix_prompt}}

${{output_schema}}

</prompt>

</rail>
\end{verbatim}

}

\end{codelisting}%

\end{document}